\documentclass{article}

\usepackage{arxiv}
\usepackage{url}
\usepackage{fancyhdr}
\usepackage{multirow}
\usepackage{diagbox}
\usepackage[T1]{fontenc}
\usepackage[numbers]{natbib}
\RequirePackage{algorithm}
\usepackage[backref]{hyperref}

\usepackage{bibentry}
\usepackage{amssymb}


\usepackage{xcolor}
\definecolor{lightgray}{gray}{0.893}
\usepackage{colortbl}

\usepackage{xcolor}         
\usepackage{bm}
\usepackage{amsmath}
\usepackage{amsthm} 
\usepackage{multicol} 

\usepackage{subcaption}

\usepackage{algorithm}
\usepackage{algorithmic}
\usepackage{soul}
\usepackage{booktabs}

\usepackage{graphicx}

\title{Multi-agent Embodied AI:\\ Advances and Future Directions}

\author{%
 Zhaohan Feng$^{1}$, Ruiqi Xue$^{2}$, Lei Yuan$^{2}$, Yang Yu$^{2}$, Ning Ding$^{3}$, Meiqin Liu$^{3,4}$, Bingzhao Gao$^{5}$, \\
 \textbf{Jian Sun$^{1}$, Xinhu Zheng$^{6}$, and Gang Wang$^{1}$\thanks{Corresponding Author}}\\
 \\
  $^1$ National Key Lab of Autonomous Intelligent Unmanned Systems, \\
  Beijing Institute of Technology, Beijing 100081, China\\
  $^2$ National Key Laboratory for Novel Software Technology, \\
  Nanjing University, Nanjing 210000, China\\
  $^3$ State Key Laboratory of Human-Machine Hybrid Augmented Intelligence, \\
  Xi’an Jiaotong University, Xi’an 710049, China\\
  $^4$ College of Electrical Engineering, Zhejiang University, Hangzhou 310007, China\\
  $^5$ Clean Energy Automotive Engineering Center, Tongji University, Shanghai 201804, China\\
  $^6$ Intelligent Transportation Thrust, \\
  The Hong Kong University of Science and Technology (Guangzhou), Guangzhou 511453, China\\
  \\
\texttt{zhfeng@bit.edu.cn},~~\texttt{xuerq@lamda.nju.edu.cn},~~\texttt{yuanl@lamda.nju.edu.cn},\\
\texttt{yuy@nju.edu.cn},~~\texttt{ding.ning@xjtu.edu.cn},~~\texttt{liumeiqin@zju.edu.cn},\\
\texttt{gaobz@jlu.edu.cn},~~\texttt{sunjian@bit.edu.cn},~~\texttt{xinhuzheng@hkust-gz.edu.cn},~~\texttt{gangwang@bit.edu.cn}\\
}

\begin{document}

\maketitle
\date{}
\begin{abstract}
Embodied artificial intelligence (Embodied AI) plays a pivotal role in the application of advanced technologies in the intelligent era, where AI systems are integrated with physical bodies that enable them to perceive, reason, and interact with their environments. Through the use of sensors for input and actuators for action, these systems can learn and adapt based on real-world feedback, allowing them to perform tasks effectively in dynamic and unpredictable environments. As techniques such as deep learning (DL), reinforcement learning (RL), and large language models (LLMs) mature, embodied AI has become a leading field in both academia and industry, with applications spanning robotics, healthcare, transportation, and manufacturing. However, most research has focused on single-agent systems that often assume static, closed environments, whereas real-world embodied AI must navigate far more complex scenarios. In such settings, agents must not only interact with their surroundings but also collaborate with other agents, necessitating sophisticated mechanisms for adaptation, real-time learning, and collaborative problem-solving. Despite increasing interest in multi-agent systems, existing research remains narrow in scope, often relying on simplified models that fail to capture the full complexity of dynamic, open environments for multi-agent embodied AI. Moreover, no comprehensive survey has systematically reviewed the advancements in this area. As embodied AI rapidly evolves, it is crucial to deepen our understanding of multi-agent embodied AI to address the challenges presented by real-world applications. To fill this gap and foster further development in the field, this paper reviews the current state of research, analyzes key contributions, and identifies challenges and future directions, providing insights to guide innovation and progress in this field.
\end{abstract}

\section{Introduction} \label{sec:introduction}
Embodied artificial intelligence (Embodied AI) \cite{pfeifer2004embodied,duan2022survey} is an interdisciplinary research area at the intersection of artificial intelligence (AI), robotics, and cognitive science, aimed at equipping robots with the capabilities to perceive, plan, decide, and act, thereby enabling them to interact with and adapt to their environments actively. This concept was first introduced by Alan Turing in the 1950s, who explored how machines could perceive the world and make corresponding decisions~\cite{turing1950computing}. Later, in the 1980s, researchers including Rodney Brooks reconsidered symbolic AI, suggesting that intelligence should emerge through active interaction with the environment rather than passive data learning, thus laying the foundation for embodied AI~\cite{brooks1991new}. In recent years, driven by advances in deep learning (DL), reinforcement learning (RL), and other techniques, embodied AI has made significant progress, particularly through the application of large pre-trained models such as large language models (LLMs) and vision-language models (VLMs). These models have enhanced intelligent systems’ capabilities in perception, decision-making, and task planning, achieving outstanding performance in visual understanding, natural language processing, and multi-modal integration tasks~\cite{wang2023large}. Leveraging these large models, embodied AI has achieved breakthroughs in image recognition, speech interaction, and robotic control, improving robots' ability to understand and adapt to complex dynamic environments~\cite{liu2024aligning}.

The primary challenge in embodied AI is that intelligent agents must simultaneously possess robust perception and decision-making capabilities, along with the capacity to continuously learn and adapt through ongoing interaction with dynamic and evolving environments~\cite{roy2021machine}. Historically, early symbolic approaches, exemplified by Turing machine theory, attempted to achieve intelligence via symbolic representation and logical reasoning. However, these symbolic methods fell short in effectively addressing the dynamic interactions required between perception and action~\cite{newel1976computer}. To overcome these limitations, Rodney Brooks proposed the concept of the  ``perception-action loop'', which posits that intelligence emerges naturally through an agent's active and continuous engagement with its environment, thus laying the foundation for modern embodied AI research~\cite{brooks1986robust}. Following this line of thought, learning paradigms such as imitation learning (IL) emerged, accelerating the learning process by mimicking expert demonstrations, yet remaining heavily reliant on expert-generated data and consequently vulnerable to challenges like distributional shifts~\cite{Zare2024ImitationSurvey}. As an alternative, RL, especially deep RL, integrates deep neural networks to optimize decision-making strategies through interaction-driven feedback from the environment~\cite{eppe2020hierarchical}. Such methods excel at managing high-dimensional sensory inputs, demonstrating strong adaptability across various complex tasks. More recently, the advent of large-scale models, particularly LLMs \cite{xu2024survey} and VLMs \cite{ma2024survey}, has significantly advanced embodied AI, enhancing agents' capabilities in visual understanding, multi-modal perception, and task planning. Nevertheless, despite these advancements, most current research primarily addresses single-agent scenarios and often overlooks the inherent complexity of multi-agent interactions. In contrast, real-world embodied AI applications frequently demand effective collaboration among multiple heterogeneous agents to achieve tasks such as confronting unknown adversaries or addressing complex dynamic challenges~\cite{dorri2018multi}. 
\begin{figure*}
\centering
\includegraphics[width=0.98\columnwidth]{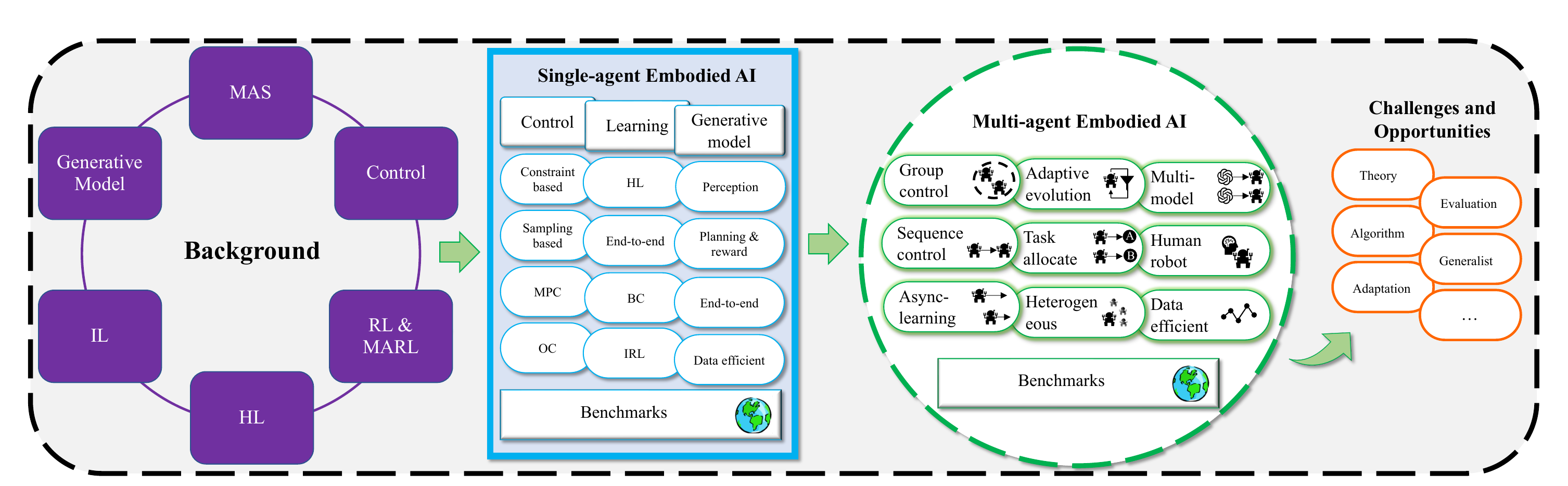}
\caption{An overview of this survey.}
\label{framwork}
\end{figure*}

Multi-agent settings fundamentally differ from single-agent scenarios, as agents must simultaneously optimize their individual policies and manage complex interactions among multiple entities. Specifically, multi-agent interactions introduce challenges such as exponential growth in problem complexity due to enlarged joint action spaces and extended planning horizons, partial observability stemming from decentralized information among agents, non-stationarity arising from concurrent agent learning processes, and difficulties associated with accurately assigning credit to individual contributions~\cite{zhang2021multi,Yuan2023OpenSurvey,albrecht2024multi}. Despite substantial advancements in single-agent embodied AI, research on embodied AI within multi-agent contexts remains relatively nascent. Current research typically adapts successful single-agent methods or employs established frameworks such as RL and LLMs. Recently, the development of specialized benchmarks tailored explicitly to embodied multi-agent scenarios has begun, aiming to support systematic progress in this evolving field. While extensive literature reviews have thoroughly explored related domains, including embodied AI~\cite{pfeifer2004embodied,liu2024aligning}, multi-agent reinforcement learning (MARL)~\cite{zhang2021multi,albrecht2024multi}, and multi-agent cooperation~\cite{Yuan2023OpenSurvey,oroojlooy2023review}, comprehensive surveys explicitly focusing on embodied multi-agent AI remain limited. For example, Literature~\cite{tarafdar2025embodied} systematically summarizes recent progress in embodied MARL, covering topics such as social learning, emergent communication, Sim2Real transfer, hierarchical methods, and safety considerations. Literature~\cite{wu2025generative} further extends this perspective by reviewing the integration of generative foundation models into embodied multi-agent systems (MAS), proposing a taxonomy of collaborative architectures, and discussing essential components like perception, planning, communication, and feedback mechanisms. Nevertheless, these surveys primarily address specific dimensions of multi-agent embodied AI, lacking a systematic and comprehensive overview of the entire field.

Recognizing the substantial potential of multi-agent embodied AI for addressing complex coordination tasks in real-world environments, this paper provides a systematic and comprehensive review of recent advances in this emerging research area. As shown in Figure~\ref{framwork}, we first introduce foundational concepts, including MAS, RL, and relevant methodologies. Next, we discuss embodied AI within single-agent contexts, clearly outlining core definitions, primary research directions, representative methods, and established evaluation benchmarks. Building on this foundation, we extend our discussion to multi-agent embodied AI, highlighting widely employed techniques and examining recent prominent benchmarks specifically designed for multi-agent scenarios. Finally, we summarize the primary contributions of this review, present insightful perspectives on the future development of multi-agent embodied AI, and aim to stimulate further research and innovation in this promising and rapidly evolving field.

\section{Preliminaries} \label{sec:background}
In this section, we present the core techniques that underpin Embodied AI, starting with a formal definition of Embodied AI itself. We also define the concept of MAS and explore a range of essential methods, including optimal control (OC), RL, MARL, hierarchical learning, IL, and generative models.

\subsection{Embodied AI} 
\begin{figure*}
\centering
\includegraphics[width=.98\columnwidth]{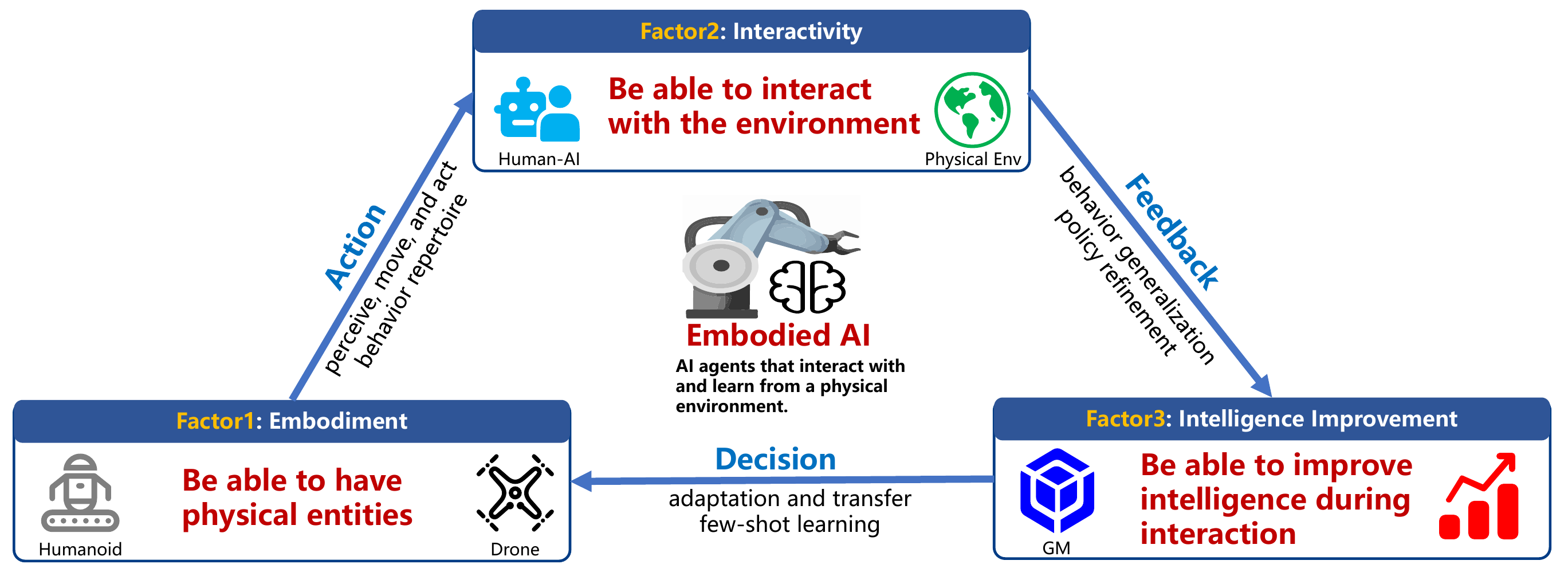}
\caption{An illustration of embodied AI.}
\label{emb_ai}
\end{figure*}

Embodied AI (Figure~\ref{emb_ai}) refers to a class of intelligent agents equipped with physical bodies that enable them to perceive, act upon, and adapt to the environments through continuous interaction~\cite{liu2024aligning}. The concept of embodied AI trace back to Alan Turing’s early proposition in the 1950s, suggesting that genuine intelligence must arise from sensory and motor experiences rather than purely symbolic computation~\cite{turing1950computing}. This notion was further formalized in the 1980s through embodied cognition theory, which posits that cognition is inherently shaped by the agent’s physical form and interaction with the world~\cite{brooks1991new}. In contrast to traditional AI paradigms that rely on abstract reasoning or passive learning from static datasets, embodied AI emphasizes real-world interaction as the foundation for learning and decision-making.

At a system level, embodied AI architectures are typically composed of three tightly integrated components: perception, cognition, and action. Firstly, agents use sensors to acquire real-time, context-dependent information from their surroundings. Then the sensory data is processed by cognitive modules that support reasoning, interpretation and planning. Finally, resulting decisions are translated into actions via actuators, which modify the environment and initiate new perceptual inputs. These processes form a continuous feedback loop known as the perception–cognition–action cycle~\cite{bridgeman2011embodied}, which enables embodied agents to adapt their behavior dynamically based on feedback. The paradigm of embodied AI mainly include three foundational attributes that govern how intelligence emerges and evolves:
\begin{itemize}
    \item \textbf{Embodiment} Embodied AI is rooted in physical agents equipped with the capabilities to perceive, move, and act in the real world. These agents come in diverse forms, including humanoid robots, quadrupeds, unmanned vehicles and drones. The physical body serves not only as the medium through which the agent interacts with its environment, but also as the structural foundation that constrains and enables its behavioral repertoire. The morphology, sensorimotor fidelity and actuation of the body jointly define the range and granularity of the agent’s interactions, shaping the scope of its situated intelligence. 
    \item \textbf{Interactivity} Built upon this physical grounding, embodied intelligence emerges through continuous, closed-loop interaction with the surrounding environment. Operating from a first-person perspective, the agent engages in a dynamic cycle of perception, decision-making, and action. Each behavior not only responds to environmental stimuli but also alters future sensory inputs, forming a feedback-rich loop that supports adaptive learning. Through this persistent engagement, the agent refines its policy, acquires task-specific competencies, and generalizes behaviors across varying contexts, allowing for robust, context-aware performance in real-world scenarios.
    \item \textbf{Intelligence Improvement} The development of embodied AI is characterized by a capacity for continual improvement in both cognition and behavior. This progression is increasingly enabled by the integration of large-scale multimodal models, which endow agents with semantic understanding, instruction following, and contextual reasoning. These models facilitate few-shot learning, in-context adaptation, and knowledge transfer across tasks. As the agent interacts with its environment, it incrementally aligns its perceptual inputs, decision processes, and physical actions, leading to both immediate task success and sustained growth in autonomy, adaptability, and generalization over time.
\end{itemize}

Recently, the advancements in generative models such as LLMs \cite{zhao2023survey} have further expanded the cognitive capabilities of embodied agents. Through leveraging their powerful reasoning and generalization abilities, LLMs enable embodied agents to understand language instructions, ground semantic knowledge in physical experiences, and perform zero-shot or few-shot adaptations. These developments have accelerated the deployment of embodied AI in various real-world domains such as robotics, autonomous driving, intelligent manufacturing and healthcare~\cite{ma2024survey}. Importantly, embodied AI is not merely the integration of powerful AI models with robotic platforms. It rather represents a co-evolutionary paradigm where intelligent algorithms which serve as brains, physical structures which serve as bodies and dynamic environments evolve jointly to support adaptive embodied intelligence. 

\subsection{Multi-Agent Systems} \label{mas}
\begin{figure*}
\centering
\includegraphics[width=.98\columnwidth]{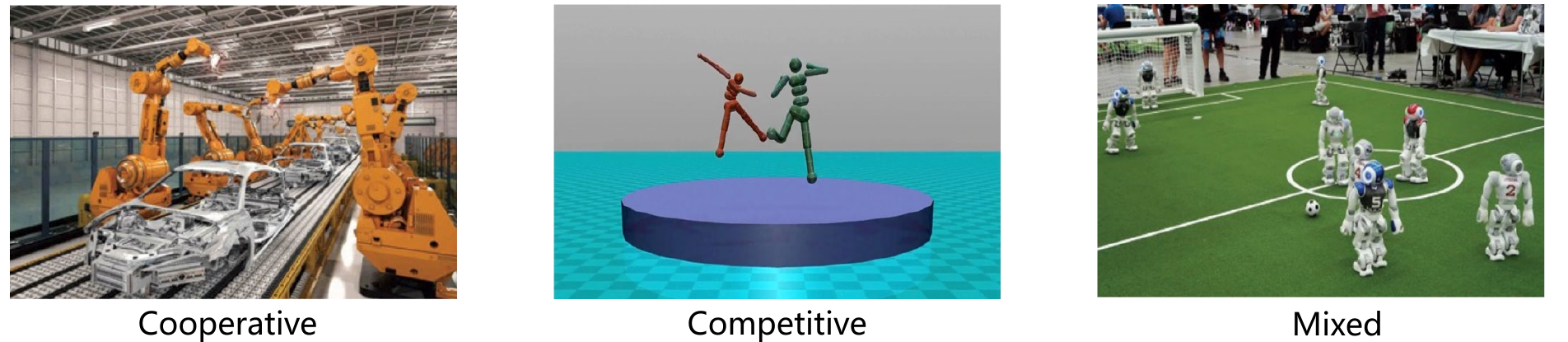}
\caption{Three common settings of MAS.}
\label{3mas}
\end{figure*}

MAS are composed of multiple autonomous agents, each capable of perceiving its environment, making independent decisions and executing actions accordingly~\cite{dorri2018multi}. In contrast to centralized control paradigms, MAS adopt a decentralized architecture in which agents interact locally while achieving global coordination. This decentralized design offers notable advantages in scalability, fault tolerance, and adaptability, particularly in dynamic, partially observable, or non-stationary environments. Core attributes of MAS include autonomy, decentralization, inter-agent communication, local information access, and dynamic adaptability. These features collectively empower MAS to tackle a broad spectrum of complex, high-dimensional tasks that demand parallel sensing, distributed planning, and real-time coordination, with prominent applications in domains such as robotics, autonomous driving, and smart infrastructure.

In recent years, MAS research has undergone a significant paradigm shift, fueled by the convergence of learning-based methods and advances in neural architectures. At the forefront of this transformation is MARL~\cite{albrecht2024multi}, which provides a powerful framework for enabling agents to learn complex behaviors through interaction. Techniques such as centralized training with decentralized execution (CTDE), parameter sharing, credit assignment, and opponent modeling have been widely adopted to address core challenges including non-stationarity, coordination, and partial observability. Complementing these advances, the integration of LLMs has opened up new capabilities for MAS. LLM-enabled agents can access vast pretrained knowledge, communicate via natural language, and engage in high-level reasoning and abstraction, capabilities that transcend the limitations of conventional policy-driven systems. The fusion of reinforcement learning and foundation models is thus reshaping the MAS landscape, paving the way for more generalizable, interpretable, and human-aligned agent architectures~\cite{guo2024large}. 

\subsection{Optimal Control} \label{subsec:OC}
\begin{figure*}
\centering
\includegraphics[width=.98\columnwidth]{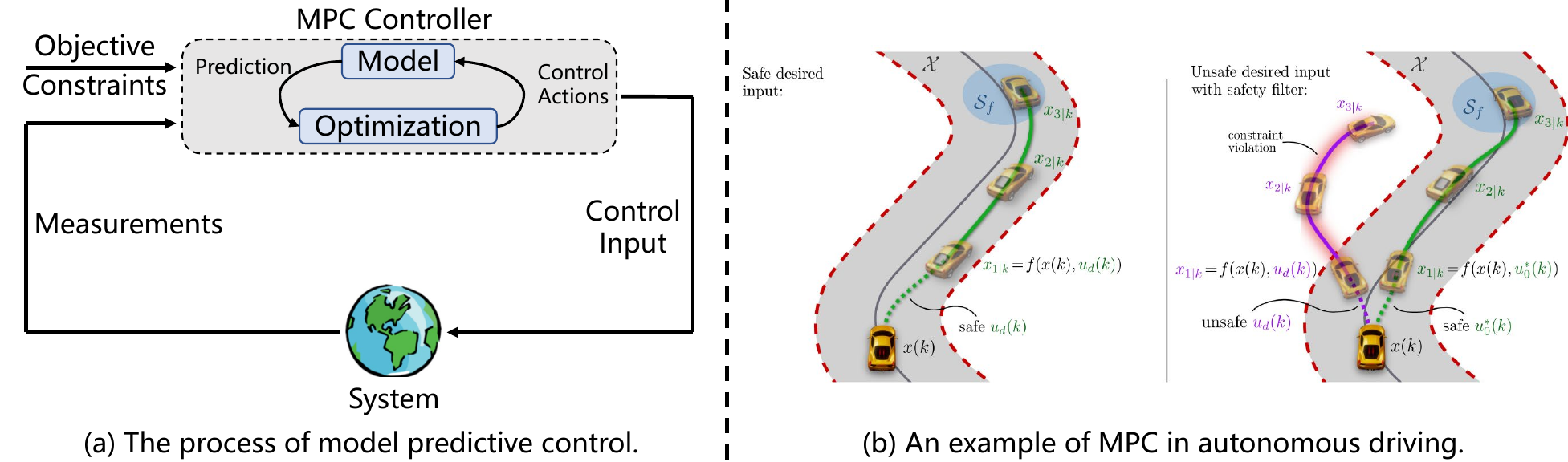}
\caption{An illustration of MPC.}
\label{mpc}
\end{figure*}
OC is a fundamental branch of modern control theory that focuses on designing control policies to optimize a specific performance criterion, such as time, energy consumption, or cost, while satisfying dynamic and control constraints of the system~\cite{ross2015primer}. Owing to its ability to achieve global optimality, handle multiple objectives and constraints simultaneously, and maintain robustness in complex systems, OC has found wide-ranging applications in fields such as aerospace engineering, industrial automation, economic management, and energy systems. 

Formally, a standard OC problem can be described as follows: Given a dynamic system governed by $\dot{x}(t)=f(x(t),u(t),t)$ and an initial state $x(t_0)=x_0$, the goal is to find the optimal control input $u(t)$ that minimizes the performance index:
\begin{align}
    J=\phi(x(t_f),t_f)+\int_{t_0}^{t_f}L(x(t),u(t),t)dt,
\end{align}
where $\phi$ is the terminal cost function and $L$ is the running cost function.

Common methods in OC include the calculus of variations, Pontryagin's maximum principle, dynamic programming, and model predictive control (MPC). The calculus of variations derives optimal trajectories using the Euler–Lagrange equations. PMP formulates necessary conditions for optimality by constructing a Hamiltonian function and defining both state and costate dynamics. DP addresses state-space optimization in complex systems via Bellman's principle of optimality. In contrast, MPC solves constrained optimization problems in real time using a receding horizon strategy.
Among them, MPC is the most widely adopted in industry. Its core idea is to solve a finite-horizon OC problem at each time step and implement only the first OC action, as shown in Figure~\ref{mpc}. Specifically, at time $t_k$, MPC solves the following online optimization problem:
\begin{align}
\min_{u_0,u_1,\dots,u_{N-1}} \sum_{i=0}^{N-1} L(x_i,u_i)+\phi(x_N),
\end{align}
subject to the system dynamics $x_{i+1}=f(x_i,u_i,i)$.
Due to its efficiency, real-time capability, and robustness, MPC has been widely applied in embodied AI systems with multiple agents, such as coordinated control of robot swarms and formation control of unmanned aerial vehicles (UAV).

\subsection{Reinforcement Learning} \label{subsec:RL}
\begin{figure*}
\centering
\includegraphics[width=.98\columnwidth]{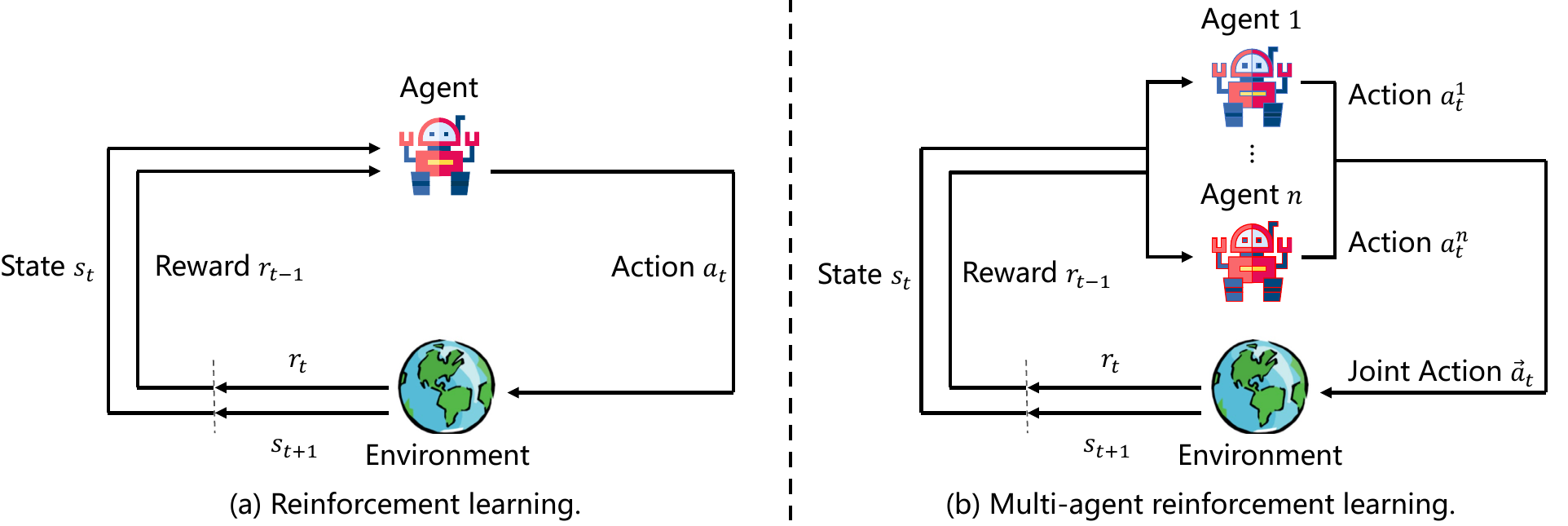}
\caption{An illustration of RL and MARL.}
\label{mdp}
\end{figure*}
Reinforcement Learning (RL) is a fundamental branch of machine learning that focuses on enabling agents to make sequential decisions by interacting with dynamic environments~\cite{Sutton1998}. Unlike supervised learning, which requires labeled data, or unsupervised learning, which uncovers hidden patterns in unlabeled data, RL is grounded in a trial-and-error process. Through continuous interaction, agents learn to select actions that maximize long-term cumulative rewards, using feedback from the environment to improve their behavior over time. Owing to its capacity for learning from interaction, RL has achieved impressive results across a broad range of domains. These include strategic game playing~\cite{lample2017playing, souchleris2023reinforcement}, robotic control~\cite{singh2022reinforcement}, industrial process optimization~\cite{nian2020review}, personalized healthcare~\cite{yu2021reinforcement}, autonomous driving~\cite{kiran2021deep,Li2024continuous}, and even alignment and instruction-following in LLMs~\cite{ouyang2022training, havrilla2024teaching}.

At its core, an RL problem is commonly formalized as a Markov decision process (MDP), depicted in Figure~\ref{mdp}(a). An MDP is defined by a tuple $\mathcal{M} = \langle \mathcal{S}, \mathcal{A}, P, R, \gamma \rangle$, where $\mathcal{S}$ and $\mathcal{A}$ represent the sets of states and actions, $P(s'|s,a)$ is the transition probability function, $R(s,a)$ denotes the reward function, and $\gamma \in [0,1]$ is a discount factor that balances short- and long-term rewards. The objective is to learn a policy $\pi(a|s)$ that maximizes the following expected cumulative return
\begin{equation}
\mathbb{E}_\pi\Big[\sum_{t = 0}^\infty \gamma^t R(s_t,a_t)\mid s_0 \Big],
\label{eq:RLtotalreward}
\end{equation}
where the expectation $\mathbb{E}_\pi[\cdot]$ is taken over trajectories generated by the policy $\pi$ and the transition dynamics $P$. While single-agent RL provides a solid foundation for autonomous decision-making, many real-world scenarios involve multiple agents acting and learning simultaneously. This motivates the extension of RL into MARL, where coordination or competition among agents becomes essential. Specifically, in fully cooperative tasks, MARL problems are often modeled as the decentralized partially observable Markov decision process (Dec-POMDP) framework as shown in Figure~\ref{mdp}(b). A Dec-POMDP extends the MDP by introducing multiple agents and partial observability. Formally, it is defined as a tuple $\mathcal{M} = \langle I, \mathcal{S}, {\mathcal{A}_i}, P, R, {\Omega_i}, O, n, \gamma \rangle$, where $I$ is the set of $n$ agents. Each agent $i$ selects actions from its own action space $\mathcal{A}_i$, based solely on local observations from $\Omega_i$. The global state transitions according to the joint action $\bm{a}$ via $P(s'|s, \bm{a})$, and all agents receive a shared reward $R(s, \bm{a})$. Observations are generated through the function $O(s', \bm{a}, \bm{o})$, providing each agent with private, potentially incomplete information. The central challenge lies in learning a set of decentralized policies $[\pi_1, \ldots, \pi_n]$, where each agent must act appropriately using local information, yet their joint behavior maximizes the team’s cumulative reward.

A wide range of algorithmic strategies has been developed to address both single-agent and multi-agent RL problems \cite{Cao2025causal,Zhou2024MultiUAVLearning,Kang2024autonomous}. Among these, two principal categories stand out: value-based methods and policy-based methods. Value-based methods, such as the deep Q-network (DQN) \cite{mnih2013playing}, focus on estimating action-value functions, enabling agents to act greedily with respect to predicted future rewards. In contrast, policy-based methods, such as DDPG \cite{lillicrap2015continuous} and PPO~\cite{Schulman2017PPO}, directly optimize parameterized policies. These methods often incorporate value functions to guide learning and are especially suited to environments with high-dimensional or continuous action spaces. Both paradigms have been extended to the multi-agent settings through distinct architectural innovations. Value decomposition methods, such as QMIX~\cite{rashid2020monotonic}, learn a global value function that can be decomposed into individual agent utilities. This structure supports decentralized execution while enabling centralized training. On the other hand, policy-based MARL approaches, including MADDPG~\cite{lowe2017multi} and MAPPO~\cite{yu2022surprising}, adopt the CTDE framework. These methods employ a shared global critic during training to provide stabilized learning signals and support effective credit assignment among agents, thereby enhancing coordination in cooperative tasks.

\subsection{Hierarchical Learning}

\begin{figure*}
\centering
\includegraphics[width=.98\columnwidth]{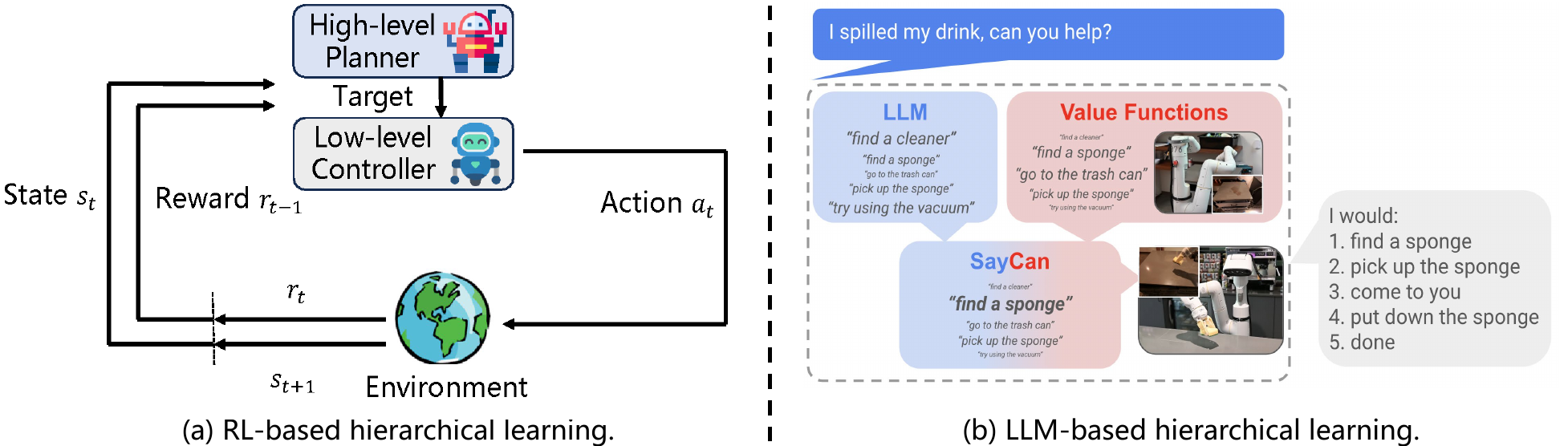}
\caption{An illustration of RL-based (high-level) hierarchical learning and LLM-based (high-level) hierarchical learning~\cite{icher2023PaLMPlanning}.}
\label{hrl}
\end{figure*}

Hierarchical learning is a learning paradigm that organizes the learning process into multiple levels of abstraction, with higher levels responsible for setting abstract goals or intentions, and lower levels focusing on executing more concrete, fine-grained subtasks (cf. Figure~\ref{hrl}). This layered structure enables models to operate across varying levels of granularity, thereby improving efficiency and scalability in solving complex tasks by breaking them down into simpler, manageable components~\cite{pateria2021hierarchical}.

The core process of hierarchical learning typically involves two stages: low-level skill learning and high-level policy learning. The low-level skill learning is aimed at mastering basic sub-tasks and is often implemented using classic controllers such as MPC~\cite{Song2022PolicySearchMPC1, Feng2024PolicySearchMPC3,Duan2025HierarchicalMPC}, or through end-to-end RL~\cite{yang2021hierarchical, li2022hierarchical}. In contrast, the high-level policy learning is responsible for orchestrating the learned low-level skills to accomplish more complex goals. This is commonly achieved through RL or via planning with LLMs. 
In RL–based high-level policy learning, the action space is defined as the set of learned low-level policies. A parameterized high-level policy is then trained using reward signals from the environment to select and sequence these low-level skills effectively~\cite{yang2021hierarchical,li2022hierarchical}. In contrast, LLM-based high-level policy learning typically involves providing the task objective and the available set of low-level skills as input to the LLM, which then directly generates a structured plan by composing and invoking the appropriate low-level skills to accomplish the task~\cite{icher2023PaLMPlanning}.

\subsection{Imitation Learning}

IL, also referred to as learning from demonstrations, is a machine learning paradigm in which an agent acquires task-solving capabilities by observing and mimicking expert behavior. In contrast to RL, which requires the manual design of a reward function to guide learning, IL leverages expert demonstrations as direct supervision. This distinction makes IL particularly advantageous in complex, high-dimensional environments where reward specification is ambiguous, costly, or even infeasible. Over the years, three approaches have emerged as the most prominent within IL: behavior cloning (BC), inverse reinforcement learning (IRL), and generative adversarial imitation learning (GAIL)~\cite{Zare2024ImitationSurvey}.

\begin{figure*}
\centering
\includegraphics[width=.98\columnwidth]{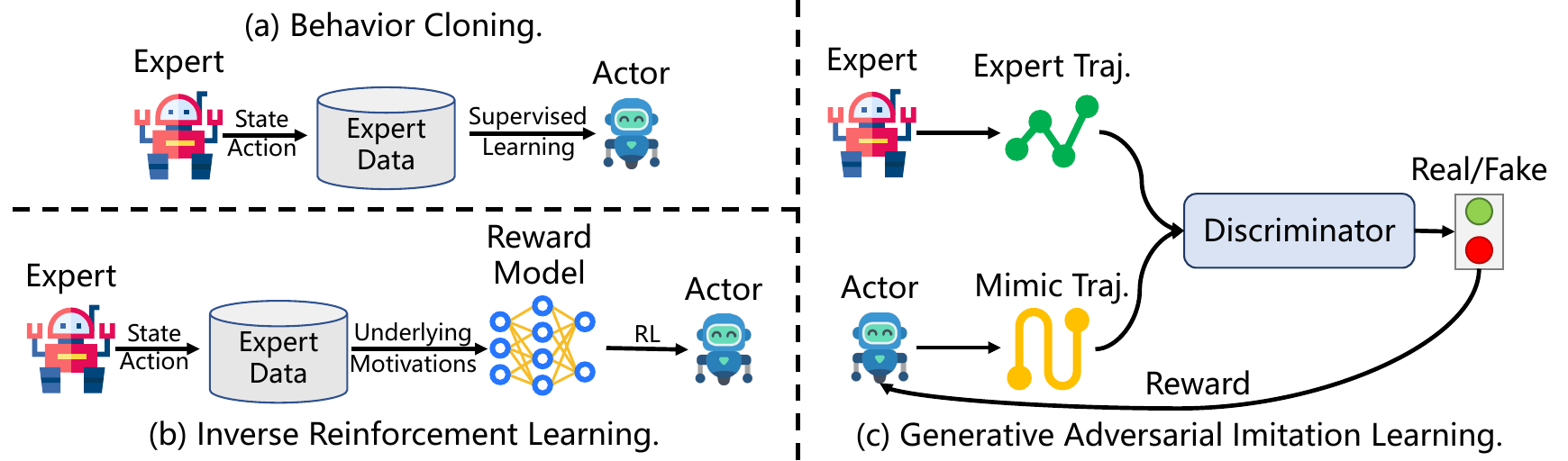}
\caption{An illustration of different IL methods.}
\label{il}
\end{figure*}

Among these, BC is the most straightforward and widely adopted technique. It casts the imitation task as a supervised learning problem, aiming to directly map observed states to corresponding expert actions. As illustrated in Figure~\ref{il}(a), the agent learns from a dataset of expert-generated state-action pairs $\mathcal{D} = \{(s_i, a_i)\}_{i=1}^N$, optimizing the following objective:
\begin{equation}
\label{eq:BehaviorCloning}
\min_{\theta} \mathbb{E}_{(s,a)\sim\mathcal{D}}[-\log\pi_{\theta}(a|s)],
\end{equation}
where $\theta$ denotes the parameters of the policy $\pi_\theta$, this loss encourages the learned policy to produce actions that closely match those of the expert when exposed to similar states. While BC is computationally efficient and simple to implement, it suffers from error accumulation due to covariate shift, especially in long-horizon tasks where the agent visits states not seen during training.

To address some of BC’s limitations, IRL adopts a fundamentally different approach by aiming to infer the expert’s underlying reward function rather than replicating their actions directly. Once the reward function is recovered, standard RL methods can be used to derive the optimal policy. This two-step process, illustrated in Figure~\ref{il}(b), enables improved generalization and policy interpretability. Formally, a representative formulation is the maximum entropy IRL objective
\begin{align}
\label{eq:IRL}
\max_{R} \mathbb{E}_{\tau \sim \mathcal{D}}[R(\tau)] - \log Z(R),
\end{align}
where $\tau = (s_0, a_0, \dots, s_T, a_T)$, $R(\tau) = \sum_{t=0}^T R(s_t, a_t)$, and $Z(R) = \int e^{R(\tau)} d\tau$ is the partition function that ensures a normalized trajectory distribution. By explicitly modeling the expert's intent, IRL provides more robust and explainable policies. However, this benefit comes at the cost of additional computational overhead due to the requirement for repeated RL iterations during reward learning.

Different from both BC and IRL, GAIL~\cite{ho2016generative} introduces an adversarial training framework that eliminates the need for explicit reward specification. Inspired by generative adversarial networks (GANs)~\cite{goodfellow2014generative}, GAIL frames imitation as a minimax game between two components: a generator (the policy to be learned) and a discriminator that distinguishes expert trajectories from those generated by the policy. As depicted in Figure~\ref{il}(c), the optimization objective is
\begin{align}
\label{eq:gail}
\min_{\pi} \max_{D} \mathbb{E}_{(s,a) \sim \mathcal{D}}[\log D(s,a)] + \mathbb{E}_{(s,a) \sim \mathcal{D}_{\pi}}[\log(1 - D(s,a))],
\end{align}
where $\mathcal{D}_{\pi}$ represents trajectories generated by the current policy. Through this adversarial setup, GAIL implicitly learns a reward signal that encourages the agent to behave indistinguishably from the expert. Compared to IRL, GAIL streamlines the training process by avoiding explicit reward modeling, resulting in better sample efficiency. However, this advantage comes at the expense of reduced interpretability, as the learned reward is never explicitly extracted. In summary, BC, IRL, and GAIL represent complementary paradigms within the broader IL framework. BC provides a simple and effective baseline, IRL offers interpretability and generalization via reward recovery, and GAIL combines the strengths of both by enabling end-to-end imitation through adversarial learning. The choice among them often depends on the trade-offs between sample efficiency, generalization capability, computational complexity, and interpretability required by the target application.

\subsection{Generative Models}

\begin{figure*}
\centering
\includegraphics[width=.98\columnwidth]{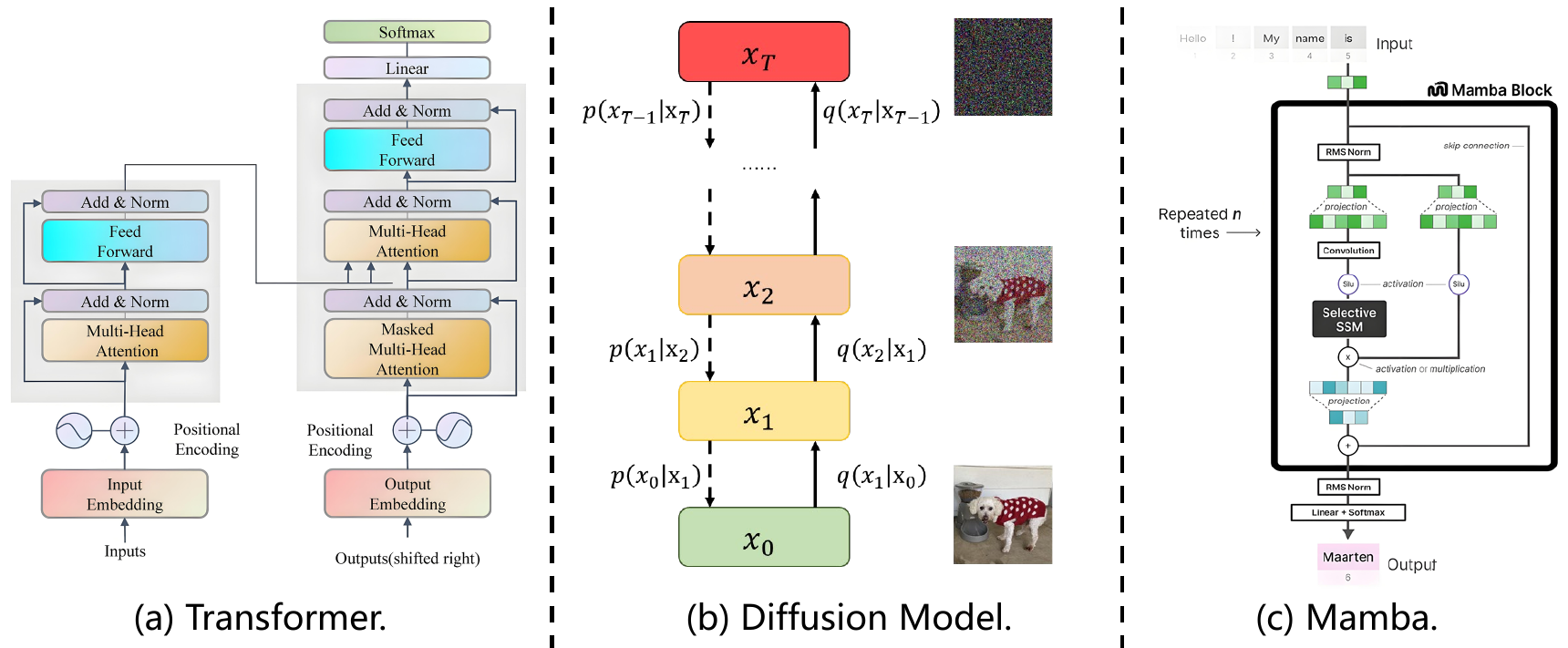}
\caption{An illustration of different generative model structures.}
\label{gen_m}
\end{figure*}

Generative models constitute a foundational paradigm in machine learning, with the central objective of capturing the underlying distribution of training data to generate novel samples that exhibit similar characteristics. These models have found widespread applications across domains such as vision, language and multi-modal learning. In recent years, the emergence of large-scale generative models, such as LLMs and VLMs, has significantly advanced the field. Their success is largely attributed to strong generalization capabilities, the availability of massive datasets and scalable architectures. At the core of these models lie several key architectural frameworks, including Transformers~\cite{Vaswani2017Transformer}, diffusion models~\cite{Ho2020Diffusion} and more recently, state space models (SSMs) such as Mamba~\cite{Gu2024Mamba}.
Among these, the Transformer architecture has played a pivotal role in revolutionizing sequence modeling. Originally proposed for machine translation, Transformers eliminate the need for recurrence or convolution by introducing an attention-based mechanism that enables each element in a sequence to directly attend to every other element (cf. Figure~\ref{gen_m}(a)). This design facilitates efficient parallel computation and allows the model to capture global contextual dependencies. The core attention mechanism is mathematically defined as:
\begin{align}
\text{Attention}(Q,K,V) = \text{softmax}\left(\frac{QK^T}{\sqrt{d_k}}\right)V,
\end{align}
where $Q$, $K$, and $V$ represent the queries, keys, and values, respectively, and $d_k$ is the dimensionality of the key vectors, acting as a scaling factor to stabilize gradients. In the case of self-attention, $Q$, $K$, and $V$ are computed from the same input sequence, enabling each token to contextualize itself with respect to all others. This mechanism is fundamental to the success of Transformers in a wide range of natural language processing and vision tasks.

In contrast to the discrete attention mechanisms of Transformers, diffusion models offer a probabilistic, noise-driven framework for generation. These models operate through a two-phase process: a forward diffusion stage that incrementally corrupts data with noise, and a learned reverse process that reconstructs the data from noisy inputs (cf. Figure~\ref{gen_m}(b)). The forward diffusion process forms a Markov chain in which Gaussian noise is added at each timestep:
\begin{align}
q(x_t|x_{t-1}) = \mathcal{N}(x_t; \sqrt{1 - \beta_t}x_{t-1}, \beta_t I),
\end{align}
where $\beta_t$ denotes a predefined variance schedule and $I$ is the identity matrix. The reverse process, which the model learns to approximate, also follows a Gaussian distribution:
\begin{align}
p(x_{t-1}|x_t) = \mathcal{N}(x_{t-1}; \mu_{\theta}(x_t, t), \Sigma_{\theta}(x_t, t)),
\end{align}
where $\mu_{\theta}$ and $\Sigma_{\theta}$ are parameterized by neural networks, typically based on U-Net architectures~\cite{ronneberger2015u}. By iteratively denoising from pure Gaussian noise, diffusion models can generate high-fidelity outputs, making them highly effective for image generation, audio synthesis, and other structured data domains.

While Transformers and diffusion models have achieved remarkable success, they both encounter limitations when scaling to long sequences or complex generative processes. To address these challenges, the Mamba architecture introduces a novel approach based on continuous-time SSMs, offering linear-time complexity and improved efficiency in long-sequence processing (cf. Figure~\ref{gen_m}(c)). Mamba represents sequential data using the following differential equations:
\begin{align}
\dot{h}(t) = A(t)h(t) + B(t)x(t), \
y(t) = C(t)h(t) + D(t)x(t),
\end{align}
where $h(t)$ is the hidden state, $x(t)$ and $y(t)$ denote the input and output signals, respectively, and $A(t)$, $B(t)$, $C(t)$, and $D(t)$ are learnable, time-dependent system parameters. In most practical scenarios, $D(t)$ is set to zero, introducing a skip connection that improves gradient flow. By discretizing the continuous system and introducing a selective scanning mechanism, Mamba achieves superior inference efficiency and scalability. These properties make it a compelling alternative for tasks involving long-range dependencies and real-time computation. 
Taken together, Transformers, diffusion models, and SSMs like Mamba represent three major architectural directions in generative modeling. Each of these paradigms provides unique advantages in terms of expressiveness, efficiency, and applicability, and understanding their design principles is critical for the development of next-generation generative systems.

\section{Single-Agent Embodied AI} \label{sec:SAembodied} 
\begin{table*}
\centering
\caption{Methods for building embodied AI in single-agent settings, where the abbreviation GM refers to generative model. }
\small
\label{tbl:SAmethods}
\resizebox{\textwidth}{!}{
\begin{tabular}{l|c|ccc|cccc|cc|ccccc|ccc}
\toprule
& \multicolumn{1}{c|}{\textbf{Year}} 
& \multicolumn{3}{c|}{\textbf{Category}} 
& \multicolumn{4}{c|}{\textbf{Input}} 
& \multicolumn{2}{c|}{\textbf{View}} 
& \multicolumn{5}{c|}{\textbf{Agent Type}} 
& \multicolumn{3}{c}{\textbf{Implement}} 
\\
& \rotatebox{90}{\quad}
& \rotatebox{90}{Perception} & \rotatebox{90}{Planning} & \rotatebox{90}{Control}
& \rotatebox{90}{Image} & \rotatebox{90}{Pointcloud} & \rotatebox{90}{Language} & \rotatebox{90}{Proprioception}
& \rotatebox{90}{Global} & \rotatebox{90}{Local}
& \rotatebox{90}{Humanoid} & \rotatebox{90}{Quadruped/Bipedal} & \rotatebox{90}{UAV/UGV/USV} & \rotatebox{90}{Manipulator} & \rotatebox{90}{No certain type} 
& \rotatebox{90}{Solver-based} & \rotatebox{90}{NN-based} & \rotatebox{90}{GM-based} 
\\
\midrule
RRT \cite{Lavalle1998RRT} & 1998 &  & \checkmark &  &  &  &  & \checkmark & \checkmark &  &  &  &  &  & \checkmark & \checkmark &  &  \\
MINCO \cite{Wang2022MINCO} & 2022 &  & \checkmark & \checkmark &  &  &  & \checkmark & \checkmark &  &  &  & \checkmark &  &  & \checkmark &  &  \\
High-MPC \cite{Song2022PolicySearchMPC1} & 2022 &  & \checkmark & \checkmark &  &  &  & \checkmark & \checkmark &  &  &  & \checkmark &  &  & \checkmark & \checkmark &  \\
SteveEye \cite{Zheng2023LLMPerception1} & 2023 & \checkmark & \checkmark &  & \checkmark &  & \checkmark &  &  & \checkmark &  &  &  &  & \checkmark &  &  & \checkmark \\
SayCan \cite{icher2023PaLMPlanning} & 2023 & \checkmark & \checkmark & \checkmark & \checkmark &  & \checkmark &  &  & \checkmark &  &  & \checkmark & \checkmark &  &  & \checkmark & \checkmark \\
VisionRacing \cite{Fu2023EndtoEndUAV} & 2023 & \checkmark &  & \checkmark & \checkmark &  &  & \checkmark &  & \checkmark &  &  & \checkmark &  &  &  & \checkmark &  \\
GNFactor \cite{Ze2023BCMultitask} & 2023 & \checkmark & \checkmark & \checkmark & \checkmark &  & \checkmark & \checkmark &  & \checkmark &  &  &  & \checkmark &  &  & \checkmark &  \\
PaLM-E \cite{Driess2023EndtoEndLLM7} & 2023 & \checkmark & \checkmark &  & \checkmark &  & \checkmark & \checkmark &  & \checkmark &  &  & \checkmark & \checkmark &  &  &  & \checkmark \\
InnerMonologue \cite{Huang2023LLMReflection} & 2023 & \checkmark & \checkmark &  & \checkmark &  & \checkmark &  &  & \checkmark &  &  & \checkmark & \checkmark &  &  &  & \checkmark \\
Infer \& Adapt \cite{Wu2024IRL4} & 2024 &  &  & \checkmark &  &  &  & \checkmark &  & \checkmark &  & \checkmark &  &  &  &  & \checkmark &  \\
RT-H \cite{Belkhale2024EndtoEndLLM11} & 2024 & \checkmark & \checkmark & \checkmark & \checkmark &  & \checkmark &  &  & \checkmark &  &  & \checkmark & \checkmark &  &  &  & \checkmark \\
OpenVLA \cite{Kim2024EndtoEndLLM8} & 2024 & \checkmark & \checkmark & \checkmark & \checkmark &  & \checkmark &  &  & \checkmark &  &  & \checkmark & \checkmark &  &  &  & \checkmark \\
Octo \cite{Mees2024EndtoEndLLM9} & 2024 & \checkmark & \checkmark & \checkmark & \checkmark &  & \checkmark & \checkmark &  & \checkmark &  &  &  &  & \checkmark &  & \checkmark & \checkmark \\
$\pi_{0.5}$ \cite{Intelligence2025EndtoEndLLM10} & 2025 & \checkmark & \checkmark & \checkmark & \checkmark &  & \checkmark & \checkmark &  & \checkmark &  &  & \checkmark & \checkmark &  &  &  & \checkmark \\
\bottomrule
\end{tabular}
}
\end{table*}

The mainstream research in Embodied AI predominantly focuses on the single-agent settings, where an individual agent operates in a static environment to complete various tasks. In this context, existing approaches can be broadly categorized into classic methods and modern learning-based techniques. Early efforts primarily relied on knowledge-driven, task-specific planning and control strategies \cite{Li2023Control,Liu2023Control,Wang2023Control}. However, as agents are increasingly deployed in dynamic and unstructured environments, manually designing bespoke planners and controllers for each scenario has become infeasible. 
Thus, learning-based methods with inherent adaptability and strong generalization across diverse tasks have been the prevailing paradigm for instilling embodied agents with intelligence in complex, dynamic environments \cite{10204557,10100655,10611494}. More recently, advances in generative models have further enhanced the potential of embodied agents, enabling them to exhibit more adaptive and broadly applicable behaviors.

\subsection{Classic Control and Planning} \label{subsec:SAcontrol} 

To accomplish long-horizon tasks in the physical world, embodied agents must effectively plan and control their motions, generating trajectories grounded in both perceptual and task-specific inputs. 
To this end, a rich body of classic planning methods has been developed, which can be broadly grouped into constraint‐based, sampling‐based, and optimization‐based approaches~\cite{Zhao2024PlanningSurvey}.
Constraint-based methods encode task objectives and environmental conditions as logical constraints, translating the planning domain into symbolic representations, and employ constraint-solving techniques, such as symbolic search, to identify feasible solutions~\cite{Lozano2014ConstrainedPlanning1,Neil2018ConstrainedPlanning2}. However, these methods typically focus on feasibility rather than solution quality, often neglecting optimality. Additionally, when perceptual inputs are high-dimensional or structurally complex, the difficulty of constraint solving increases substantially. Sampling-based methods address these limitations by incrementally exploring the feasible solution space through random sampling techniques, such as rapidly-exploring random trees (RRT) and its variants, progressively building a tree or graph structure to discover viable motion trajectories~\cite{Lavalle1998RRT,Yu2022IRRT,OrtizIDBRRT}.

\begin{figure*}
\centering
\includegraphics[width=.98\columnwidth]{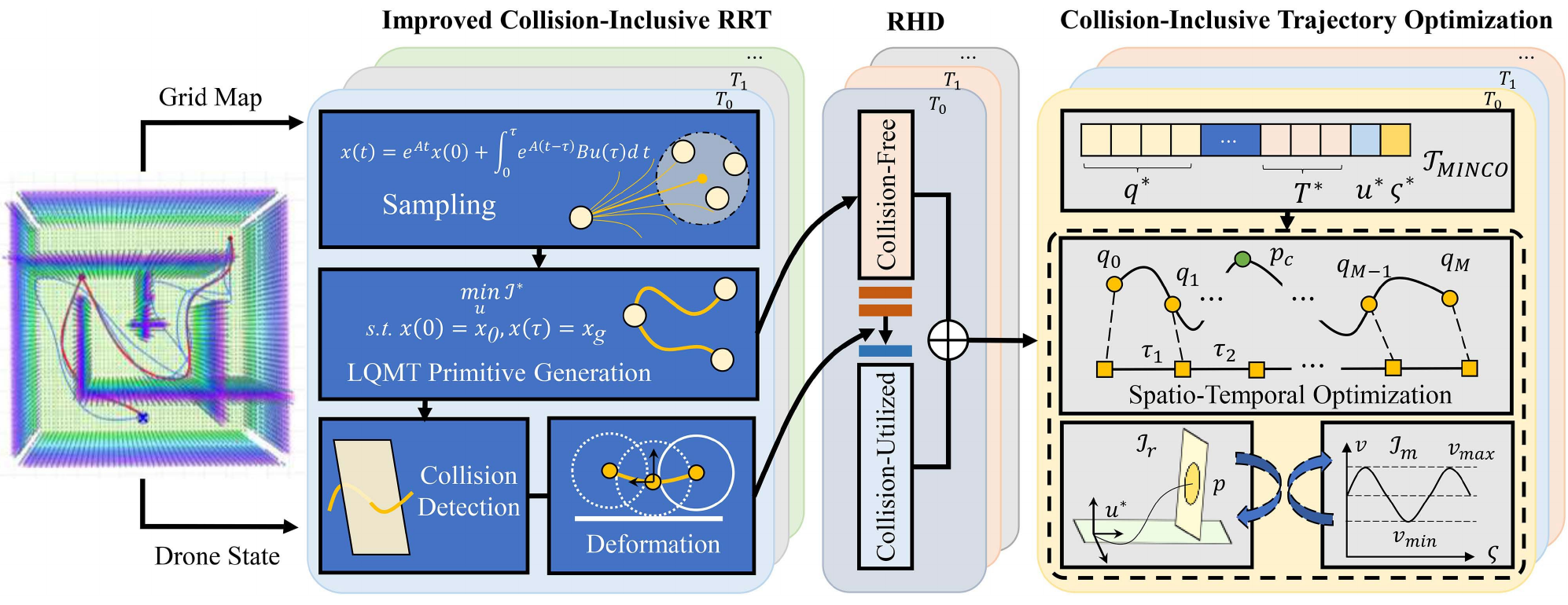}
\caption{An illustration of control-based motion planning~\cite{10720609}.}
\label{sa_control}
\end{figure*}

To further optimize motion planning within the feasible domain, optimization-based methods explicitly model task objectives and performance metrics as objective functions, while representing feasibility conditions as hard constraints. Advanced optimization techniques are then utilized to search for optimal solutions within the constrained solution space. Representative optimization-based methods include polynomial trajectory planning~\cite{Mellinger2011PolynominalPlanning1}, MPC \cite{kulic2024mpcc++} and OC \cite{Philipp2021OptimalControl1}. Among these, polynomial trajectory planning generally exhibits relatively low computational cost, ensuring real-time responsiveness and smooth trajectory generation, making it particularly suitable for autonomous navigation tasks in complex scenarios~\cite{yang2024trace,qin2024time,10700666,wang2025miner,fork2023euclidean}. In contrast, MPC and OC approaches are better suited for highly constrained scenarios requiring time-optimal maneuvers, such as collision avoidance during UAV operations~\cite{Philipp2021OptimalControl1,10803033,Zhou2023OptimalControl3,10720609}. For example, as illustrated in Figure~\ref{sa_control}, the work~\cite{10720609} introduces an aggressive collision-inclusive motion planning framework that strategically utilizes collisions to improve navigation efficiency in constrained environments. This method integrates an enhanced collision-inclusive rapidly-exploring random tree (ICRRT) for generating deformable motion primitives, alongside a trajectory optimization approach guided by a receding horizon decision strategy for autonomous collision avoidance. Experimental results demonstrate a substantial reduction in task completion time compared to conventional collision-free methods, showing the effectiveness of this approach and  underscoring its considerable potential for practical applications.

\subsection{Learning based Methods} \label{subsec:SARL}

Classic control and planning methods have long been employed for constrained real-time decision-making due to their ability to produce high-precision solutions. 
However, these approaches are often computationally intensive, limiting scalability and responsiveness, especially in high-dimensional, nonlinear or non-stationary systems. Moreover, they typically exhibit poor generalization to unseen scenarios or dynamic environments. 
To overcome these limitations, learning-based decision-making paradigms have gained increasing attention, aiming to deliver real-time performance with improved robustness and generalization by learning directly from interactive data \cite{Wang2025one}. Among these, end-to-end RL has emerged as a prominent approach as it enables direct policy optimization through interaction with the environment. Nonetheless, in tasks with complex objectives or large state-action spaces, end-to-end RL suffers from low sample efficiency and prolonged training times due to the expansive exploration space. To alleviate this issue, hierarchical RL has been introduced, which decomposes overall tasks into a sequence of sub-tasks, thereby improving learning efficiency and scalability. Despite these advances, both end-to-end and hierarchical RL approaches fundamentally depend on the availability of well-designed reward functions, which are often difficult or impractical to define in real-world settings. In such scenarios, IL provides a compelling alternative by enabling agents to acquire effective policies directly from expert demonstrations, circumventing the need for explicit reward engineering and facilitating faster deployment in practice.
\begin{figure*}
\centering
\includegraphics[width=.98\columnwidth]{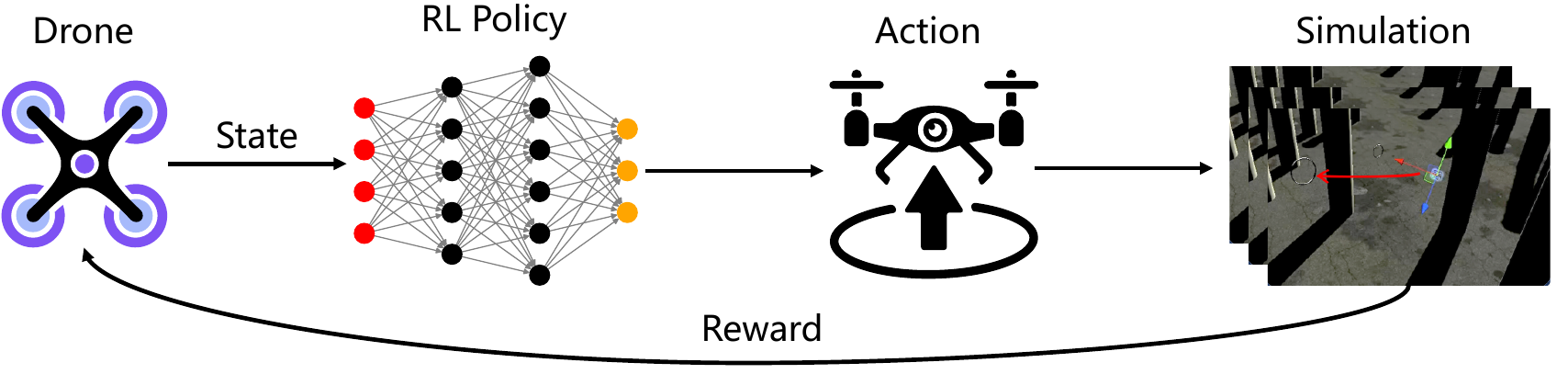}
\caption{An illustration of end-to-end RL methods, where the learned policy $\pi$ directly interacts with the environment~\cite{Xiao2025EndtoEndUAV}.}
\label{fig:endtoendrl}
\end{figure*}

\paragraph{End-to-end learning.} 
In sophisticated environments, OC problems often exhibit non-convex characteristics, making it challenging for classic planning and control methods to identify feasible solutions within a reasonable computational time. A natural approach to addressing this issue is to optimize the entire system in an end-to-end manner, employing RL techniques to directly map perceptual information to action decisions through neural networks. These methods optimize the policy by maximizing the expected cumulative rewards as defined in \eqref{eq:RLtotalreward}, ensuring stable and high-frequency control outputs (cf. Figure~\ref{fig:endtoendrl}). Consequently, end-to-end learning is widely recognized as an ultimate pathway toward embodied AI, as it removes the necessity for manual decomposition, alignment, and tuning of individual modules, enabling agents to directly translate raw sensory inputs into effective actions. Indeed, end-to-end RL has achieved impressive performance on diverse robotic platforms equipped with multi-modal inputs, including UAVs~\cite{Xiao2025EndtoEndUAV,Fu2023EndtoEndUAV}, autonomous vehicles~\cite{Gao2025EndtoEndDriving}, manipulators~\cite{Geng2023EndtoEndManipulation} and legged robots~\cite{Bohlinger2024EndtoEndQuadruped,He2024EndtoEndQuadruped1}.

\paragraph{Hierarchical learning.} 
\begin{figure*}
\centering
\includegraphics[width=.98\columnwidth]{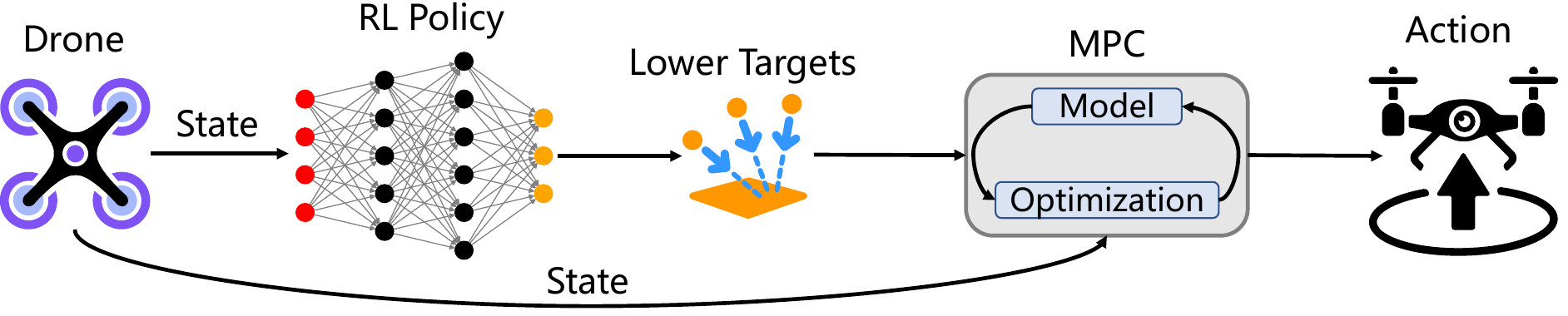}
\caption{An illustration of hierarchical learning using the example of embodied drone control.}
\label{sa_hl}
\end{figure*}
When dealing with tasks characterized by complex objectives or large state-action spaces, end-to-end RL often faces challenges related to low sample efficiency and prolonged training times due to expansive exploration spaces. To address this issue, researchers frequently adopt hierarchical frameworks integrating RL with classic control methods. In such frameworks, RL typically manages high-level planning or policy generation, while robust low-level controllers such as MPC or proportional integral derivative (PID) controllers execute precise and reliable action commands~\cite{Song2022PolicySearchMPC1,Feng2024PolicySearchMPC3,Hong2024HierarchicalLearning1}. This hierarchical approach enables embodied agents to effectively tackle challenging unstructured problems. For example, guiding a UAV through a dynamically swinging gate in minimal time is particularly difficult for classic methods alone (Figure~\ref{sa_hl}), as the gate's position evolves overtime, making it impractical to define a fixed optimization objective. Instead, RL can learn policies that determine intermediate target states and planning horizons for MPC, using reward signals shaped by traversal success and timing to generate efficient trajectories~\cite{Song2022PolicySearchMPC1}. Furthermore, when gate dynamics are unknown, a learnable mixing coefficient can dynamically adjust the control policy, enabling smooth transitions between pursuit and passage behaviors based on real-time state information, thereby enhancing trajectory continuity~\cite{Feng2024PolicySearchMPC3}. Additionally, carefully structured hierarchical designs can naturally extend this framework to support multi-agent collaborative tasks in complex embodied scenarios~\cite{Hong2024HierarchicalLearning1}.

\paragraph{Learning from demonstrations.} 
\begin{figure*}
\centering
\includegraphics[width=.95\columnwidth]{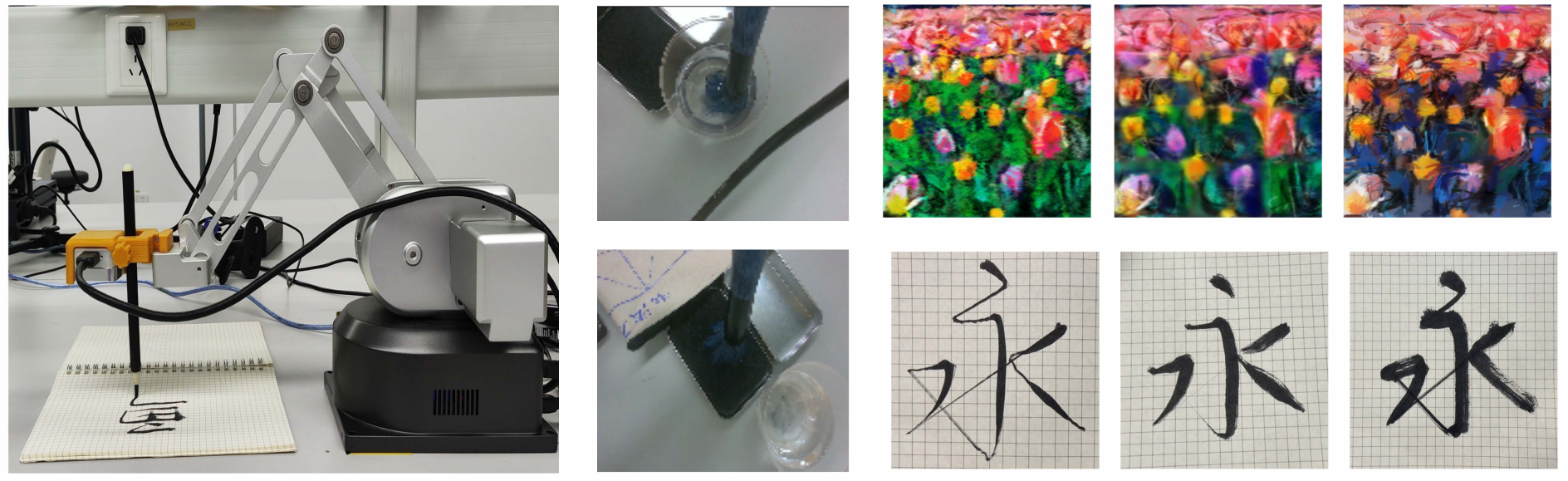}
\caption{An illustration of the writing and painting skills learned from demonstrations~\cite{Jia2024BCDraw}.}
\label{fig:learningfromdem}
\end{figure*}
Unlike OC methods, learning-based approaches such as RL require extensive data collection through interaction with the environment in a trial-and-error manner, incurring high costs in real-world embodied settings. Furthermore, RL often relies heavily on meticulously designed objective functions or reward signals, which become challenging to specify accurately in complex and unstructured environments due to the necessity for flexibility and adaptability. In these scenarios, learning from expert demonstrations, commonly referred to as IL, presents a more practical alternative. 
For example, \cite{Jia2024BCDraw} successfully transfers a painting policy from a simulator to a real-world robotic arm using BC, demonstrating high-level painting and calligraphy skills as shown in Figure~\ref{fig:learningfromdem}. Beyond this, a range of more advanced IL techniques have been introduced to further enhance embodied intelligence with improved generalization and robustness, including interactive IL~\cite{Oh2024InteractiveImitationLearning2}, constrained IL~\cite{Shi2023ConstrainedIL}, and IRL~\cite{zakka2022xirl}.

\subsection{Generative Model Based Methods} \label{subsec:SAWM} 
\begin{figure*}
\centering
\includegraphics[width=.95\columnwidth]{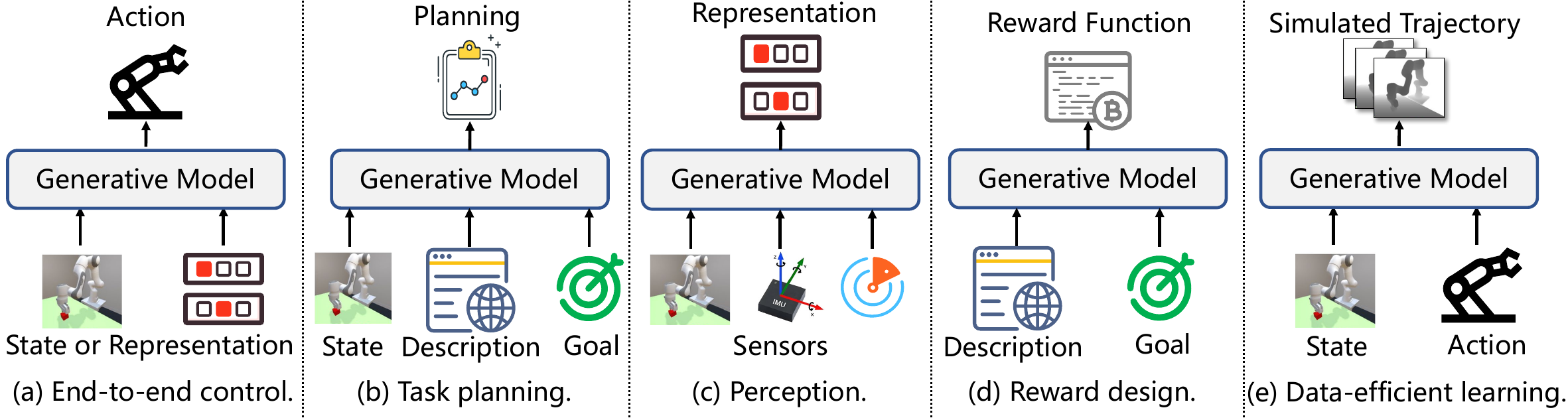}
\caption{An illustration of different roles generative models played in embodied control.}
\label{fig:sa_gene}
\end{figure*}

Despite their progress, traditional learning methods often suffer from limited representational capacity, making them inadequate for handling complex scenarios such as multi-goal or multi-task decision making. These methods typically struggle with poor learning performance, inefficient use of external prior knowledge and low sample efficiency. To address these challenges, researchers have begun exploring the use of generative models, which offer stronger representational capabilities, to enable more efficient and flexible embodied intelligence \cite{Gao2025EndtoEndDriving,Wang2025VLMAutonomousDriving}. The most straightforward application of generative models is to serve as \textbf{end-to-end controllers} by directly outputting executable actions, thereby transferring their internal prior knowledge into embodied agents. However, many embodied tasks involve fine-grained low-level behaviors which generative models often struggle to understand and control effectively. Inspired by the principles of hierarchical learning, a growing body of work has adopted generative models for \textbf{generating high-level plans}, while relying on traditional control or learning-based methods to execute low-level skills. Beyond direct decision-making, generative models are also widely used to support the learning process. Applications include \textbf{integrating sensory information for perception}, \textbf{generating reward functions}, and \textbf{synthesizing data to improve sample efficiency}.

\paragraph{End-to-end control.}
To enhance the utilization of prior knowledge and the processing of complex, multi-modal observations during decision-making, a growing body of research has explored using generative models, particularly large VLMs, as direct decision-makers, as shown in Figure~\ref{fig:sa_gene}(a). Early approaches typically involve formatting inputs such as states and goals in natural language and feed them into pre-trained generative models, enabling action inference without additional training~\cite{chen2023towards, Jiang2024EndtoEndLLM6}. However, due to the limited scope of scenarios covered during pre-training, the models often struggle to generalize to the wide range of real-world decision-making tasks. Consequently, more studies have focused on fine-tuning generative models with domain-specific data~\cite{Xu2024EndtoEndLLM1, Huang2024EndtoEndLLM5, Driess2023EndtoEndLLM7, Kim2024EndtoEndLLM8, Mees2024EndtoEndLLM9, Intelligence2025EndtoEndLLM10}, or integrating them with additional networks~\cite{Dong2024EndtoEndLLM2, Shao2024EndtoEndLLM3, Bjorck2025EndtoEndLLM4} to improve the generalizability.

\paragraph{Task planning.}
Due to the limited capability of generative models in understanding low-level controls for many tasks, combined with the substantial resources and costs required for domain-specific fine-tuning, directly employing pre-trained generative models for high-level task planning presents a simpler and more widely applicable approach. Leveraging the reasoning~\cite{Wei2022ChainofThought} and reflection capabilities~\cite{Xie2023SearchReflection,Lin2023AgentReflection,Hu2023GPTReflection,Huang2023LLMReflection} inherent, pre-trained generative models can effectively make goal-oriented plans through iterative reasoning and reflection processes upon inputs such as environmental descriptions, states and objectives~\cite{icher2023PaLMPlanning,song2023llm,Vemprala2024GPTPlanning}, as shown in Figure~\ref{fig:sa_gene}(b). For example, given a high-level task like ``pour a glass of water and place it on the table'', an LLM can decompose the task into a clear sequence of actionable steps: (S1) pick up the cup from the table, (S2) locate the water dispenser, (S3) align the cup with the dispenser's spout, (S4) activate the dispenser and wait until the cup is filled, and (S5) place the filled cup back on the table. Meanwhile, safety in embodied AI planning has garnered increasing attention in recent years~\cite{zhang2024badrobot,yin2024safeagentbench}.

\paragraph{Perception.}

Beyond decision-making, the strong expressive capabilities of generative models also make them effective in assisting embodied agents with environmental perception. For instance, HPT~\cite{Wang2024TransformerPerception2} utilizes the powerful sequence processing ability of Transformer architectures to fuse sensory data collected across multiple timesteps from various sources like images, IMU sensors, and radar, generating efficient observational representations, as shown in Figure~\ref{fig:sa_gene}(c). Similarly, numerous methods leverage Transformer, diffusion Model, and Mamba architectures for perceptual fusion~\cite{Li2020TransformerPerception1,Mei2025TransformerPerception3,Li2024MambaPerception1,Xu2024DiffusionPerception1,Zhong2025DiffusionPerception2,Tian2025DiffusionPerception3}. Additionally, leveraging pre-trained VLMs for semantic understanding of multimodal perception data provides another effective approach to support embodied agents in perception tasks~\cite{Zheng2023LLMPerception1}.

\paragraph{Reward design.}
To address the challenge of designing high-quality reward functions in complex real-world scenarios, utilizing pre-trained generative models for reward design has also become an effective approach to facilitate embodied agents in learning policies beyond perception. Reward design methods can be broadly categorized into reward signal generation and reward function generation. In reward signal generation methods, generative models receive state, action, and goal information at each decision step, producing reward signals in real-time~\cite{Kwon2023RewardSignalwithLLM1,lee2024llm}. Although this method offers flexibility and allows for dynamic adaptation to environmental changes, it incurs substantial computational costs due to frequent calls to generative models during training. In contrast, reward function generation methods only require a few pre-training calls to generative models before training to produce a reward function based on task descriptions and goals~\cite{yu2023language,Ma2023RewardCodewithLLM2,xie2023text2reward,zeng2024learning,Sun2024RewardCodewithLLM1}, as shown in Figure~\ref{fig:sa_gene}(d). During training, the reward function rather than the generative model is employed at each step, significantly reducing computational overhead.

\paragraph{Data-efficient learning.}
Furthermore, since interactions in real physical environments are costly and errors in robotic manipulation can lead to physical damage or system failures, employing generative models for data generation to enhance sample efficiency is another crucial approach in assisting policy learning \cite{Zhang2023WM1,Robine2023WM2,Micheli2023WM3}. Learning world models based on generative model architectures like Transformer and diffusion Models, and subsequently generating data via these world models, represents a mainstream method to boost sample efficiency~\cite{hu2023gaia,jia2023adriver,zhang2023copilot4d,zhang2024bevworld,mazzaglia2024gen,wang2025world}, as shown in Figure~\ref{fig:sa_gene}(e). In higher-level language-abstracted environments, some methods directly use pre-trained generative models as world models for future trajectory predictions~\cite{nottingham2023embodied}. Another class of approaches, rather than learning explicit world models, utilizes generative models to create diverse simulation environments, generating additional data through interactions between agents and simulated environments~\cite{wang2023robogen,wang2023gensim,yang2024holodeck,zala2024envgen}.

\subsection{Benchmarks} \label{subsec:SABM} 
\begin{figure*}
	\centering    
	\subfloat[ALFRED \cite{Shridhar2020Alfred}]{
		\label{fig:ALFRED}     
		\includegraphics[width=0.24\columnwidth]{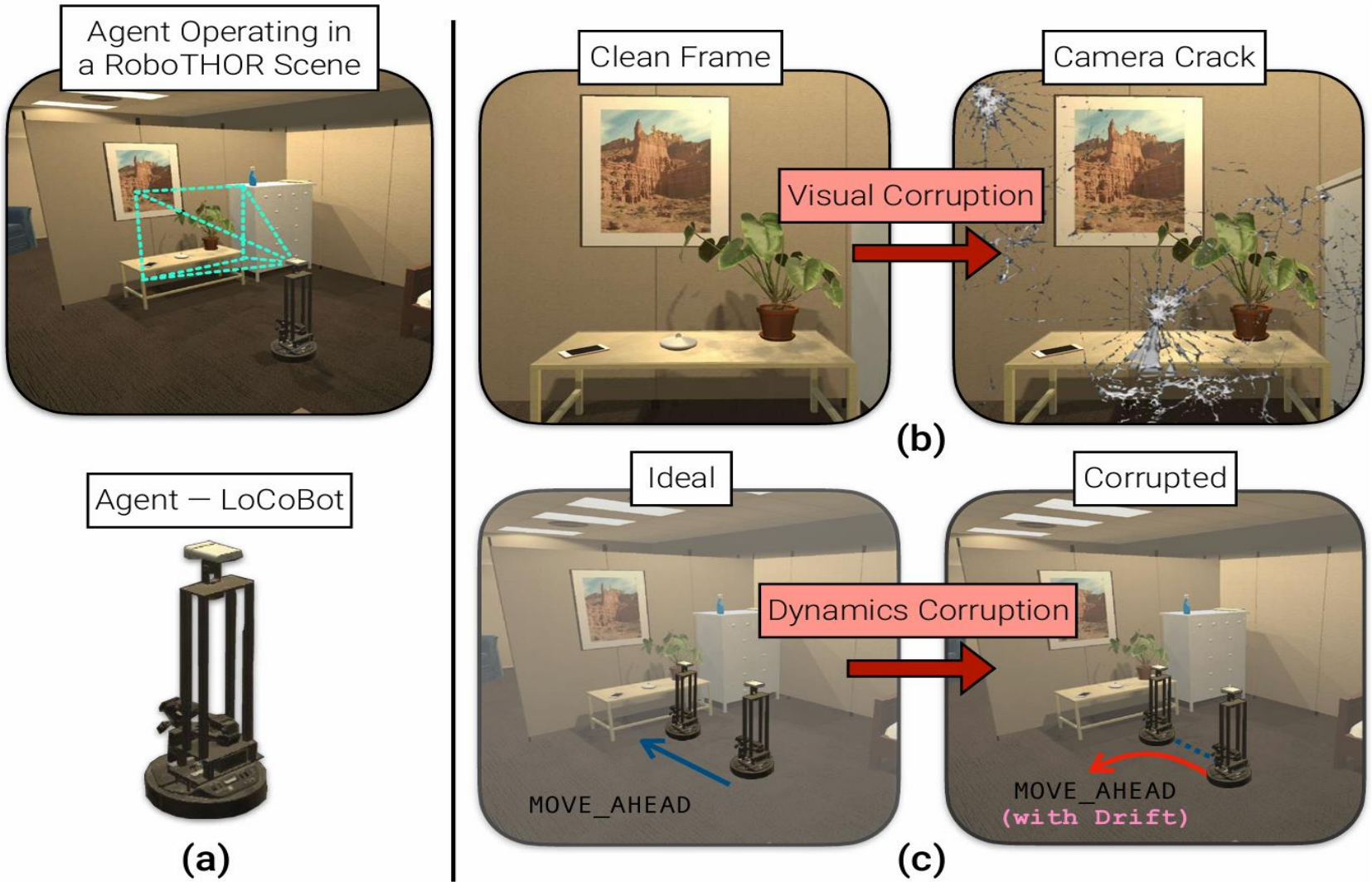}  
	}
        \subfloat[RoboTHOR \cite{Deitke2020RoboTHOR}]{ 
		\label{fig:RoboTHOR}     
		\includegraphics[width=0.24\columnwidth]{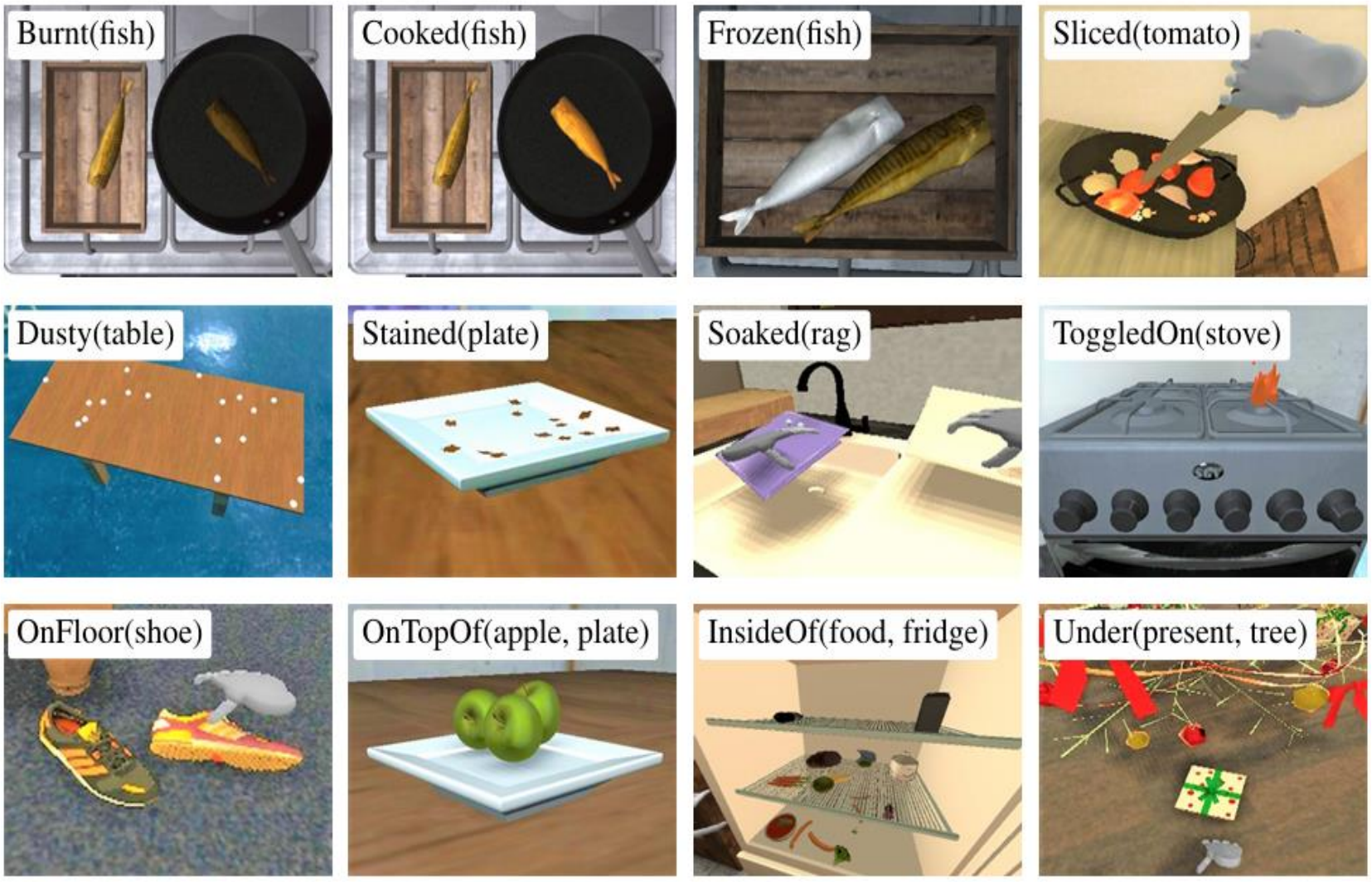}     
	}    
        \subfloat[RobustNav \cite{Chattopadhyay2021RobustNav}]{
		\label{fig:RobustNav}     
		\includegraphics[width=0.24\columnwidth]{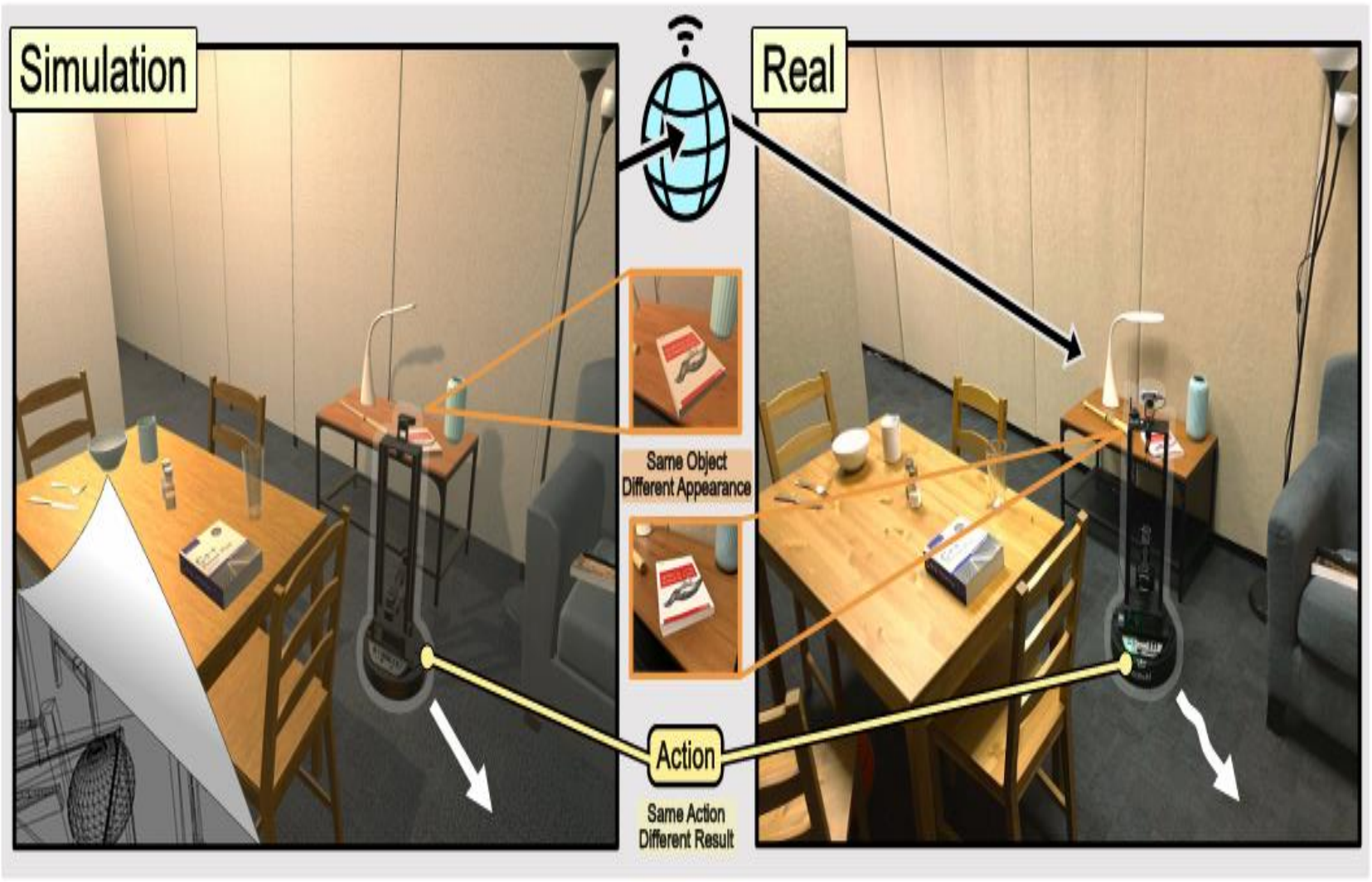}     
	}
	\subfloat[Behavior \cite{Srivastava2022Behavior}]{ 
		\label{fig:Behavior}     
		\includegraphics[width=0.24\columnwidth]{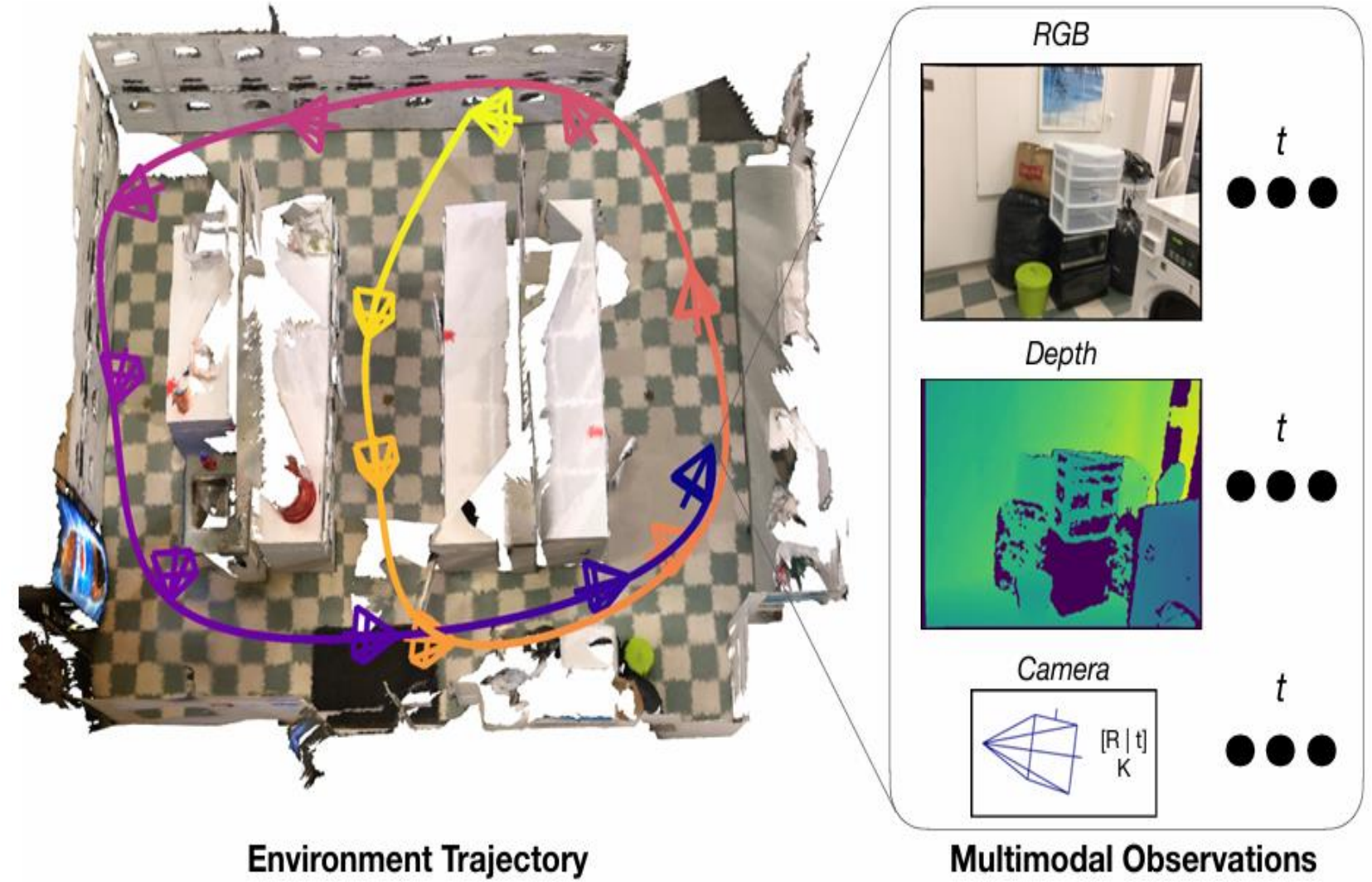}     
	}  
    \\
	\subfloat[ManiSkill2 \cite{Gu2023ManiSkill2}]{ 
		\label{fig:ManiSkill2}     
		\includegraphics[width=0.24\columnwidth]{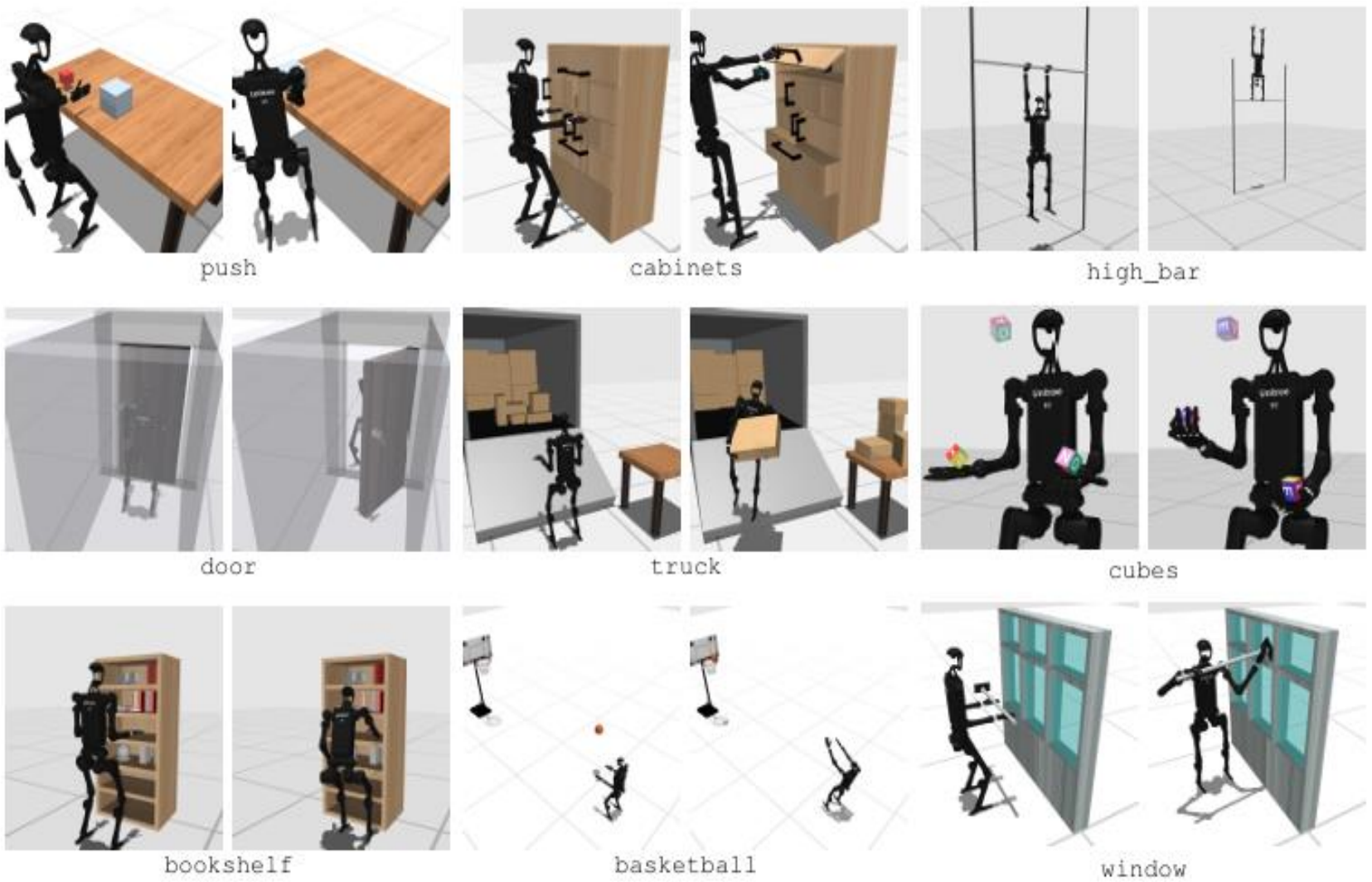}     
	}  
	\subfloat[EgoCOT \cite{Yao2023EmbodiedGPT}]{ 
		\label{fig:EgoCOT}     
		\includegraphics[width=0.24\columnwidth]{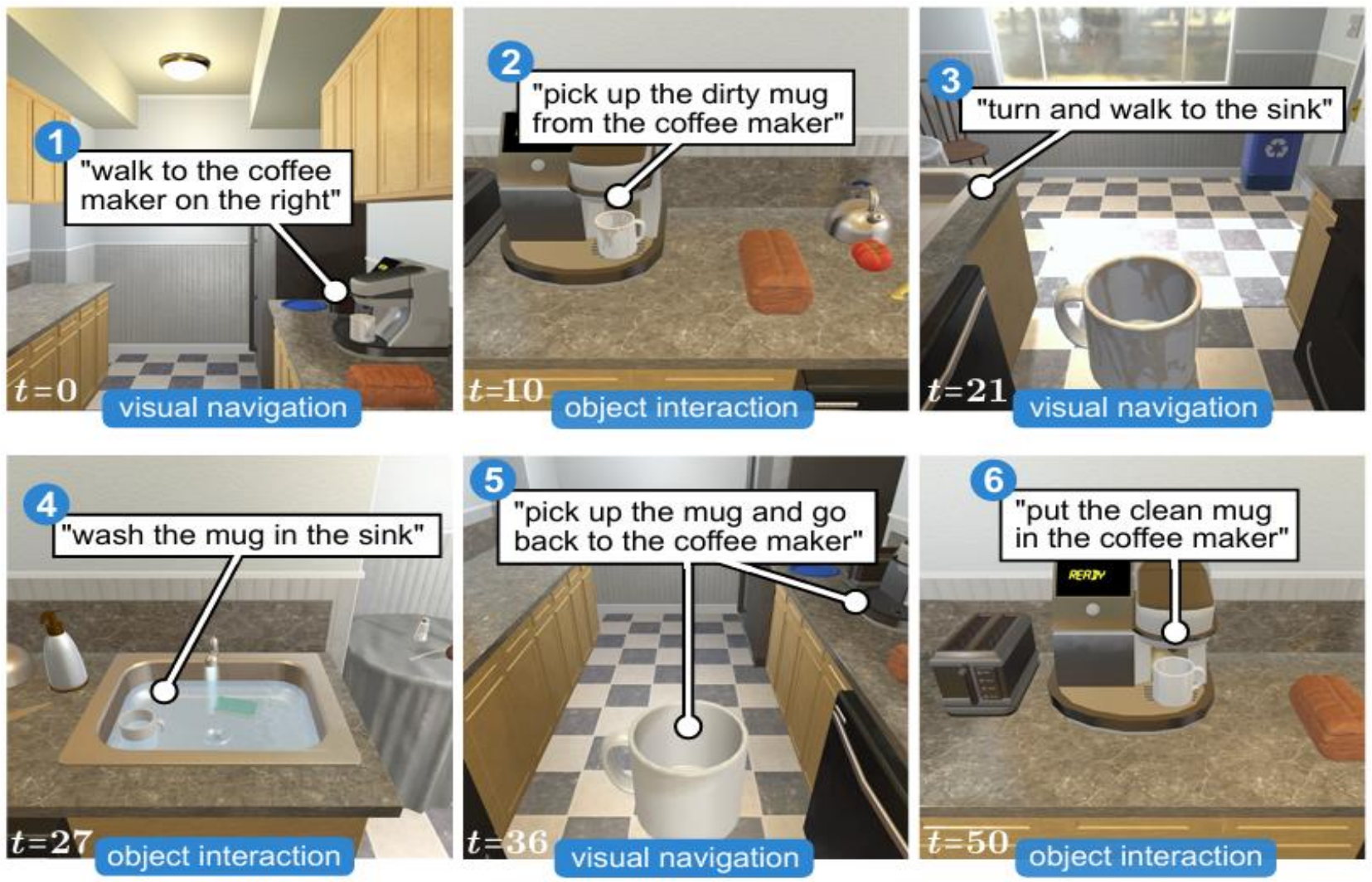}     
	}
	\subfloat[PhysBench \cite{Chow2025physbench}]{ 
		\label{fig:PhysBench}     
		\includegraphics[width=0.24\columnwidth]{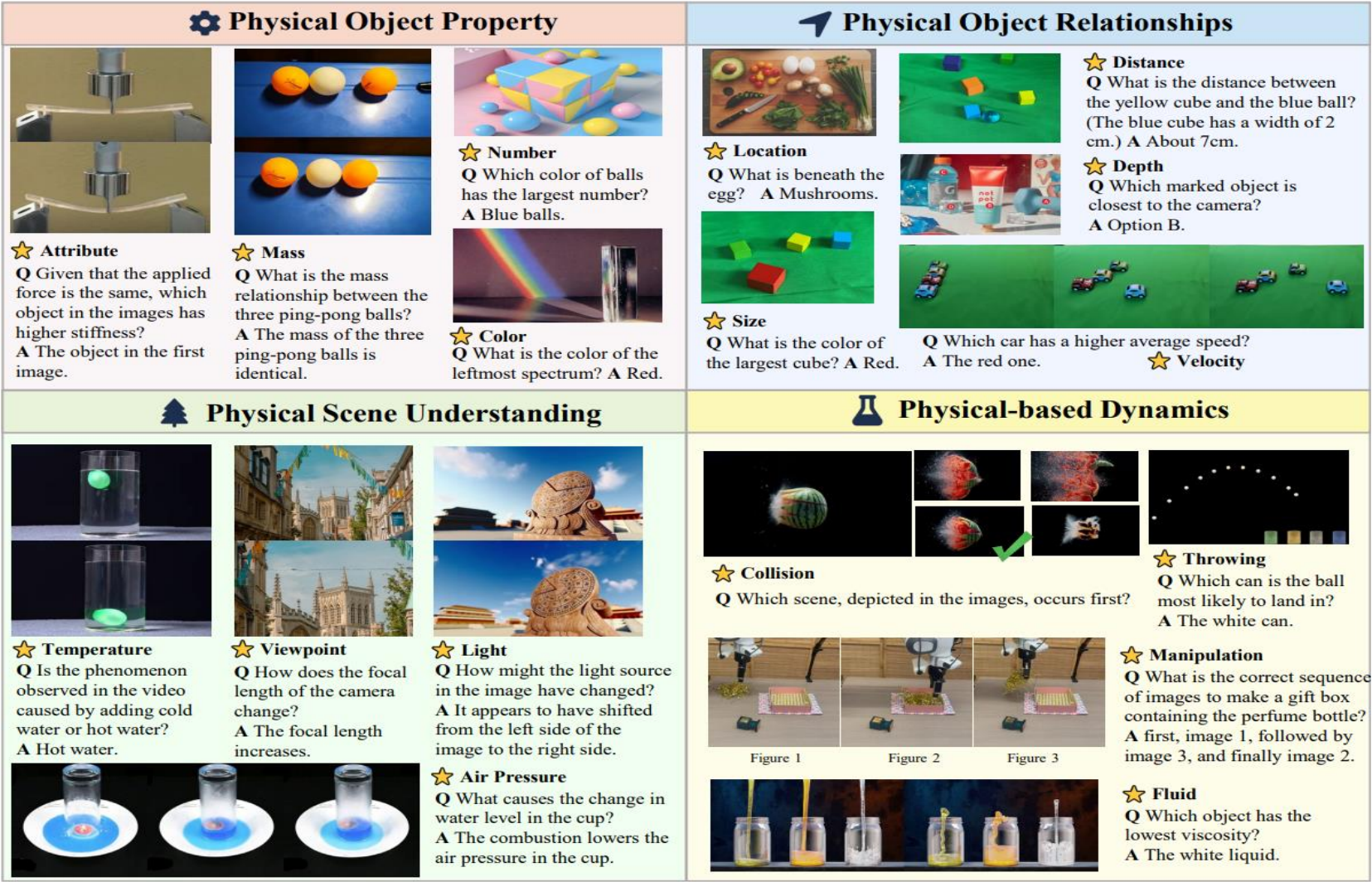}     
	}
	\subfloat[HumanoidBench \cite{Sferrazza2024HumanoidBench}]{ 
		\label{fig:HumanoidBench}     
		\includegraphics[width=0.24\columnwidth]{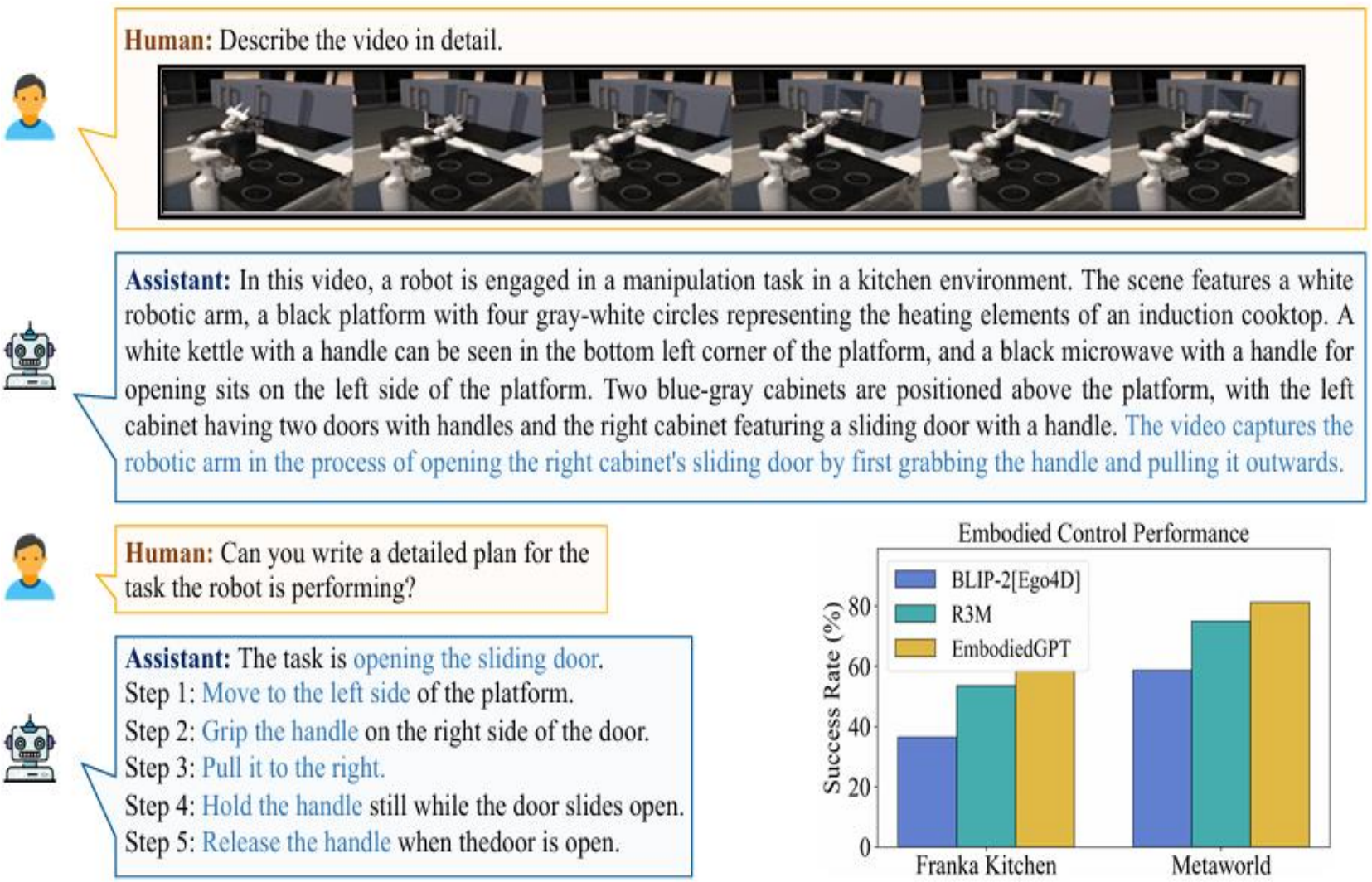}     
	}
    \\
	\subfloat[Open X-Embodiement \cite{Neill2024SABenchmark}]{ 
		\label{Open X-Embodiement}     
		\includegraphics[width=0.24\columnwidth]{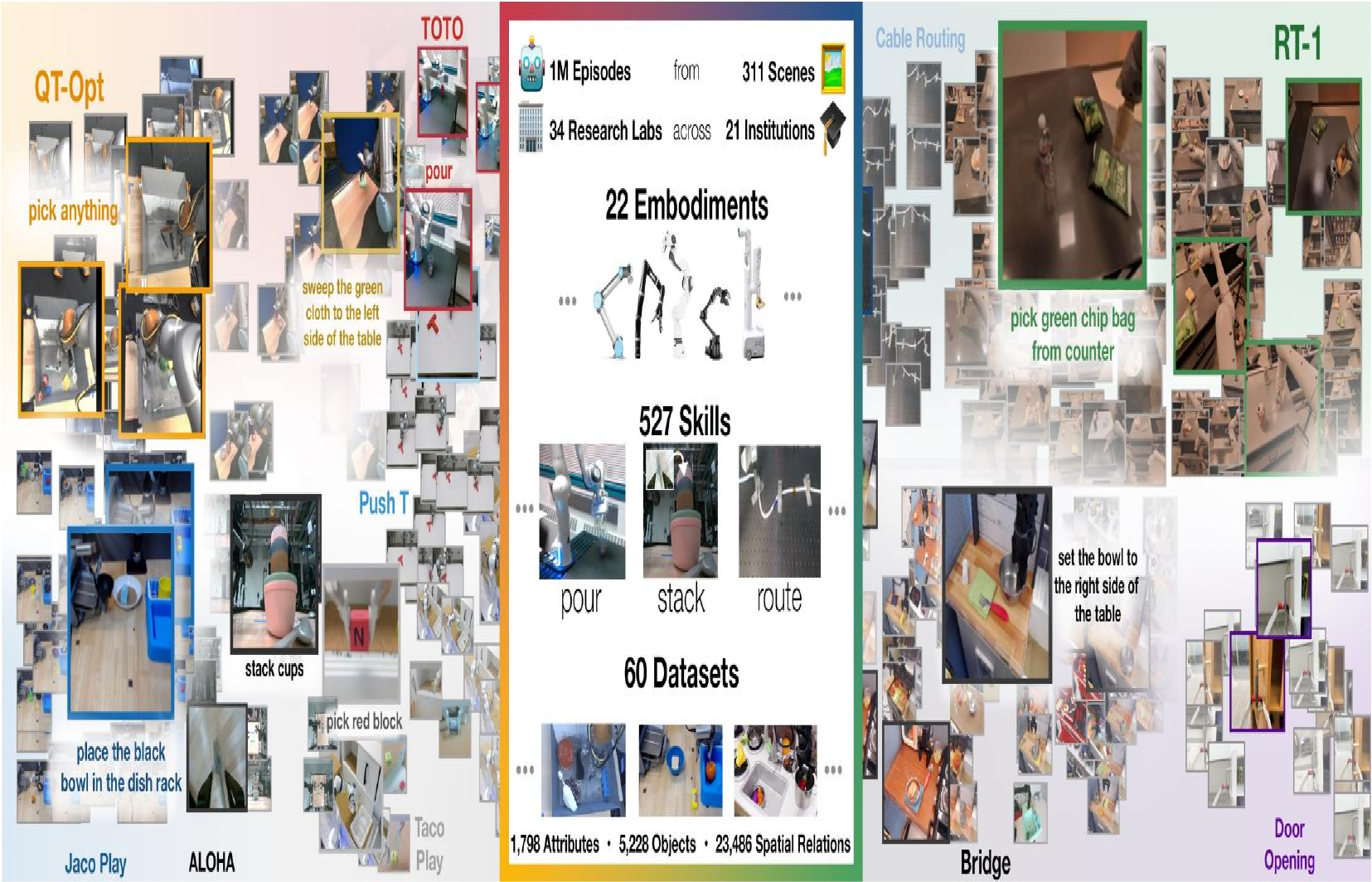}     
	}
	\subfloat[OpenEQA \cite{Majumdar2024OpenEQA}]{ 
		\label{fig:OpenEQA}     
		\includegraphics[width=0.24\columnwidth]{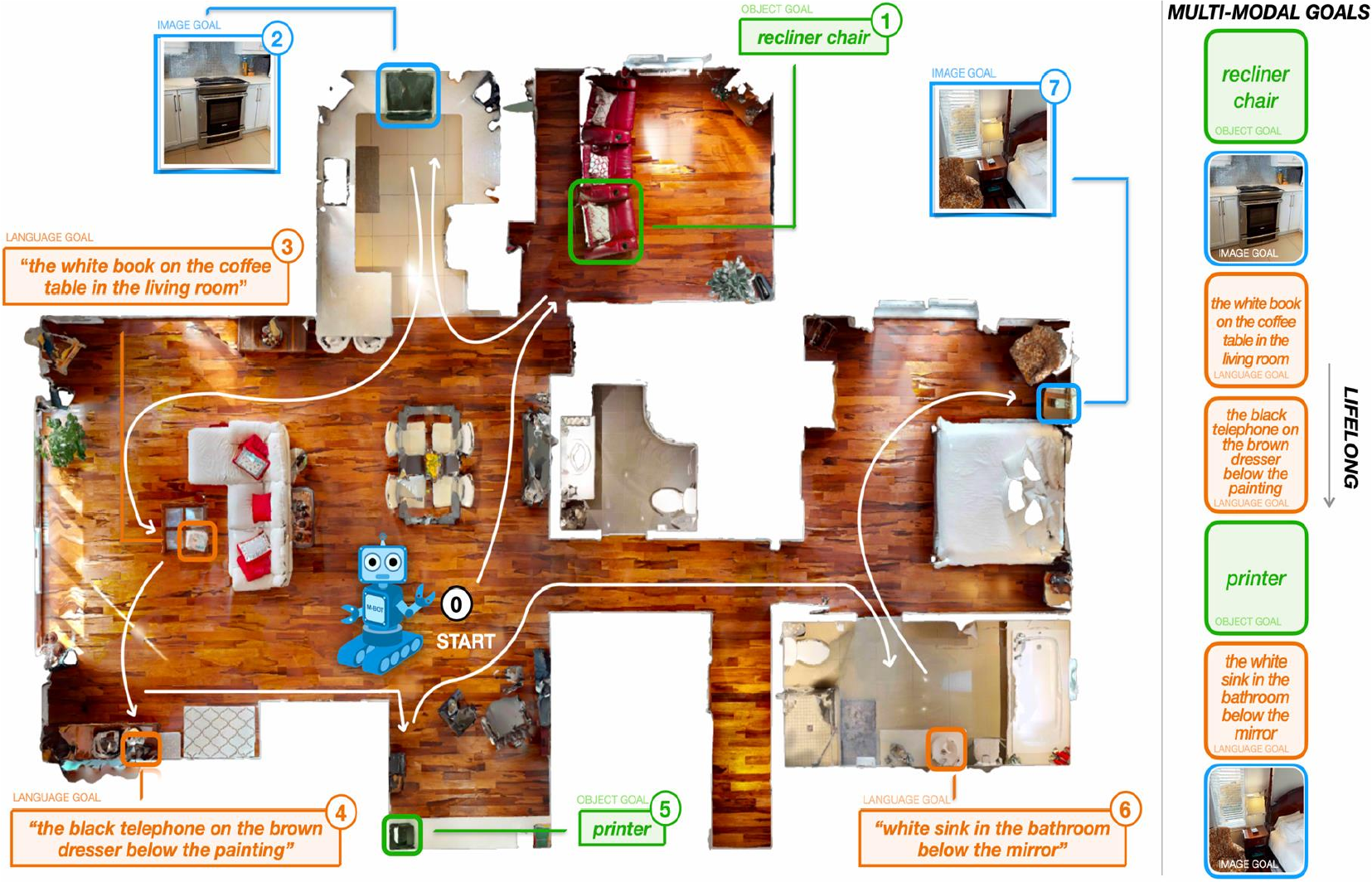}     
	}
	\subfloat[EAI \cite{Li2024EmbodiedAgentInterface}]{ 
		\label{fig:Embodied-Agent-Interface}     
		\includegraphics[width=0.24\columnwidth]{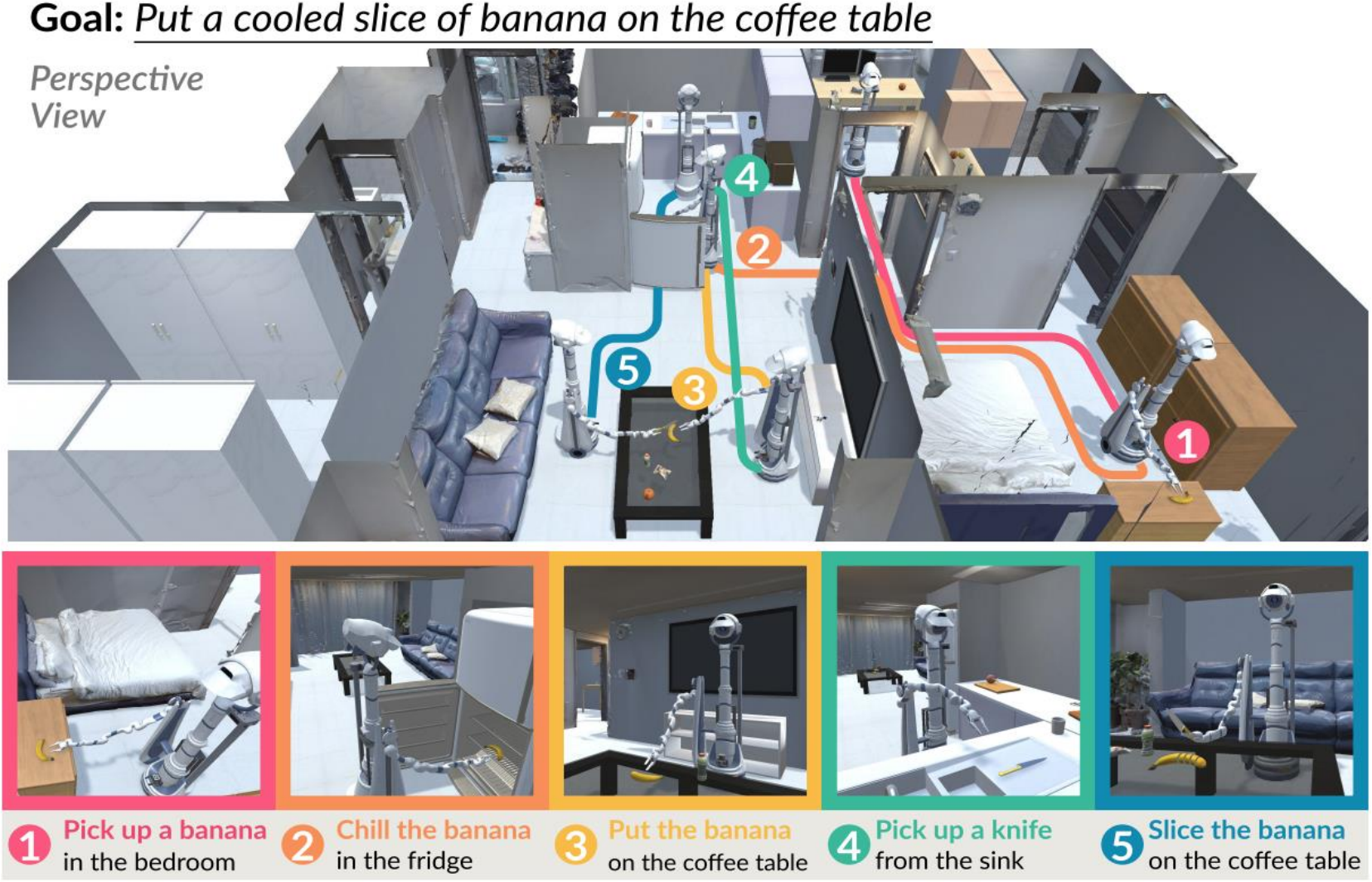}     
	}
	\subfloat[ReALFRED \cite{Kim2024ReALFRED}]{ 
		\label{fig:ReALFRED}     
		\includegraphics[width=0.24\columnwidth]{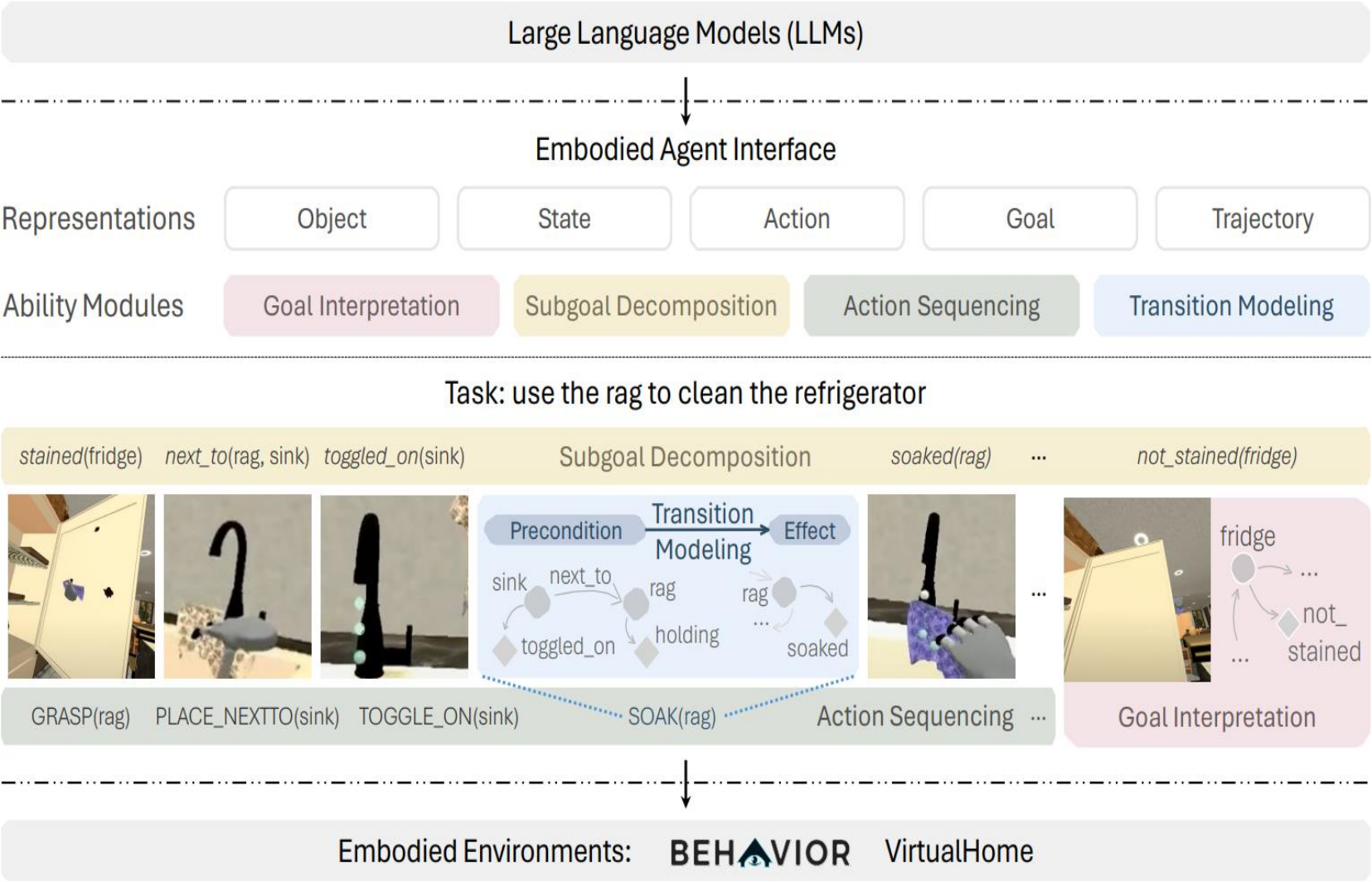}     
	}   
	\caption{An overview of all multi-embodied benchmarks listed in Table \ref{tbl:SABenchmarks}.}
    \label{fig:SABenchmarks}
\end{figure*}
\begin{table*}[h]
\centering
\caption{Benchmarks for testing single agent embodied AI. }
\small
\label{tbl:SABenchmarks}
\resizebox{\textwidth}{!}{
\begin{tabular}{l|c|ccc|cccc|cc|ccc|ccccc}
\toprule
& \multicolumn{1}{c|}{\textbf{Year}} 
& \multicolumn{3}{c|}{\textbf{Category}} 
& \multicolumn{4}{c|}{\textbf{Input}} 
& \multicolumn{2}{c|}{\textbf{View}} 
& \multicolumn{3}{c|}{\textbf{Data}} 
& \multicolumn{4}{c}{\textbf{Agent Type}} 
\\
& \rotatebox{90}{\quad}
& \rotatebox{90}{Perception} & \rotatebox{90}{Planning} & \rotatebox{90}{Control}
& \rotatebox{90}{Image} & \rotatebox{90}{Pointcloud} & \rotatebox{90}{Language} & \rotatebox{90}{Proprioception}
& \rotatebox{90}{Global} & \rotatebox{90}{Local}
& \rotatebox{90}{Real} & \rotatebox{90}{Sim} & \rotatebox{90}{Hybrid}
& \rotatebox{90}{Humanoid} & \rotatebox{90}{Quadruped} & \rotatebox{90}{UAV/UGV/USV} & \rotatebox{90}{Manipulator} & \rotatebox{90}{No certain type}
\\
\midrule
\href{https://github.com/askforalfred/alfred}{ALFRED} \cite{Shridhar2020Alfred} & 2020 & \checkmark & \checkmark & \checkmark & \checkmark &  & \checkmark &  &  & \checkmark &  & \checkmark &  &  &  &  & \checkmark &  \\
\href{ai2thor.allenai.org/robothor}{RoboTHOR} \cite{Deitke2020RoboTHOR} & 2020 & \checkmark & \checkmark & \checkmark & \checkmark &  &  &  &  & \checkmark &  &  & \checkmark &  &  &  & \checkmark &  \\
\href{prior.allenai.org/projects/robustnav}{RobustNav} \cite{Chattopadhyay2021RobustNav} & 2021 & \checkmark & \checkmark & \checkmark & \checkmark &  &  & \checkmark &  & \checkmark &  & \checkmark &  &  &  & \checkmark &  &  \\
\href{https://behavior.stanford.edu/}{Behavior} \cite{Srivastava2022Behavior} & 2021 & \checkmark & \checkmark & \checkmark & \checkmark &  & \checkmark & \checkmark &  & \checkmark &  & \checkmark &  & \checkmark &  & \checkmark & \checkmark &  \\
\href{https://github.com/haosulab/ManiSkill2}{ManiSkill2} \cite{Gu2023ManiSkill2} & 2023 &  & \checkmark & \checkmark & \checkmark & \checkmark &  & \checkmark &  & \checkmark &  & \checkmark &  &  &  &  & \checkmark &  \\
\href{https://embodiedgpt.github.io/}{EgoCOT} \cite{Yao2023EmbodiedGPT} & 2023 & \checkmark & \checkmark & \checkmark & \checkmark &  & \checkmark &  &  & \checkmark &  &  & \checkmark &  &  &  & \checkmark &  \\
\href{https://humanoid-bench.github.io/}{HumanoidBench} \cite{Sferrazza2024HumanoidBench} & 2024 & \checkmark & \checkmark & \checkmark & \checkmark &  &  & \checkmark &  & \checkmark &  & \checkmark &  & \checkmark &  &  & \checkmark &  \\
\href{https://robotics-transformer-x.github.io/}{Open X-Embodiment} \cite{Neill2024SABenchmark} & 2024 & \checkmark & \checkmark & \checkmark & \checkmark & \checkmark & \checkmark &  &  & \checkmark & \checkmark &  &  &  &  & \checkmark & \checkmark &  \\
\href{https://open-eqa.github.io/}{OpenEQA} \cite{Majumdar2024OpenEQA} & 2024 & \checkmark & \checkmark & \checkmark & \checkmark & \checkmark & \checkmark &  &  & \checkmark &  &  & \checkmark &  &  & \checkmark &  &  \\
\href{embodied-agent-interface.github.io}{Embodied-Agent-Interface} \cite{Li2024EmbodiedAgentInterface} & 2024 &  & \checkmark & \checkmark & \checkmark &  & \checkmark &  &  & \checkmark &  & \checkmark &  & \checkmark &  &  &  &  \\
\href{https://github.com/snumprlab/realfred}{ReALFRED} \cite{Kim2024ReALFRED} & 2024 & \checkmark & \checkmark & \checkmark & \checkmark &  & \checkmark &  &  & \checkmark &  & \checkmark &  &  &  & \checkmark & \checkmark &  \\
\href{https://physbench.github.io/}{PhysBench} \cite{Chow2025physbench} & 2025 & \checkmark &  &  & \checkmark &  & \checkmark &  & \checkmark &  &  &  & \checkmark &  &  &  &  & \checkmark \\
\bottomrule
\end{tabular}
}
\end{table*}
Although embodied agents are designed for interactive tasks in the real world, benchmarks still play a crucial role in standardizing performance measurement, guiding research focus, and reducing real-world testing costs. In recent years, there has been a surge of progress in the development of embodied AI benchmarks. We summarize several representative examples in Table \ref{tbl:SABenchmarks} and present them in Figure \ref{fig:SABenchmarks}. The listed benchmarks include:
\begin{itemize}
  \item ALFRED \cite{Shridhar2020Alfred} is a benchmark for evaluating embodied agents’ ability to ground and execute free-form natural language instructions through physical interactions. Built on the AI2-THOR simulator, it includes 120 visually and functionally diverse household scenes, where agents are tasked with completing goals involving navigation, object manipulation, and irreversible state changes. These tasks span seven categories of household activities and involve 58 distinct object types. To support learning and evaluation, the dataset provides 25,743 human-authored instructions paired with 8,055 expert demonstrations. Agents perceive the environment from egocentric RGB-D inputs and act using a predefined set of 13 discrete low-level actions. Interaction targets are defined through pixel-wise masks, which guide object-level behaviors. Task performance is evaluated primarily based on goal completion, and the benchmark includes all code, annotations, and environment assets necessary to support reproducible research.
    \item RoboTHOR \cite{Deitke2020RoboTHOR} is a sim-to-real embodied benchmark that provides paired simulated and physical environments designed for consistent cross-domain evaluations. It includes 75 training and validation scenes, along with 24 held-out test scenes (14 for test-dev and 10 for test-standard), which all constructed from a modular asset library that supports flexible reconfiguration and expansion. The agents in RoboTHOR interact with the environment via a unified AI2-THOR API, and the policies can be remotely deployed on real LoCoBot robots whose noisy dynamics are closely matched to those in the simulations. RoboTHOR is centered on semantic navigation tasks with performance evaluated using success rate, SPL and path efficiency in both virtual and physical settings. To promote reproducible research, all code, assets and remote-access infrastructure are publicly released.
    \item RobustNav \cite{Chattopadhyay2021RobustNav} is a benchmark framework for evaluating the robustness of embodied navigation agents under realistic environmental and sensor corruptions. It extends standard PointGoal and ObjectGoal navigation tasks by introducing seven visual corruptions (e.g., motion blur, camera crack, low lighting) and three dynamics corruptions (e.g., motion bias, drift, motor failure) across 15 validation scenes. Agents are evaluated based on Success Rate and SPL, both before and after a fixed unsupervised “calibration budget.” The benchmark highlights significant performance degradation under corruption, emphasizing the importance of robust perception, multi-modal sensing, and adaptive policy learning.
    \item Behavior \cite{Srivastava2022Behavior}  is a benchmark for evaluating embodied AI agents in everyday household activities within virtual, interactive environments. It defines 100 realistic, diverse, and complex chores using a predicate logic–based language, and supports infinite scene-agnostic instantiations within iGibson 2.0. The benchmark includes 500 human VR demonstrations and provides evaluation metrics such as success score, task efficiency, and human-centric performance. All code, task definitions, and data are publicly available to facilitate reproducible research in embodied AI.
    \item ManiSkill2 \cite{Gu2023ManiSkill2} is a unified benchmark for generalizable robotic manipulation, comprising 20 task families that span stationary and mobile platforms, single- and dual-arm setups, and both rigid- and soft-body interactions. It includes over 2,000 object models and more than 4 million demonstration frames. The suite supports point-cloud, RGB-D, and state-based inputs, with control at both the joint and end-effector levels. Built on the SAPIEN engine \cite{Xiang2020SAPIENdataset} and interfaced via OpenAI Gym \cite{brockman2016openai}, ManiSkill2 features asynchronous rendering and a warp MPM–based soft-body simulator for high-throughput data collection. An open-source codebase and online challenge platform support reproducible evaluation across perception–planning–execution pipelines, RL, and IL methods. 
    \item EgoCOT and EgoVQA \cite{Yao2023EmbodiedGPT} provides 2,900 hours of egocentric human–object interaction video clips accompanied by 3.85 million machine-generated, semantics-filtered, and human-verified chain-of-thought planning instructions. Complementarily, EgoVQA offers 200 million video–question–answer pairs automatically generated and CLIP-filtered from the Ego4D corpus. Together, these two datasets establish a unified, large-scale benchmark for end-to-end evaluation of embodied planning, control, and visual question answering in realistic first-person settings.
    \item HumanoidBench \cite{Sferrazza2024HumanoidBench}  is a high-dimensional simulation benchmark for humanoid robots, featuring 27 tasks including 15 whole-body manipulation tasks (e.g., truck unloading, basketball catch-and-throw) and 12 locomotion tasks (e.g., running, maze navigation). The benchmark uses a Unitree H1 humanoid equipped with dual Shadow Hands, yielding a 61-dimensional action space and a 151-dimensional observation space. Built on MuJoCo \cite{todorov2012mujoco}, it enables standardized evaluation of both flat and hierarchical RL methods in terms of success, efficiency, and robustness. Its open-source codebase and unified interfaces support reproducible research in complex humanoid control and learning.
    \item Open X-Embodiment \cite{Neill2024SABenchmark} aggregates over one million real-world robot trajectories collected from 22 distinct robot embodiments, spanning a total of 60 datasets. All trajectories are converted into a unified RLDS format \cite{Ramos2021RLDS}, ensuring consistency and interoperability across platforms. To support broad adoption and reproducible research, the benchmark provides standardized APIs, a suite of open-source tools, and pretrained RT-X model checkpoints, enabling efficient training, evaluation, and comparison of generalist robotic policies across diverse embodiments and task domains.
    \item OpenEQA \cite{Majumdar2024OpenEQA} is the first open-vocabulary benchmark for embodied question answering, featuring over 1,600 human-authored questions grounded in more than 180 real-world scanned environments. It supports two evaluation paradigms—episodic memory and active exploration—where agents perceive the environment through RGB-D inputs and must reason about spatial, semantic, or temporal cues. Agent performance is assessed using an automated LLM-Match protocol, which enables consistent and reproducible evaluation of open-ended, language-driven understanding in complex 3D environments.
    \item Embodied-Agent-Interface \cite{Li2024EmbodiedAgentInterface} is a standardized benchmark for evaluating LLM-driven decision-making in embodied environments. It formalizes task goals using linear temporal logic and structures the decision-making process into four core modules: goal interpretation, sub-goal decomposition, action sequencing, and transition modeling. Each module is evaluated using fine-grained metrics such as grammar correctness, hallucination frequency, step completeness, and ordering accuracy. The benchmark is implemented on top of BEHAVIOR and VirtualHome, and incorporates automated sub-task checkers as well as PDDL-based planners to enable systematic evaluation and reproducible comparison across 18 LLMs in translating natural language instructions into executable plans.
    \item ReALFRED \cite{Kim2024ReALFRED} is a photo-realistic benchmark for embodied instruction following, designed to bridge the gap between simulation and real-world complexity. It features 150 multi-room, 3D-scanned interactive environments, along with 30,696 human-annotated natural language directives and planner-generated expert demonstrations spanning seven household task categories. Evaluation is based on standardized metrics for task success and execution efficiency, capturing both correctness and behavioral quality. To support reproducible research, ReALFRED provides open-source code and a complete dataset for studying vision-and-language–conditioned decision-making in realistic settings.
    \item PhysBench \cite{Chow2025physbench} is a large benchmark designed for evaluating VLMs on their understanding of physical concepts in real-world scenarios. It consists of 10,002 multiple-choice instances that interleave video, image and text modalities, covering four core domains: physical object properties, object relationships, scene understanding and physical dynamics. These domains are further divided into 19 fine-grained sub-task types, enabling detailed assessment of model capabilities across diverse aspects of physical reasoning. The dataset of PhysBench is constructed from a combination of multiple resources including real-world video recordings, web-sourced media and high-fidelity physical simulations. To support fair and reproducible evaluation, PhysBench includes an online platform with withheld test answers, allowing standardized comparison across 75 representative VLMs.
\end{itemize}
\begin{table*}[h]
\centering
\caption{Methods for building embodied AI in multi-agent settings. }
\small
\label{tbl:MAmethods}
\resizebox{\textwidth}{!}{
\begin{tabular}{l|c|ccc|cccc|cc|ccccc|ccc}
\toprule
& \multicolumn{1}{c|}{\textbf{Year}} 
& \multicolumn{3}{c|}{\textbf{Category}} 
& \multicolumn{4}{c|}{\textbf{Input}} 
& \multicolumn{2}{c|}{\textbf{View}} 
& \multicolumn{5}{c|}{\textbf{Agent Type}} 
& \multicolumn{3}{c}{\textbf{Implement}} 
\\
& \rotatebox{90}{\quad}
& \rotatebox{90}{Perception} & \rotatebox{90}{Planning} & \rotatebox{90}{Control}
& \rotatebox{90}{Image} & \rotatebox{90}{Pointcloud} & \rotatebox{90}{Language} & \rotatebox{90}{Proprioception}
& \rotatebox{90}{Global} & \rotatebox{90}{Local}
& \rotatebox{90}{Humanoid} & \rotatebox{90}{Quadruped/Bipedal} & \rotatebox{90}{UAV/UGV/USV} & \rotatebox{90}{Manipulator} & \rotatebox{90}{No certain type} 
& \rotatebox{90}{Solver-based} & \rotatebox{90}{NN-based} & \rotatebox{90}{GM-based}
\\
\midrule
ACE \cite{Yu2023MAAsynchronousLearning2} & 2023 & \checkmark & \checkmark &  & \checkmark &  &  &  &  & \checkmark &  &  & \checkmark &  &  &  & \checkmark &  \\
HetGPPO \cite{Matteo2023MAHeterogeneousLearning1} & 2023 &  & \checkmark & \checkmark &  &  &  & \checkmark &  & \checkmark &  &  & \checkmark &  &  &  & \checkmark &  \\
Self-play \cite{Gao2023MASelfPlayLearning1} & 2023 & \checkmark & \checkmark & \checkmark & \checkmark &  &  & \checkmark & \checkmark & \checkmark &  &  & \checkmark &  &  &  & \checkmark &  \\
TwoStep \cite{Singh2024MALLMPlanning3} & 2024 &  & \checkmark &  &  &  & \checkmark &  & \checkmark &  &  &  &  &  & \checkmark &  &  & \checkmark \\
ZeroCAP \cite{Venkatesh2024MALLMPlanningandExecuting1} & 2024 & \checkmark & \checkmark &  & \checkmark &  & \checkmark &  & \checkmark &  &  &  & \checkmark &  &  &  &  & \checkmark \\
COHERENT \cite{Liu2024MALLMPlanningandExecuting2} & 2024 &  & \checkmark &  & \checkmark &  & \checkmark &  & \checkmark &  &  & \checkmark & \checkmark & \checkmark &  &  &  & \checkmark \\
LLaMAC \cite{Zhang2024MALLMLargeScaleDecentralizedCollabration1} & 2024 & \checkmark & \checkmark & \checkmark &  &  & \checkmark &  & \checkmark &  &  &  &  &  & \checkmark &  &  & \checkmark \\
FaGeL \cite{Liu2024MALLMHumanAgentCollaboration6} & 2024 & \checkmark & \checkmark &  &  &  & \checkmark &  &  & \checkmark &  &  &  &  & \checkmark &  &  & \checkmark \\
MacNet \cite{Qian2025MALLMLargeScaleDecentralizedCollabration2} & 2025 &  & \checkmark &  &  &  & \checkmark &  & \checkmark &  &  &  &  &  & \checkmark &  &  & \checkmark \\
CoELA \cite{Zhang2024MALLMHumanAgentCollaboration1} & 2025 & \checkmark & \checkmark & \checkmark & \checkmark & \checkmark & \checkmark & \checkmark &  & \checkmark & \checkmark &  &  &  &  &  &  & \checkmark \\
FISER \cite{Wan2025MALLMHumanAgentCollaboration5} & 2025 &  & \checkmark & \checkmark &  &  & \checkmark &  & \checkmark &  &  &  &  &  & \checkmark &  &  & \checkmark \\
REMAC \cite{Yuan2025MADynamicLearning1} & 2025 & \checkmark & \checkmark &  & \checkmark &  & \checkmark &  &  & \checkmark &  &  & \checkmark & \checkmark &  &  &  & \checkmark \\
LDPD \cite{Liu2025MALLMCentralizedTeacher} & 2025 &  & \checkmark & \checkmark &  &  &  & \checkmark &  & \checkmark &  &  & \checkmark &  &  &  & \checkmark & \checkmark \\
\bottomrule
\end{tabular}
}
\end{table*}
\section{Multi-agent Embodied AI}

Real-world embodied tasks often involve multiple agents or human-AI collaboration and competition, where dynamic interactions among agents in a shared environment give rise to emergent group-level behaviors that individuals cannot achieve alone. As a result, directly transferring methods designed for single-agent settings to MAS is typically inefficient. MAS research primarily focuses on enabling effective collaboration. Here, we review recent advances in multi-agent collaboration along the same three aspects as in the single-agent setting in Section \ref{sec:SAembodied}: \textbf{control and planning methods}, \textbf{learning-based methods} and \textbf{generative model-based methods}, typical methods are listed in Table \ref{tbl:MAmethods}. 

\subsection{Multi-agent Control and Planning} \label{subsec:MAcontrol} 
\begin{figure*}
\centering
\includegraphics[width=.95\columnwidth]{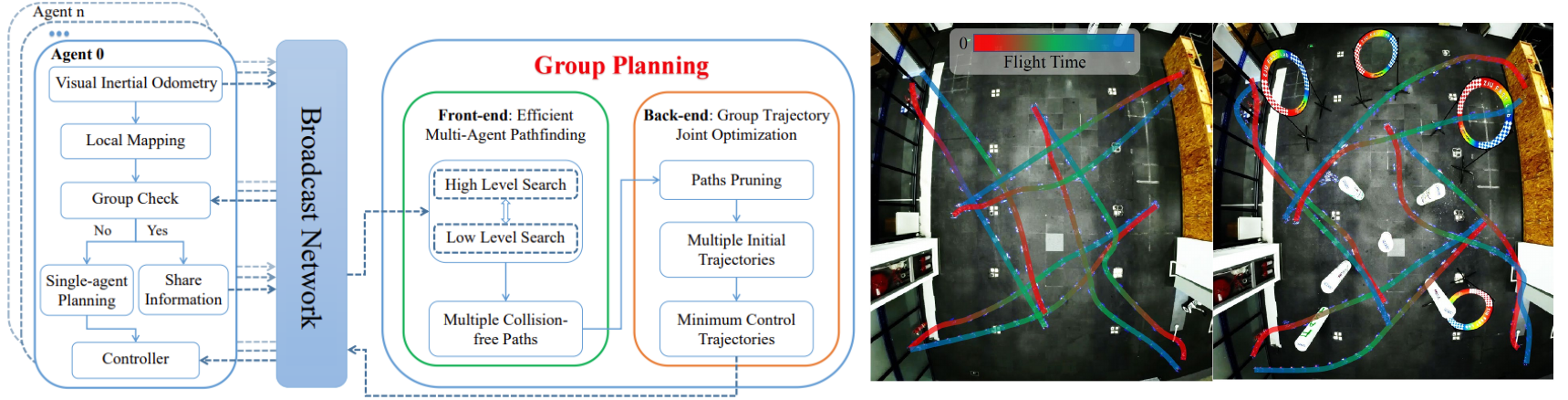}
\caption{An illustration of control-based multi-agent trajectory optimization~\cite{Hou2022GroupedSwarmPlan1}.}
\label{fig:ma_control}
\end{figure*}

In MAS, control-based methods remain a fundamental approach for achieving high-precision, real-time decision-making under task constraints. Early methods model MAS as single agents and perform centralized control and planning~\cite{Silver2005CentralizedSwarmPlan1, Augugliaro2012CentralizedSwarmPlan2} methods on them. However, these approaches face significant scalability challenges. To address this, some methods adopt distributed strategies that control each agent in the MAS independently~\cite{Tordesillas2022DecentralizedSwarmPlan1}, making them more suitable for large-scale MAS. Nevertheless, such fully decentralized methods often struggle with resolving conflicts between agents. To overcome these limitations, EMAPF~\cite{Hou2022GroupedSwarmPlan1} proposes a grouped multi-agent control framework. It dynamically clusters agents based on their spatial proximity, applying centralized control within each group while ensuring that inter-group control remains independent, as shown in Figure~\ref{fig:ma_control}. This enables efficient coordination among large teams of aerial robots. Alternatively, Swarm-Formation~\cite{Quan2022AsynchronousSwarmPlan1} employs a decentralized sequential decision-making scheme: agents are pre-ordered and each agent plans its trajectory based on its predecessor’s plan in turn, thereby eliminating inter-agent conflicts.

\subsection{Learning for Multi-agent Interaction} \label{subsec:MARL} 
Control-based methods in MAS, much like in single-agent scenarios, still face challenges such as high computational overhead and limited generalizability. As a result, learning-based approaches remain crucial in constructing multi-agent embodied AI. However, unlike single-agent settings, learning-based methods in MAS must additionally address unique challenges, including \textbf{asynchronous decision-making}, \textbf{heterogeneous team composition} and \textbf{open multi-agent environments}.

\begin{figure*}
\centering
\includegraphics[width=.9\columnwidth]{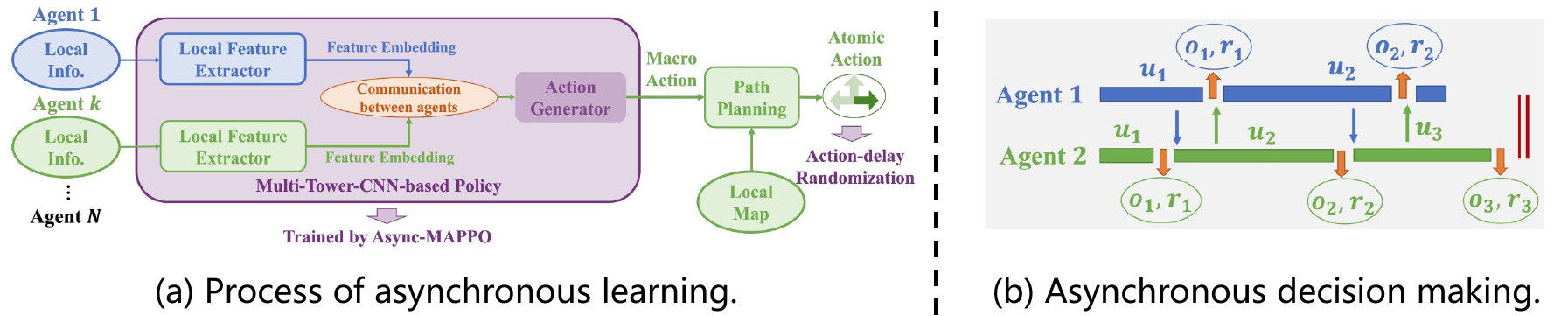}
\caption{An illustration of multi-agent asynchronous learning~\cite{Yu2023MAAsynchronousLearning2}.}
\label{fig:ma_async}
\end{figure*}

\paragraph{Asynchronous collaboration.} 
In multi-agent embodied systems, challenges such as communication delays and hardware heterogeneity across agents often disrupt synchronized interactions and feedback from the real environment, making effective policy learning under asynchronous decision-making a significant challenge. To address this, ACE~\cite{Yu2023MAAsynchronousLearning2} introduces the concept of macro-actions, where a macro-action serves as a centralized goal for the entire MAS. Individual agents then make multiple asynchronous decisions based on this goal, with delayed feedback from the environment provided only after the macro-action is completed. To facilitate policy learning in this setup, ACE employs a macro-action-based MAPPO algorithm, as shown in Figure~\ref{fig:ma_async}. This approach has proven effective and has inspired several other works tackling asynchronous decision-making challenges~\cite{Xiao2022MAAsynchronousLearning1, Xiao2025MAAsynchronousLearning3}.
\begin{figure*}
\centering
\includegraphics[width=.95\columnwidth]{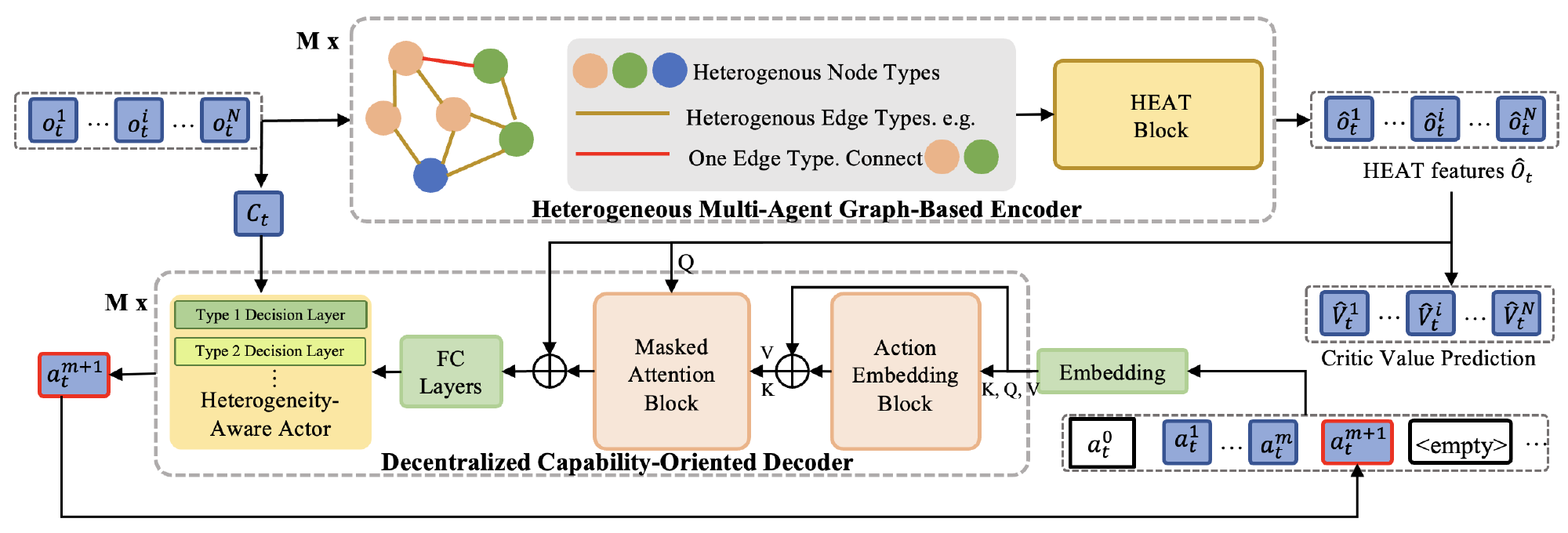}
\caption{An illustration of multi-agent heterogeneous learning~\cite{Cai2024MAchangescale1}.}
\label{fig:ma_hetor}
\end{figure*}

\paragraph{Heterogeneous collaboration.} 
Beyond differences in decision timing, another key distinction in embodied MAS is agent heterogeneity. This refers to the variations among agents in terms of perception capabilities, action spaces, task objectives, physical properties, communication abilities, and decision-making models. For instance, in a collaborative manufacturing scenario, autonomous vehicles might be responsible for transporting goods, while robotic arms handle sorting tasks. These two agent types differ significantly in their observation spaces, action spaces, and task goals, making them a prime example of heterogeneous agents. To tackle the challenges posed by such differences in observation and action spaces, approaches like HetGPPO~\cite{Matteo2023MAHeterogeneousLearning1} and COMAT~\cite{Cai2024MAchangescale1} propose using separate observation and policy networks for different agent types. These networks are connected through graph-based communication, enabling effective information exchange, as illustrated in the architecture of COMAT in Figure~\ref{fig:ma_hetor}. Another approach focuses on adapting the learning algorithm itself, such as by decomposing the advantage function across heterogeneous agents to facilitate more effective credit assignment~\cite{Zhong2024HARL}.

\begin{figure*}
\centering
\includegraphics[width=.95\columnwidth]{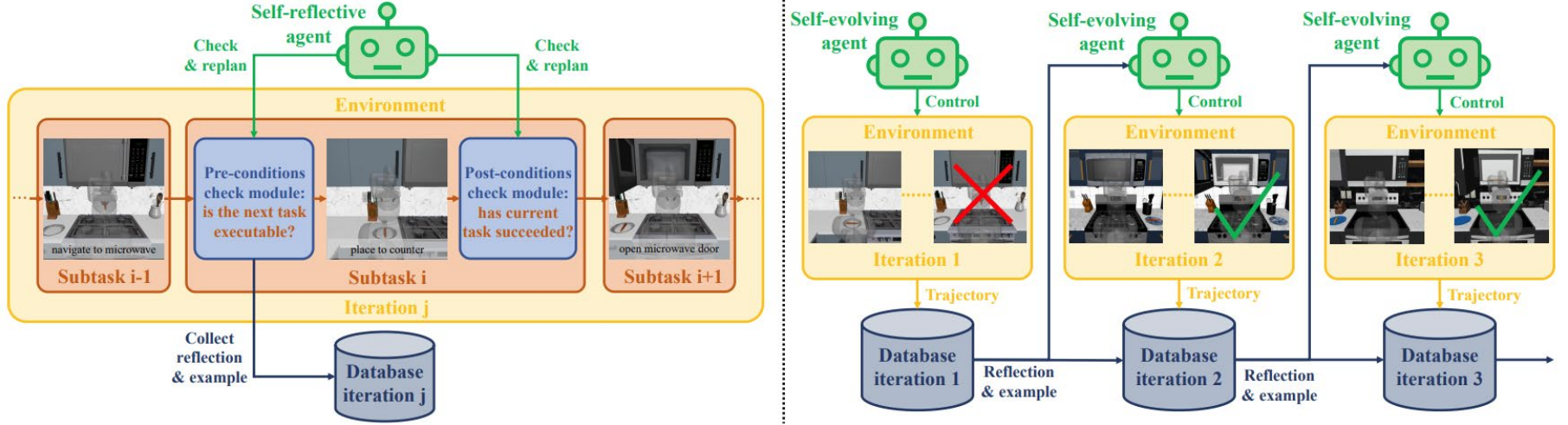}
\caption{An illustration of self-evolving multi-agent learning~\cite{Yuan2025MADynamicLearning1}.}
\label{fig:ma_evol}
\end{figure*}

\paragraph{Self-evolution in open environments.} 
Unlike well-defined simulated scenarios, real-world embodied tasks typically take place in open environments, where key elements such as task objectives, environmental factors (e.g., states, actions, and reward functions), and collaboration patterns (e.g., teammates and opponents) evolve continuously and unpredictably~\cite{Yuan2023OpenSurvey}. To address these challenges in MARL, researchers have proposed methods such as robust training~\cite{yuan2024robust,yuan2023robust} and continual coordination~\cite{yuan2024multiagent}, which have yielded promising results. However, the unpredictable nature of open environments in embodied scenarios presents even greater difficulties, requiring specialized approaches beyond conventional MARL techniques. Recently, several innovative solutions leveraging the strong generalization capabilities of generative models have emerged. For example, when the number of collaborators dynamically changes, scalable architectures like graph neural networks (GNNs) and transformers can effectively encode interaction information, as shown in Figure~\ref{fig:ma_evol}. By combining these architectures with distributed policy networks, agents can adapt seamlessly to fluctuations in team size, ensuring robust coordination~\cite{Cai2024MAchangescale1}. Moreover, by incorporating self-play and other mechanisms for self-reflection and policy evolution, MAS can continuously improve collaborative performance in dynamic and open environments~\cite{Gao2023MASelfPlayLearning1,Yuan2025MADynamicLearning1}.

\subsection{Generative Models based Multi-agent Interaction} \label{subsec:MAWM} 
Although algorithms designed for asynchronous decision-making, heterogeneous agents and open environments have advanced learning-based methods in embodied MAS, challenges persist due to the diversity of cooperative behaviors and the partial or missing nature of observations, leading to low exploration efficiency and increased learning complexity. To address these issues, generative models have emerged as a powerful tool for enhancing decision-making in embodied MAS. They can introduce prior knowledge to facilitate explicit task allocation and leverage their strong information-processing capabilities to enable inter-agent communication and observation completion for distributed decision-making. Furthermore, MAS may not only involve autonomous agents but also humans. By harnessing the language understanding and generation capabilities of generative models, human-agent interaction and collaboration can be significantly improved, marking a unique and critical application in this domain. Lastly, the significantly larger exploration space in multi-agent settings makes the challenge of sample efficiency even more pronounced than in single-agent scenarios. Consequently, data augmentation techniques based on generative models are essential for improving data efficiency in multi-agent embodied AI.

\begin{figure*}[h]
\centering
\includegraphics[width=.95\columnwidth]{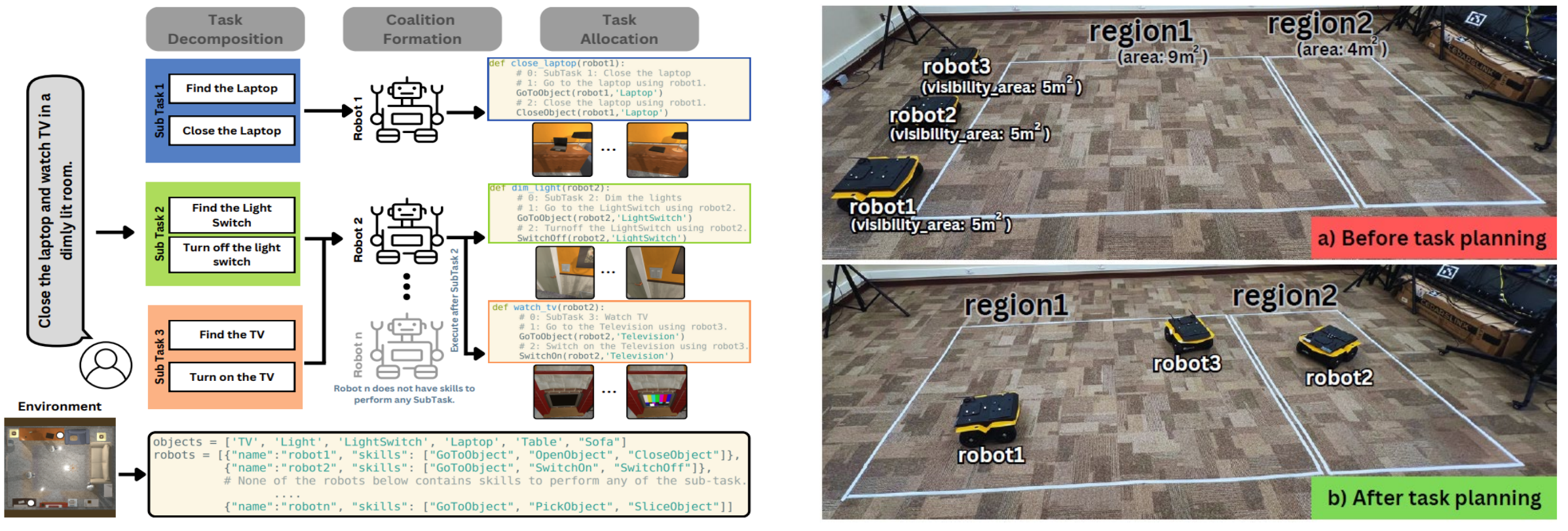}
\caption{An illustration of generative models based multi-agent task allocation~\cite{Kannan2024MALLMPlanning2}.}
\label{fig:ma_gen1}
\end{figure*}

\paragraph{Multi-agent task allocation.} 
To address the challenges posed by the diverse collaborative behaviors in embodied MAS, an increasing number of studies have begun leveraging the prior knowledge and reasoning capabilities of pre-trained generative models to explicitly assign distinct tasks to different agents, significantly reducing the individual exploration spaces of each agent. For example, SMART-LLM~\cite{Kannan2024MALLMPlanning2} utilizes a pre-trained language model to first decompose a given task into multiple parallel sub-tasks and then group agents based on their capabilities. The sub-tasks are subsequently assigned to each group accordingly, enabling multi-agent decentralized task executions. For instance, different robotic vacuum cleaners are each allocated a distinct cleaning area, as illustrated in Figure~\ref{fig:ma_gen1}. This paradigm of task decomposition and assignment has become a mainstream planning strategy in embodied MAS~\cite{Singh2024MALLMPlanning3,Chen2024MALLMPlanning4,Wan2025MALLMPlanning5}.
Building on this, a series of works integrates task assignment and execution. These approaches first use generative models for task allocation, then continue using the same or similar models to execute the tasks, adjusting the allocation based on execution outcomes to form a more complete workflow~\cite{Venkatesh2024MALLMPlanningandExecuting1,Liu2024MALLMPlanningandExecuting2,Wu2024MALLMPlanningandExecuting3,Chen2025MALLMPlanningandExecuting4}.
However, the aforementioned methods generally focus on task assignment based solely on generative reasoning, often overlooking the dependencies among sub-tasks. For instance, in the task ``retrieve the wrench from a closed box'', the sub-task ``open the box'' must be completed before ``take the wrench out of the box''. To better capture such dependencies, recent works have been focused on exploring the use of sub-task dependency graphs to enhance task allocation~\cite{Wang2024MALLMDependencyPlanning1,Obata2025MALLMDependencyPlanning2}.
In addition to explicit task decomposition, another line of work employs centralized generative models to produce global decisions. These are then implicitly distributed across agents by having each agent imitate the behavior generated by the centralized model, thus achieving implicit task assignment~\cite{Liu2025MALLMCentralizedTeacher}.

\begin{figure*}
\centering
\includegraphics[width=.95\columnwidth]{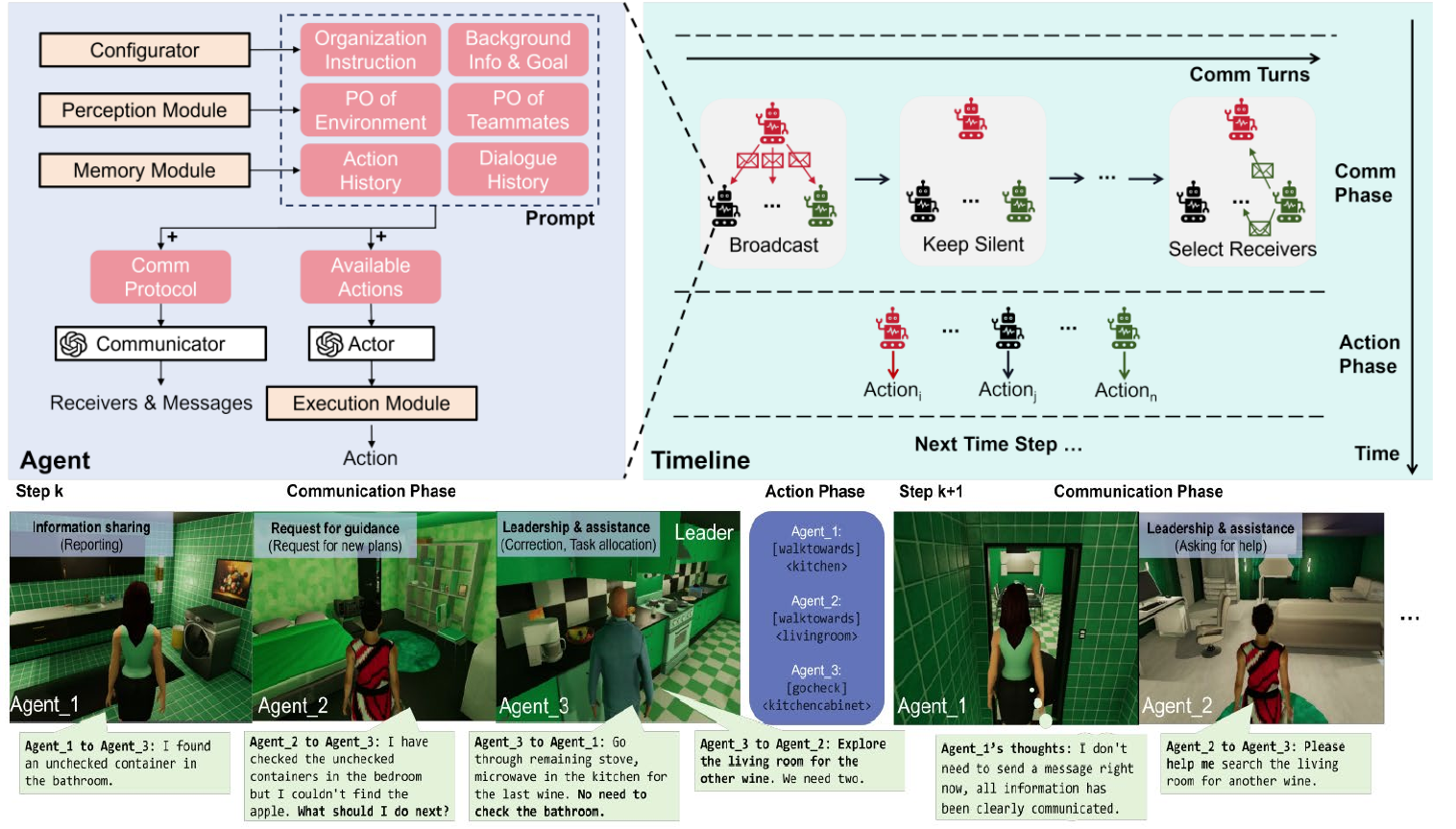}
\caption{An illustration of generative models based multi-agent distributed collaboration~\cite{Guo2024MALLMDecentralizedCollaboration4}.}
\label{fig:ma_gen2}
\end{figure*}

\paragraph{Multi-agent distributed decision making.} 
Using generative models for multi-agent embodied task decomposition and planning leverages the prior knowledge and reasoning capabilities acquired during pretraining to meet the demands of real-world collaborative tasks. However, centralized task planning and allocation can compromise the flexibility and scalability of cooperation, often requiring frequent recalls for adjustment. 
Therefore, exploring the coordination among multiple generative models is essential. 
Unlike independent methods in MARL that often fail in complex tasks, the outstanding perception and reasoning capabilities of generative models make it feasible to deploy multiple LLM-based agents that independently perform decision-making and policy evaluation effectively \cite{Zhang2024MALLMDecentralizedCollabration1,Wang2025MALLMDecentralizedCollabration2,Bo2025MALLMDecentralizedCollabration3,Guo2024MALLMDecentralizedCollaboration4,Tan2020MALLMDecentralizedCollaboration5,Wang2024MALLMDecentralizedCollaboration6,Yu2024MALLMDecentralizedCollaboration7}.
For example, \cite{Guo2024MALLMDecentralizedCollaboration4} equips each agent with a unique pre-trained generative model. These models leverage their strong information processing capabilities to communicate with other agents, either to complete missing observations or to request assistance, thus facilitating decentralized collaboration, as illustrated in Figure~\ref{fig:ma_gen2}.
However, challenges such as credit assignment and policy conflicts still hinder the effectiveness of fully distributed collaborative architectures. 
Therefore, similar to practices in MARL, some studies have begun to incorporate an additional centralized generative model to evaluate the decisions made by distributed generative models, enhancing their overall decision-making capabilities~\cite{Zhang2024MALLMDecentralizedCollabration1,Guo2024MALLMDecentralizedCollaboration4}.
Furthermore, by introducing a shared global LLM-based reflector to assess the contribution of each individual during collaboration, effective credit assignment in multi-LLM cooperation can also be achieved. \cite{Bo2025MALLMDecentralizedCollabration3}. 
With appropriately designed system topologies and hierarchical collaboration frameworks, generative model-based distributed decision making is capable of scaling to large-scale systems including up to thousands of agents \cite{Zhang2024MALLMLargeScaleDecentralizedCollabration1,Qian2025MALLMLargeScaleDecentralizedCollabration2}.

Beyond classic centralized or decentralized frameworks for embodied collaboration, generative models introduce a unique advantage: they allow negotiation among agents \cite{Liu2023MALLMDecentralizedConsensusCollabration1,Chen2023MALLMDecentralizedConsensusCollabration2,Guo2024MALLMDecentralizedCollaboration4}. 
Unlike traditional collaboration pipelines where agents plan once every step, multiple LLMs are capable of engaging in iterative negotiation rounds during each planning phase, aiming to collectively identify an optimal course of action \cite{Chen2023MALLMDecentralizedConsensusCollabration2}. 
Beyond conventional coordination, multi-LLM negotiation enables embodied MAS to dynamically select and evolve their membership, organizational frameworks and leadership roles, thereby enhancing their capacity to handle complex tasks in open environments \cite{Liu2023MALLMDecentralizedConsensusCollabration1,Guo2024MALLMDecentralizedCollaboration4}. 

\begin{figure*}
\centering
\includegraphics[width=.95\columnwidth]{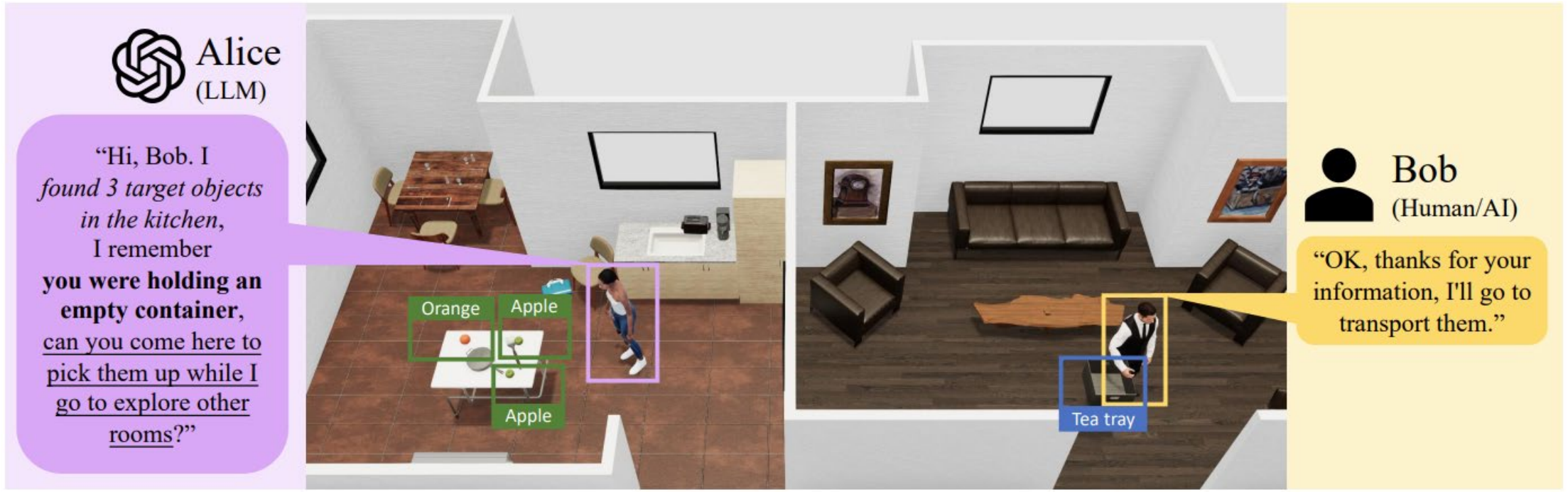}
\caption{An illustration of generative models based human-AI collaboration~\cite{Zhang2024MALLMHumanAgentCollaboration1}.}
\label{fig:ma_gen3}
\end{figure*}

\paragraph{Human-AI coordination.} 
Efficient collaboration between humans and embodied agents has long been a critical objective of research~\cite{ajoudani2018progress}. Human-AI coordination~\cite{vicentini2021collaborative}, closely related to research fields such as human-AI interaction (HAI)~\cite{van2021human} and human-robot interaction (HRI)~\cite{onnasch2021taxonomy}, focuses on enhancing teamwork between humans and robots to effectively achieve complex tasks. Cooperative MARL, recognized for its robust problem-solving capabilities, provides promising avenues for improving human-AI collaboration across diverse user groups. However, traditional RL approaches often fall short in fully capturing the intricacies and variability inherent in human behaviors. Recently, with the rise of multi-modal large models such as LLMs which inherently involve human-in-the-loop training processes, researchers have begun leveraging the extensive knowledge embedded in them to design sophisticated and adaptive strategies for human-AI collaboration~\cite{Zhang2024MALLMHumanAgentCollaboration1,Sun2024MALLMHumanAgentCollaboration2,Guo2024MALLMDecentralizedCollaboration4,Feng2024MALLMHumanAgentCollaboration3,guan2023efficient,Liu2024MALLMHumanAgentCollaboration6,Asuzu2025MALLMHumanAgentCollaboration7,Liu2024MALLMHumanAgentCollaboration8,Gao2024MALLMHumanAgentCollaboration10} as illustrated in Figure~\ref{fig:ma_gen3}.

In scenarios including multiple LLM-based agents, collaboration between humans and agents can be partially facilitated by substituting human participants for communication partners or team leaders~\cite{Zhang2024MALLMHumanAgentCollaboration1,Sun2024MALLMHumanAgentCollaboration2,Guo2024MALLMDecentralizedCollaboration4}. Nevertheless, this approach often under-utilizes the strengths that humans bring to the collaboration process. Given humans' superior capacities for understanding, reasoning, and adapting, recent methods have emphasized active human involvement to address limitations inherent in LLM-based agents~\cite{Feng2024MALLMHumanAgentCollaboration3}. For example, during navigation tasks, agents encountering uncertainty can proactively seek missing perceptual or decisional information from humans through natural language queries, allowing humans to offer guidance without directly engaging in detailed task planning or execution~\cite{Liu2024MALLMHumanAgentCollaboration4}. Additionally, LLM-based agents can proactively infer human intentions based on verbal and behavioral cues, facilitating more intuitive, flexible, and effective human-AI collaboration without explicit communication requests~\cite{Wan2025MALLMHumanAgentCollaboration5}. Furthermore, these agents can autonomously adapt and refine their collaborative behaviors over time, evolving through interactions without the need for explicit human instructions, thereby supporting sustained and effective long-term human-AI collaboration~\cite{Liu2024MALLMHumanAgentCollaboration6}.

\begin{figure*}
\centering
\includegraphics[width=.98\columnwidth]{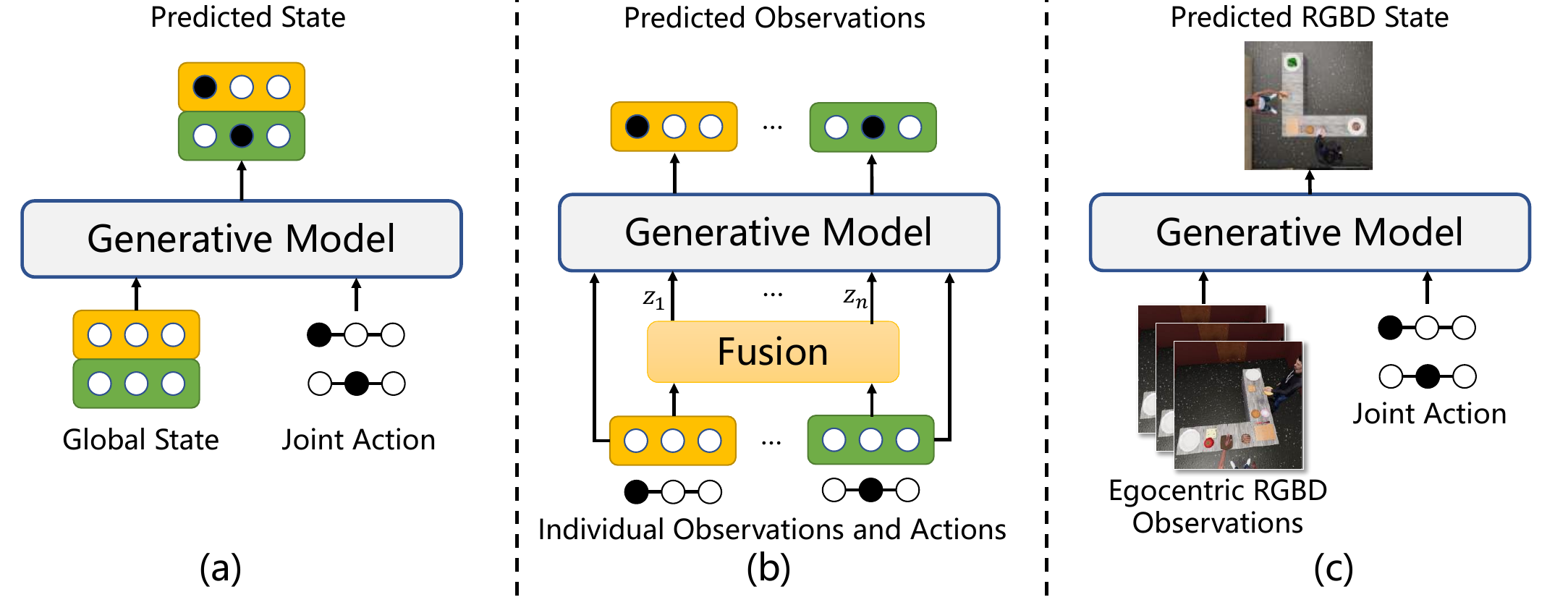}
\caption{An illustration of different multi-agent world model structures.}
\label{fig:ma_gen4}
\end{figure*}

\paragraph{Data-efficient multi-agent learning.} 
Due to the high sample efficiency of model-based RL approaches, applying world models to enhance multi-agent collaborative learning has always been a significant topic. However, modeling the interactions between agents and inferring the global state from local observations remains a key difficulty in leveraging world models to improve multi-agent collaboration efficiency. Early work attempted to model the dynamics of two collaborative robots simultaneously, achieving some success \cite{Krupnik2020MAWM1}, as shown in Figure~\ref{fig:ma_gen4}(a). However, as the number of collaborative agents increases, this approach quickly becomes inefficient.
Through the use of generative models such as VAE and Transformers for local observation fusion, decoupling of global and local modeling and auto-regressive trajectory prediction, it is now possible to model the dynamic collaboration of MAS using world models \cite{Shi2025MAWM2,Venugopal2024MAWM3,Barde2024MAWM4,Liu2024MAWM5}, as shown in Figure~\ref{fig:ma_gen4}(b).
One approach addresses these challenges in the context of urban traffic by constructing a generative world model that simulates large-scale heterogeneous agents under physical and social constraints, using destination- and personality-conditioned dynamics to promote realistic and safe multi-agent behaviors \cite{ZhangMAWM6}.
Complementarily, another line of work focuses on physically embodied agents operating with egocentric RGBD observations, introducing a diffusion-based reconstruction process to establish a shared representation of the environment and incorporating intention inference to facilitate implicit coordination without direct communication \cite{Zhang2025MAWM7}, as shown in Figure~\ref{fig:ma_gen4}(c).

\subsection{Benchmarks} \label{subsec:MABM} 
\begin{figure*}[ht]
	\centering    
	\subfloat[Grid-MAPF \cite{Stern2019MABench17}]{
		\label{fig:Grid-MAPF}     
		\includegraphics[width=0.24\columnwidth]{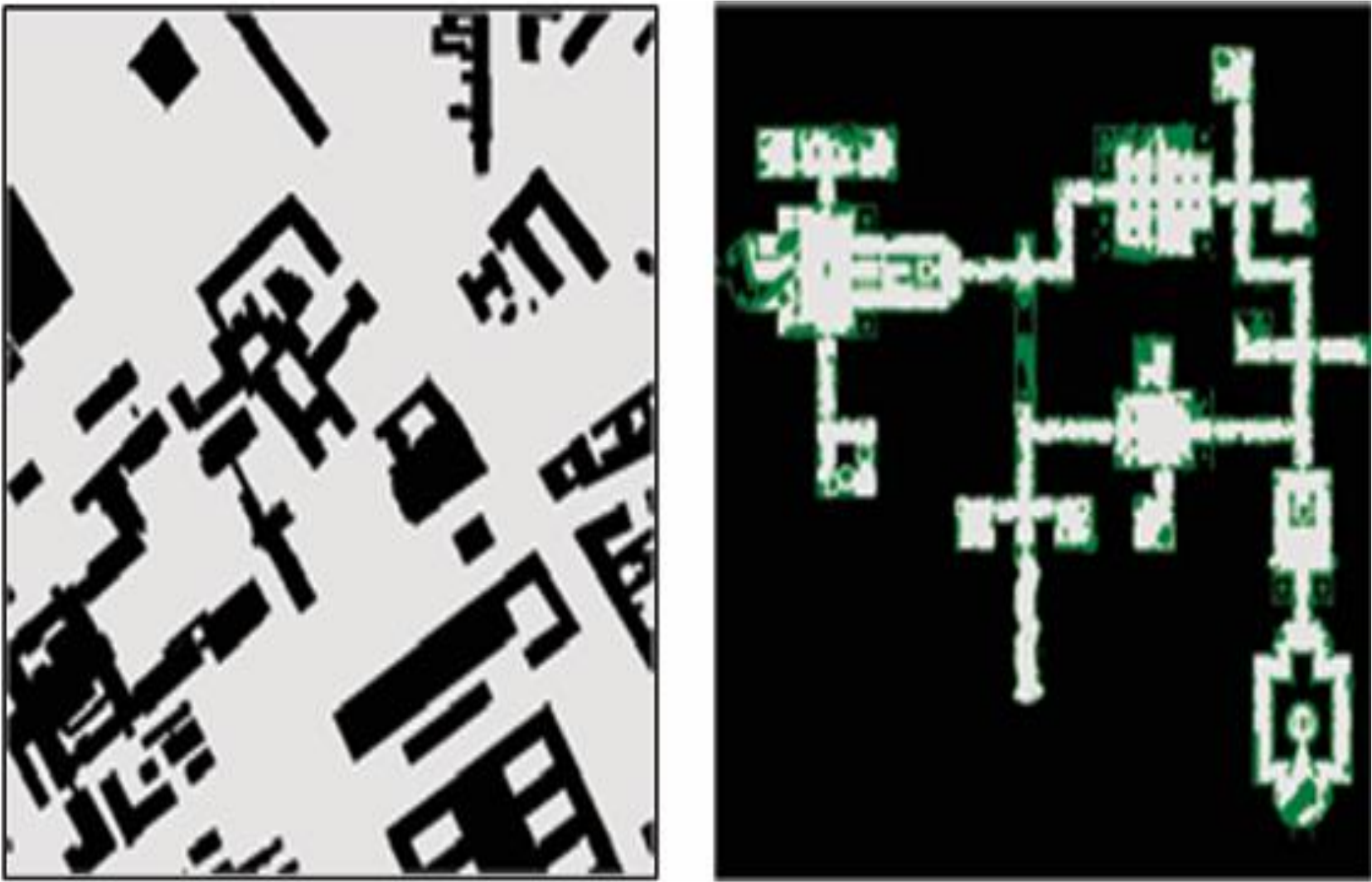}  
	}
        \subfloat[FurnMove \cite{Jain2020MABench19}]{ 
		\label{fig:FurnMove}     
		\includegraphics[width=0.24\columnwidth]{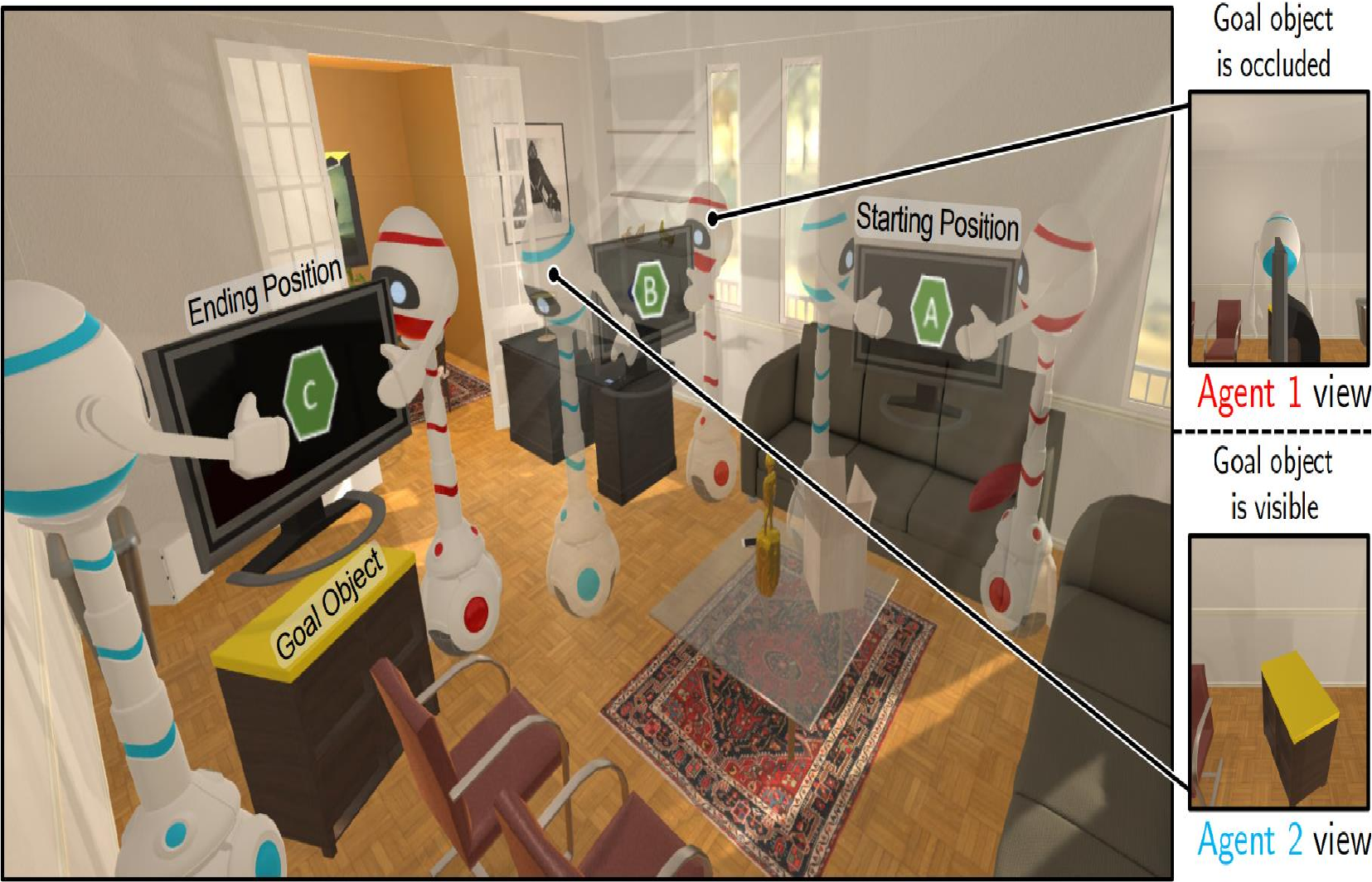}     
	}    
        \subfloat[SMARTS \cite{Zhou2021MABench20}]{
		\label{fig:SMARTS}     
		\includegraphics[width=0.24\columnwidth]{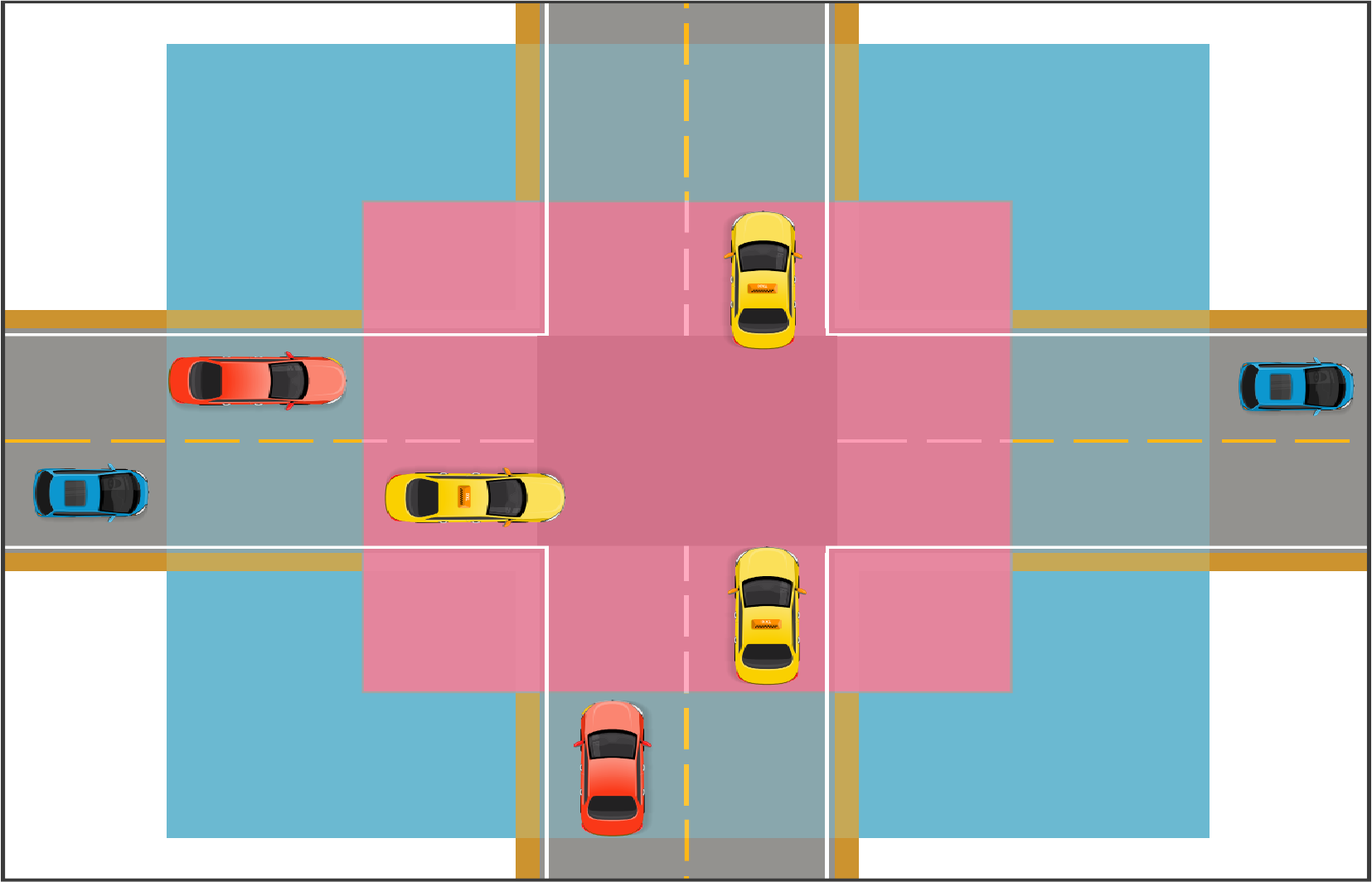}     
	}
	\subfloat[Megaverse \cite{Petrenko2021MABench1}]{ 
		\label{fig:Megaverse}     
		\includegraphics[width=0.24\columnwidth]{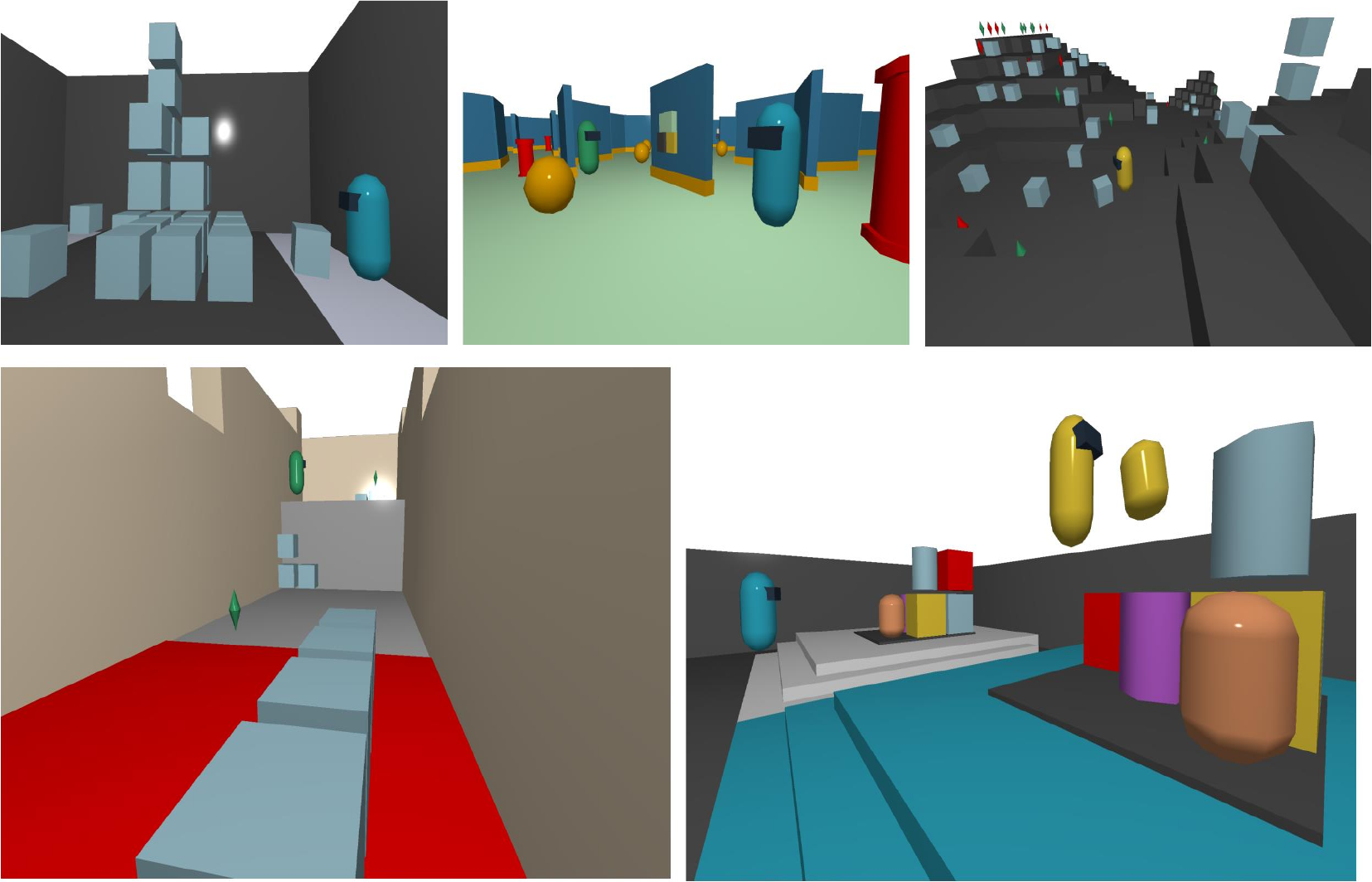}     
	}  
    \\
	\subfloat[V2X-Sim \cite{Li2022MABench18}]{ 
		\label{fig:V2X-Sim}     
		\includegraphics[width=0.24\columnwidth]{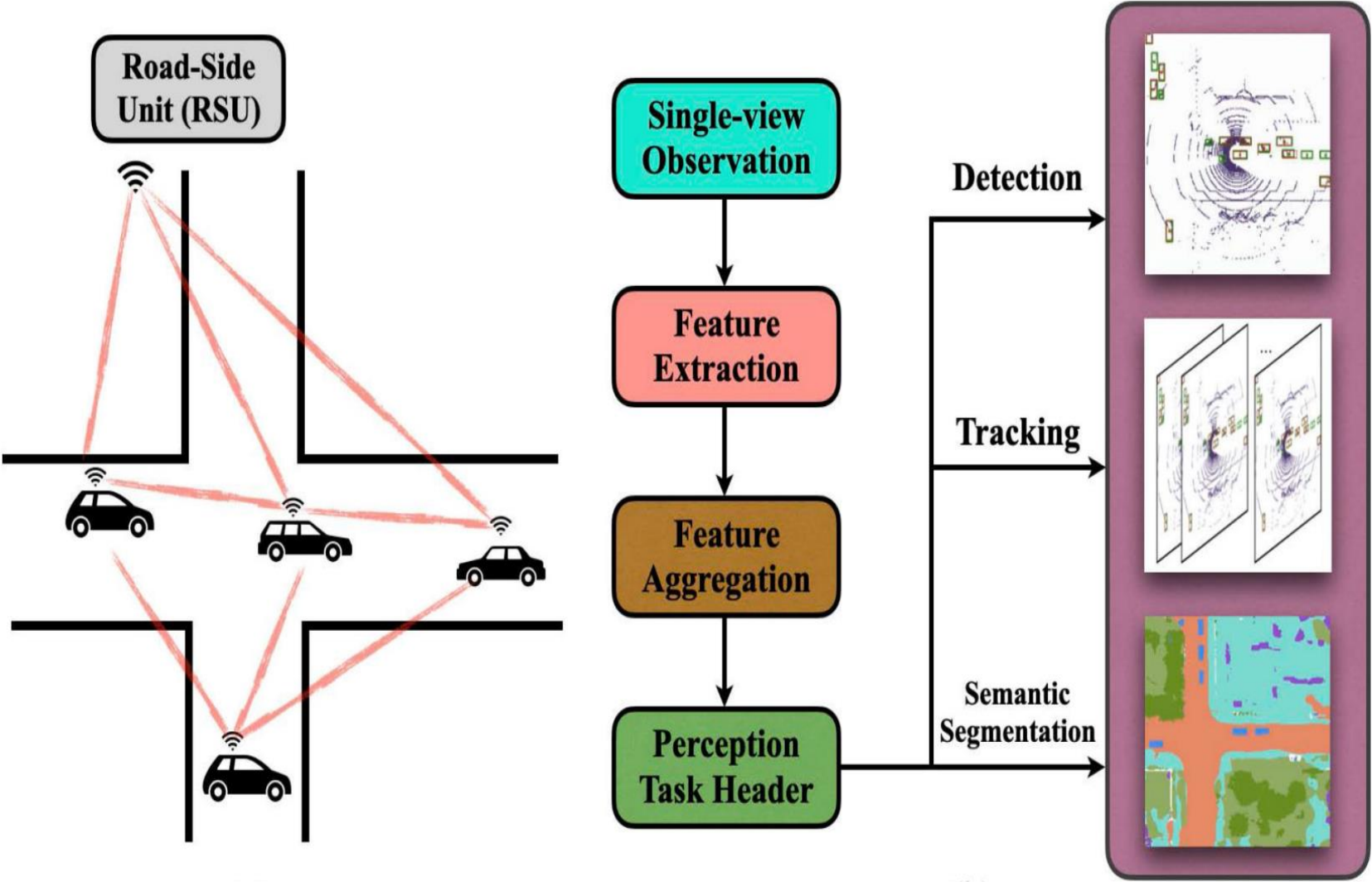}     
	}  
	\subfloat[BiDexhands \cite{Chen2022MABench2}]{ 
		\label{fig:BiDexhands}     
		\includegraphics[width=0.24\columnwidth]{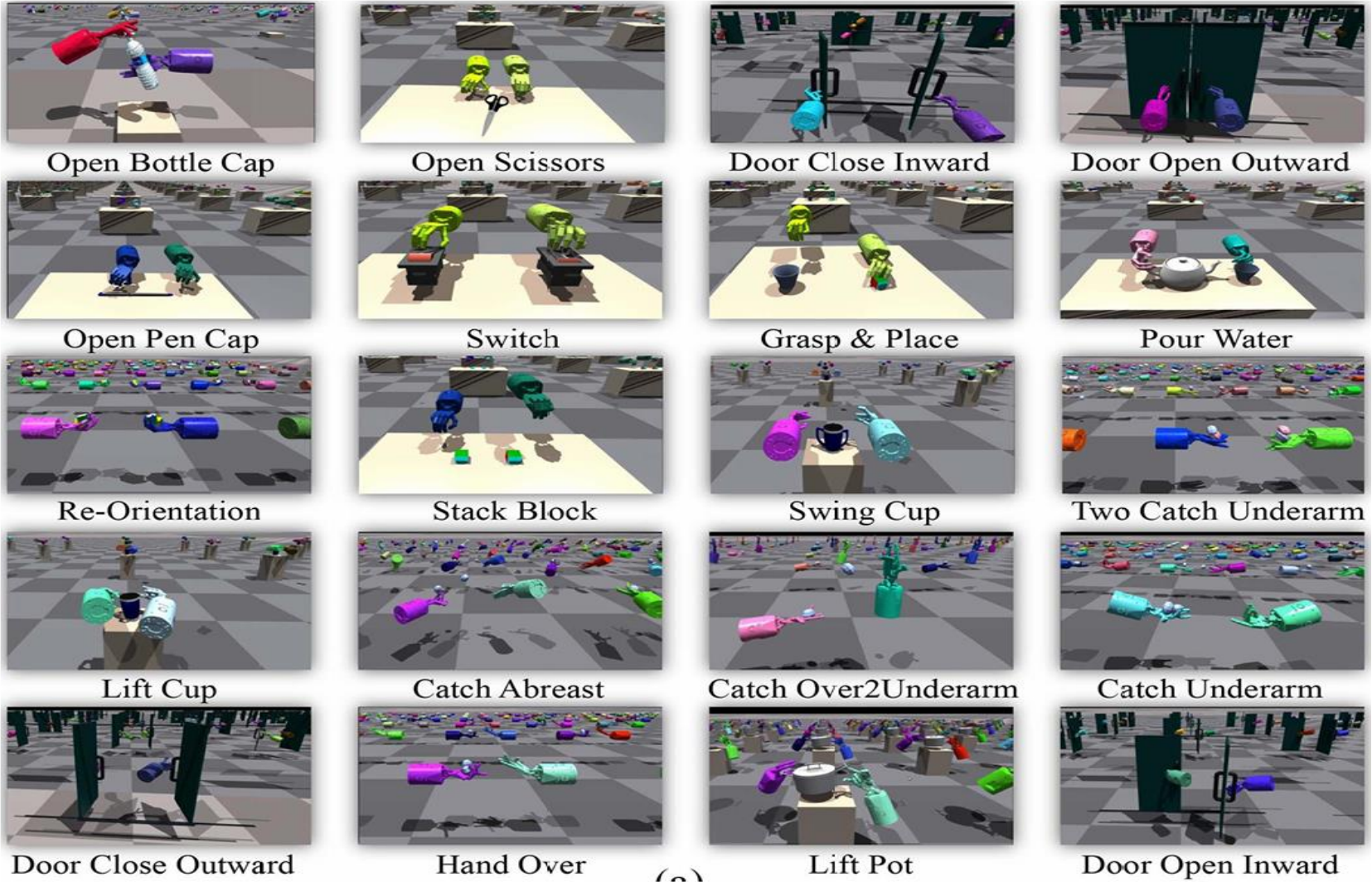}     
	}
	\subfloat[MRP-Bench \cite{Schaefer2023MABench3}]{ 
		\label{fig:MRP-Bench}     
		\includegraphics[width=0.24\columnwidth]{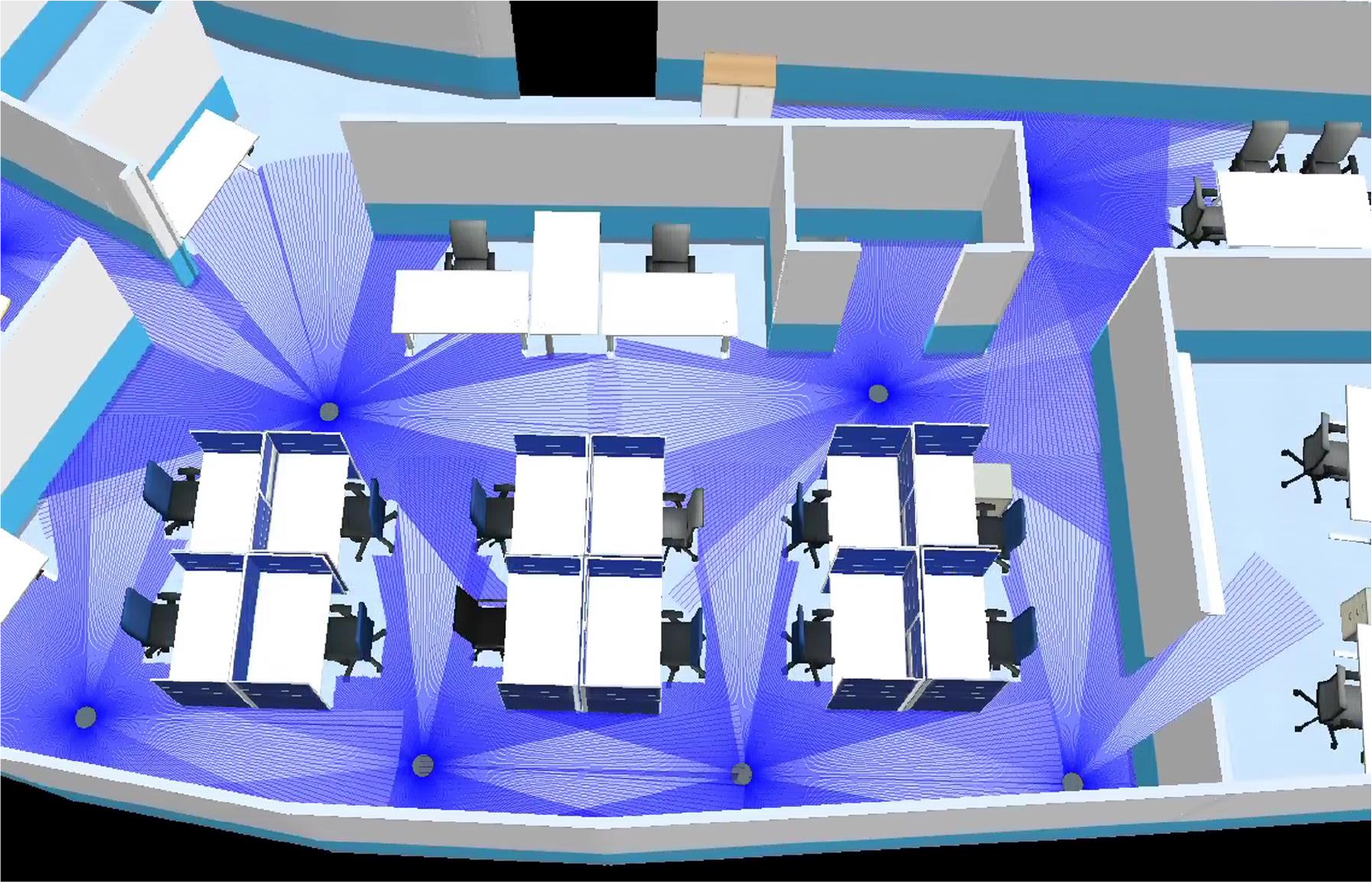}     
	}
	\subfloat[LEMMA \cite{Gong2023MABench4}]{ 
		\label{fig:LEMMA}     
		\includegraphics[width=0.24\columnwidth]{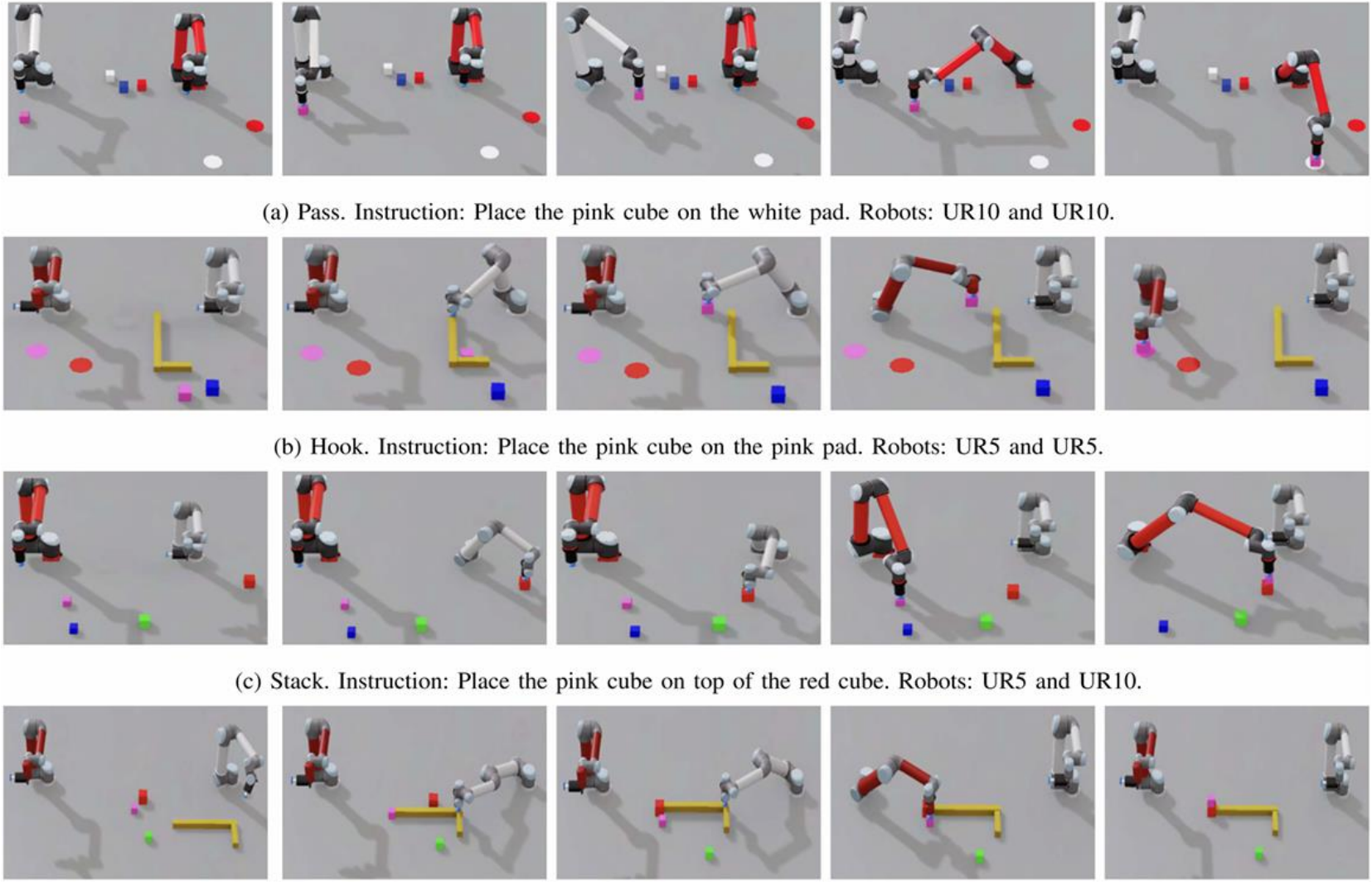}     
	}
    \\
	\subfloat[CHAIC \cite{Du2024MABench16}]{ 
		\label{fig:CHAIC}     
		\includegraphics[width=0.24\columnwidth]{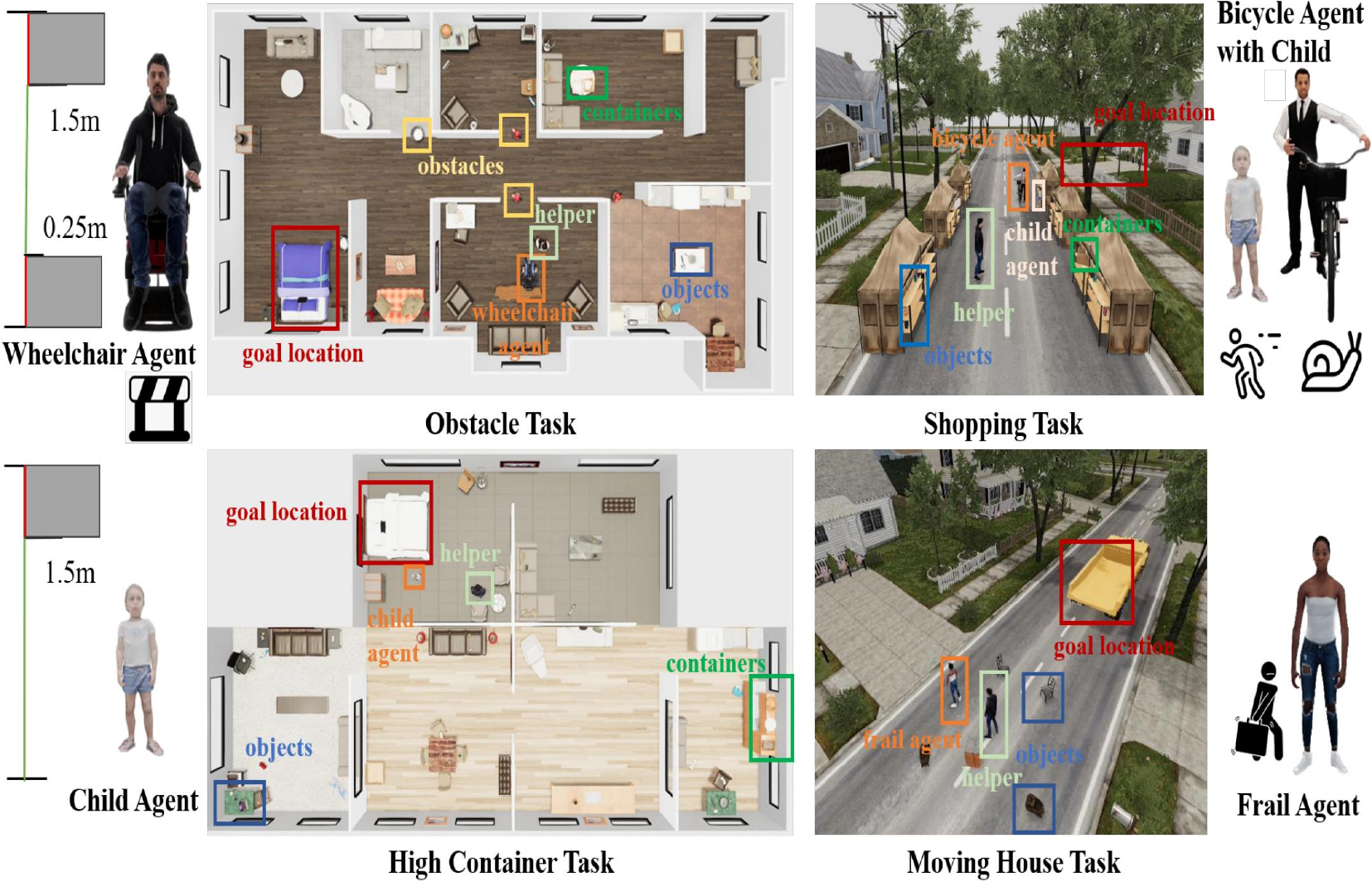}     
	}
	\subfloat[RoCo \cite{Mandi2024EnvReflection1}]{ 
		\label{fig:RoCo}     
		\includegraphics[width=0.24\columnwidth]{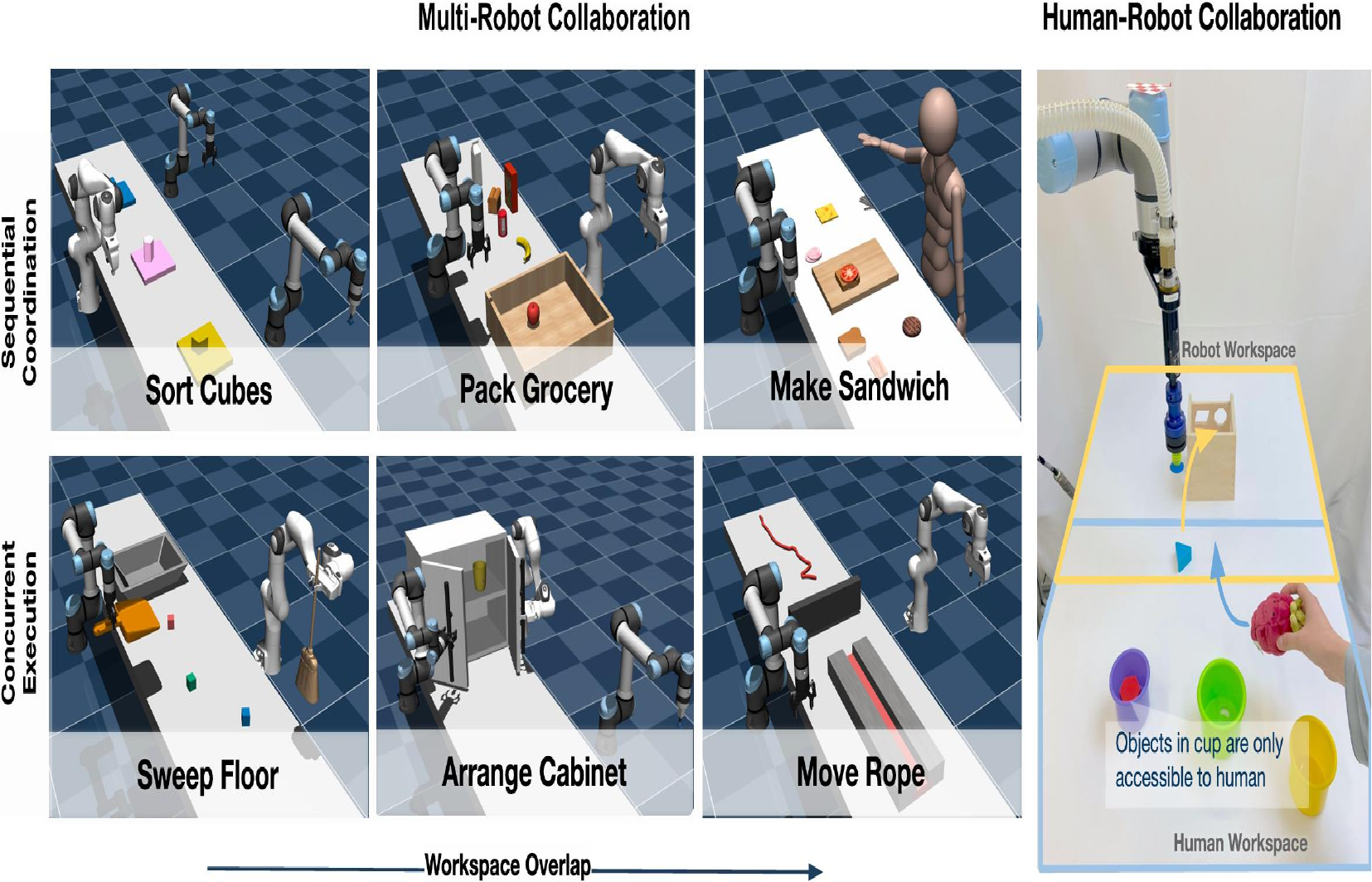}     
	}
	\subfloat[MAP-THOR \cite{Nayak2024MABench15}]{ 
		\label{fig:MAP-THOR}     
		\includegraphics[width=0.24\columnwidth]{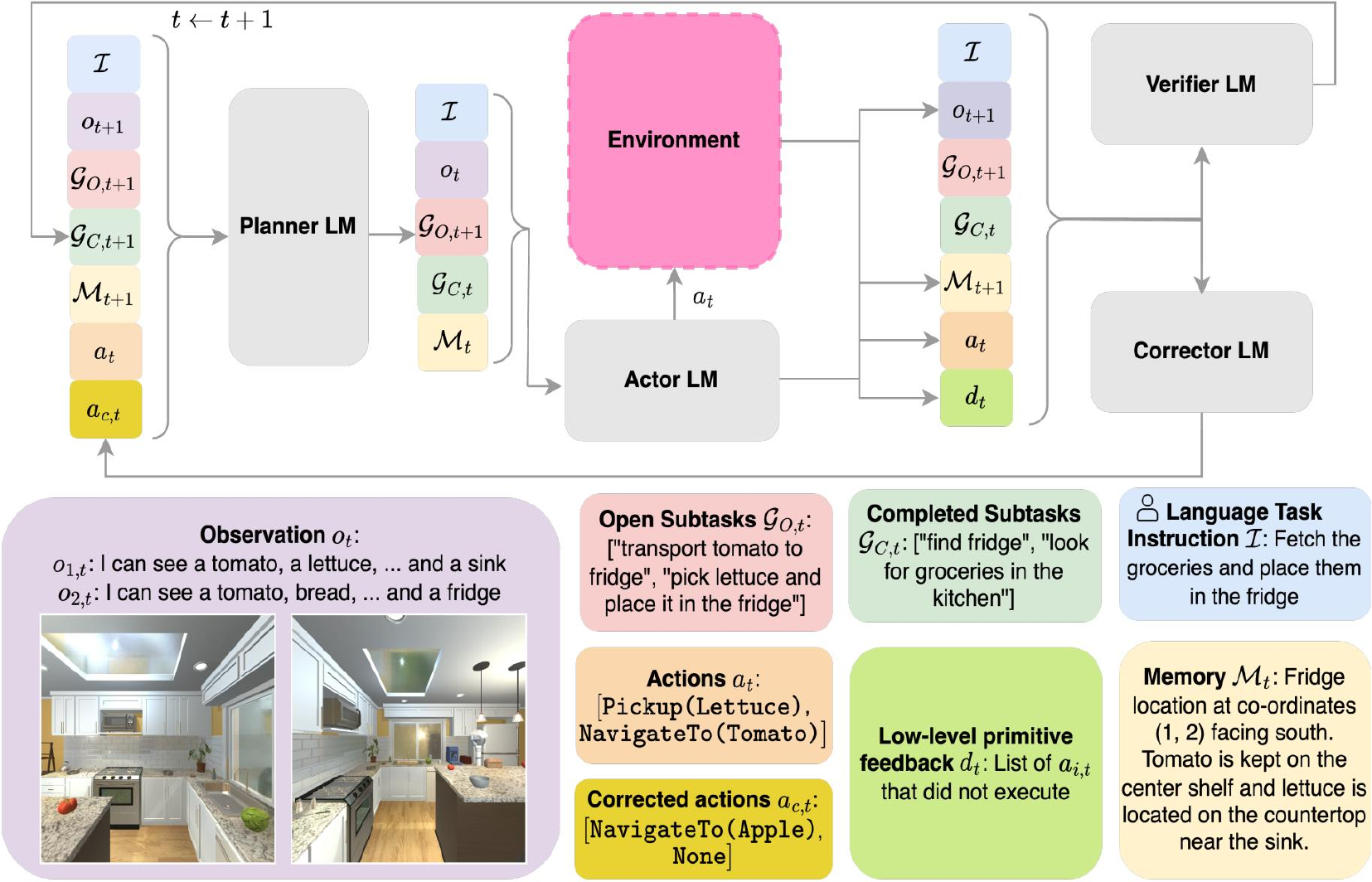}     
	}
	\subfloat[REMROC \cite{Heuer2024MABench6}]{ 
		\label{fig:REMROC}     
		\includegraphics[width=0.24\columnwidth]{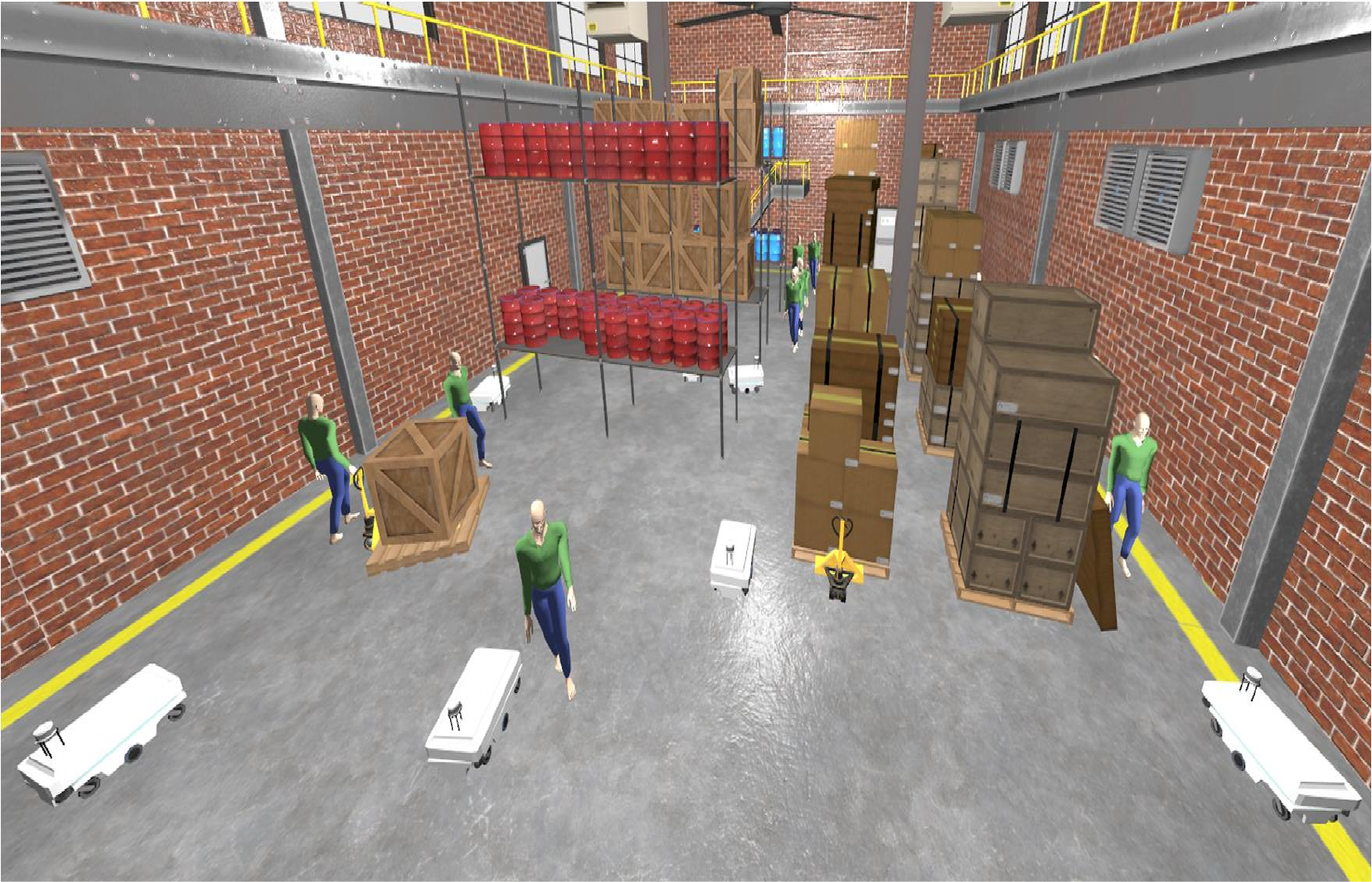}     
	}
    \\
	\subfloat[T2E \cite{Zhang2024MABench7}]{ 
		\label{fig:T2E}     
		\includegraphics[width=0.24\columnwidth]{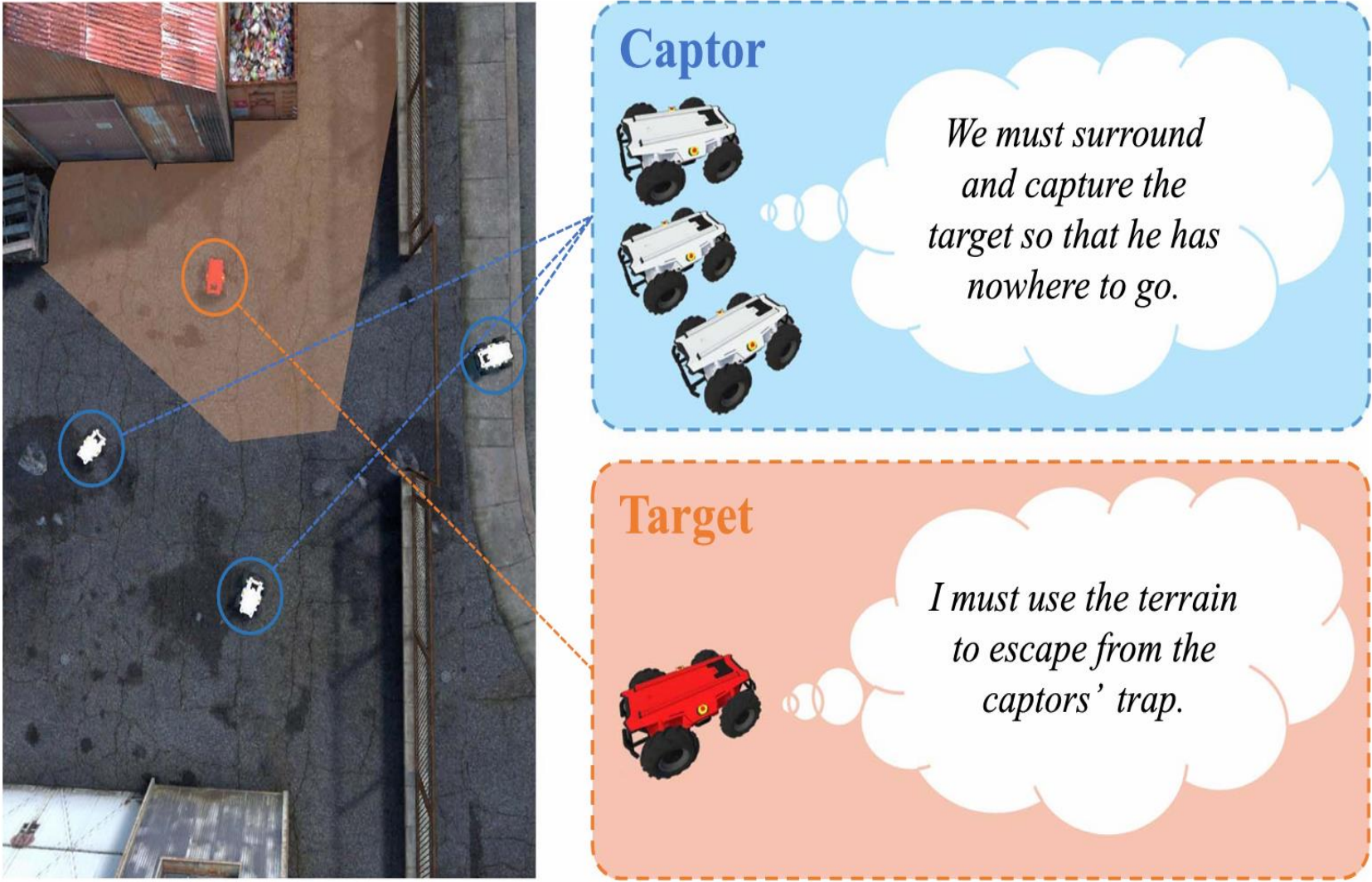}     
	}
	\subfloat[SocialGym2.0 \cite{Chandra2024MABench8}]{ 
		\label{fig:SocialGym2}     
		\includegraphics[width=0.24\columnwidth]{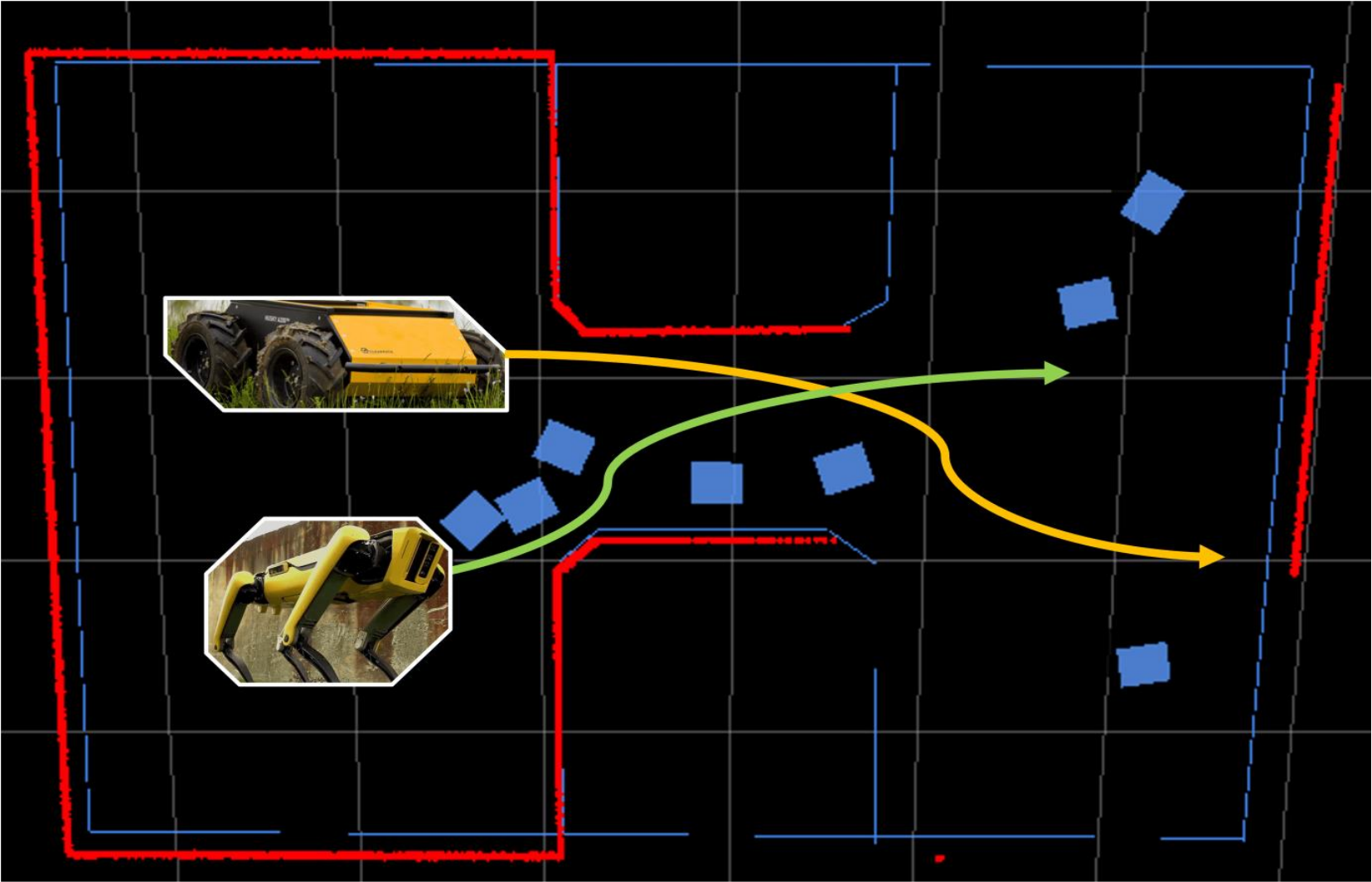}     
	}
	\subfloat[Cambrige RoboMaster \cite{Blumenkamp2024MABench9}]{ 
		\label{fig:CambrigeRoboMaster}     
		\includegraphics[width=0.24\columnwidth]{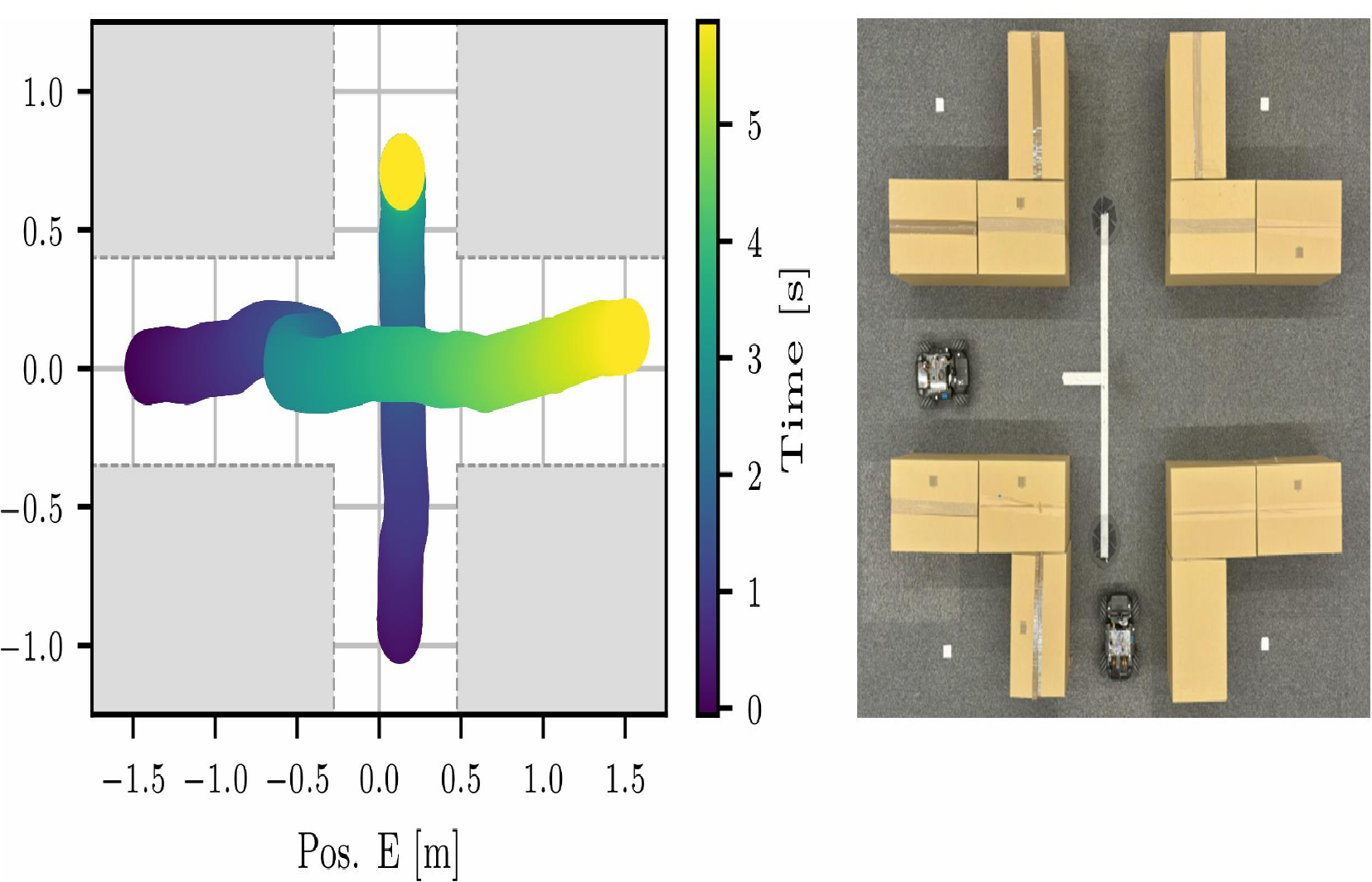}     
	}
	\subfloat[S3E \cite{Feng2024MABench10}]{ 
		\label{fig:S3E}     
		\includegraphics[width=0.24\columnwidth]{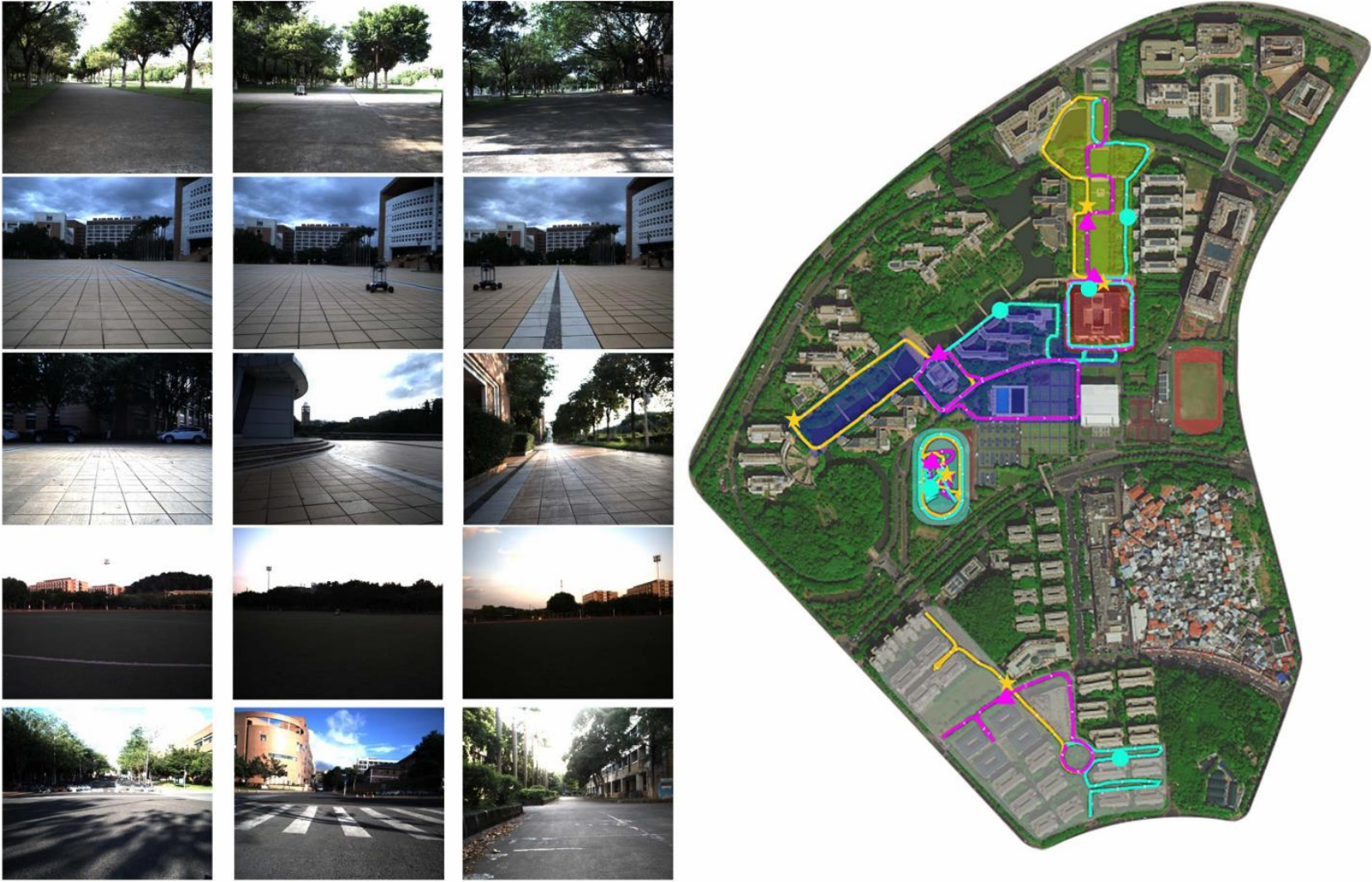}     
	}
    \\
	\subfloat[PARTNR \cite{Chang2025MABench11}]{ 
		\label{fig:PARTNR}     
		\includegraphics[width=0.24\columnwidth]{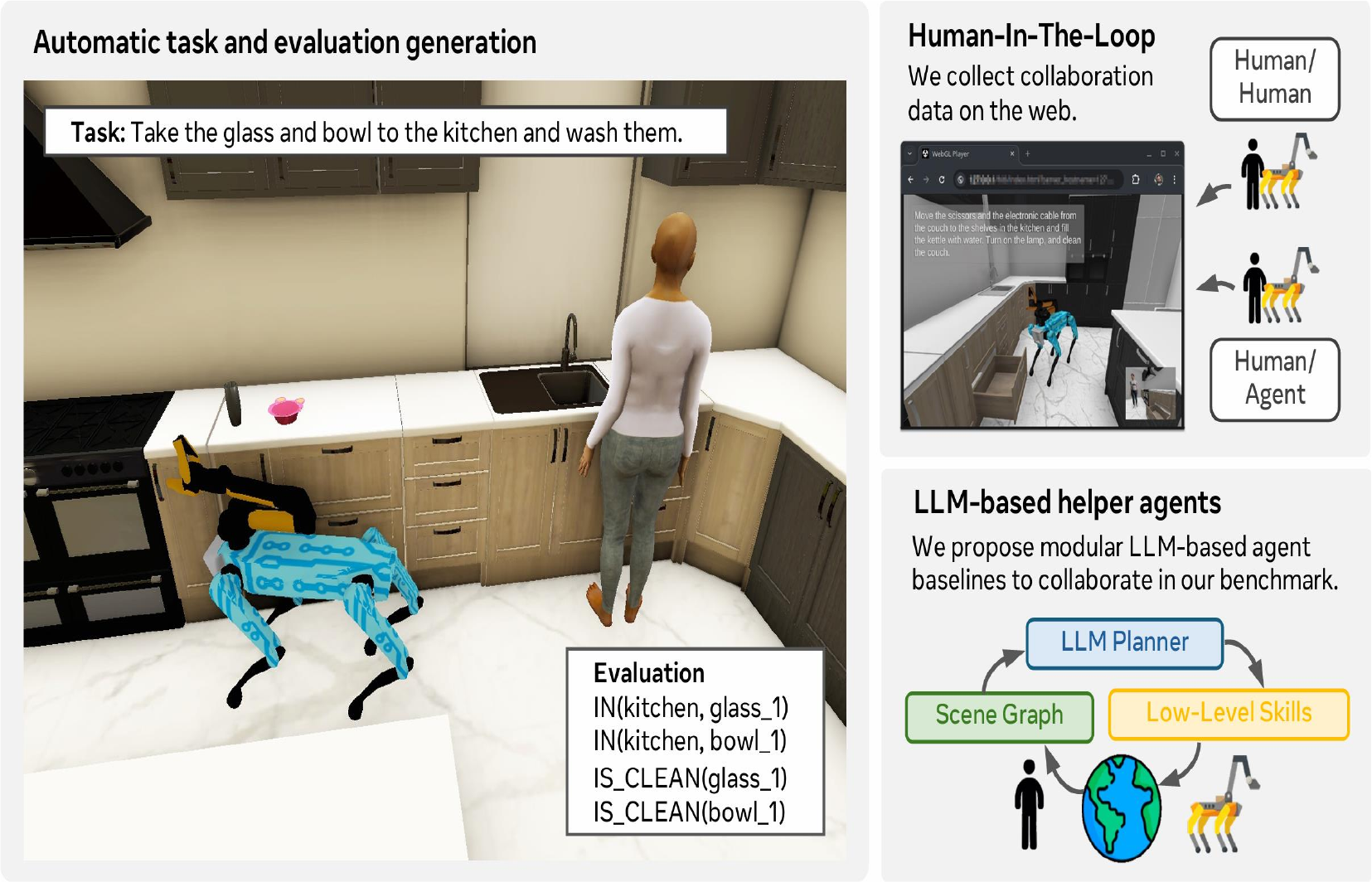}     
	}
	\subfloat[EmbodiedBench \cite{Yang2025MABench12}]{ 
		\label{fig:EmbodiedBench}     
		\includegraphics[width=0.24\columnwidth]{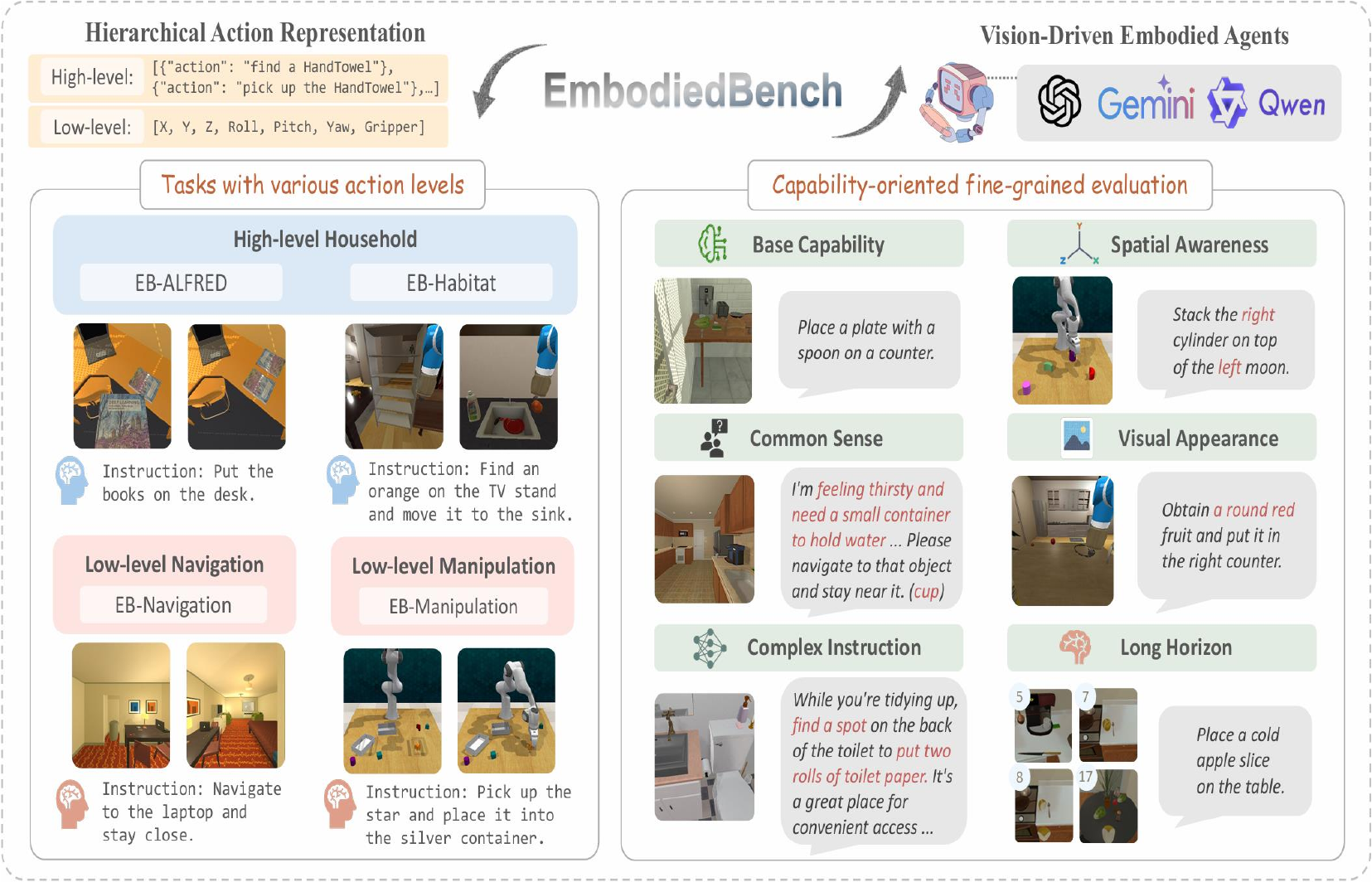}     
	}
	\subfloat[MRMG-Planning \cite{Hartmann2025MABench13}]{ 
		\label{fig:MRMG-Planning}     
		\includegraphics[width=0.24\columnwidth]{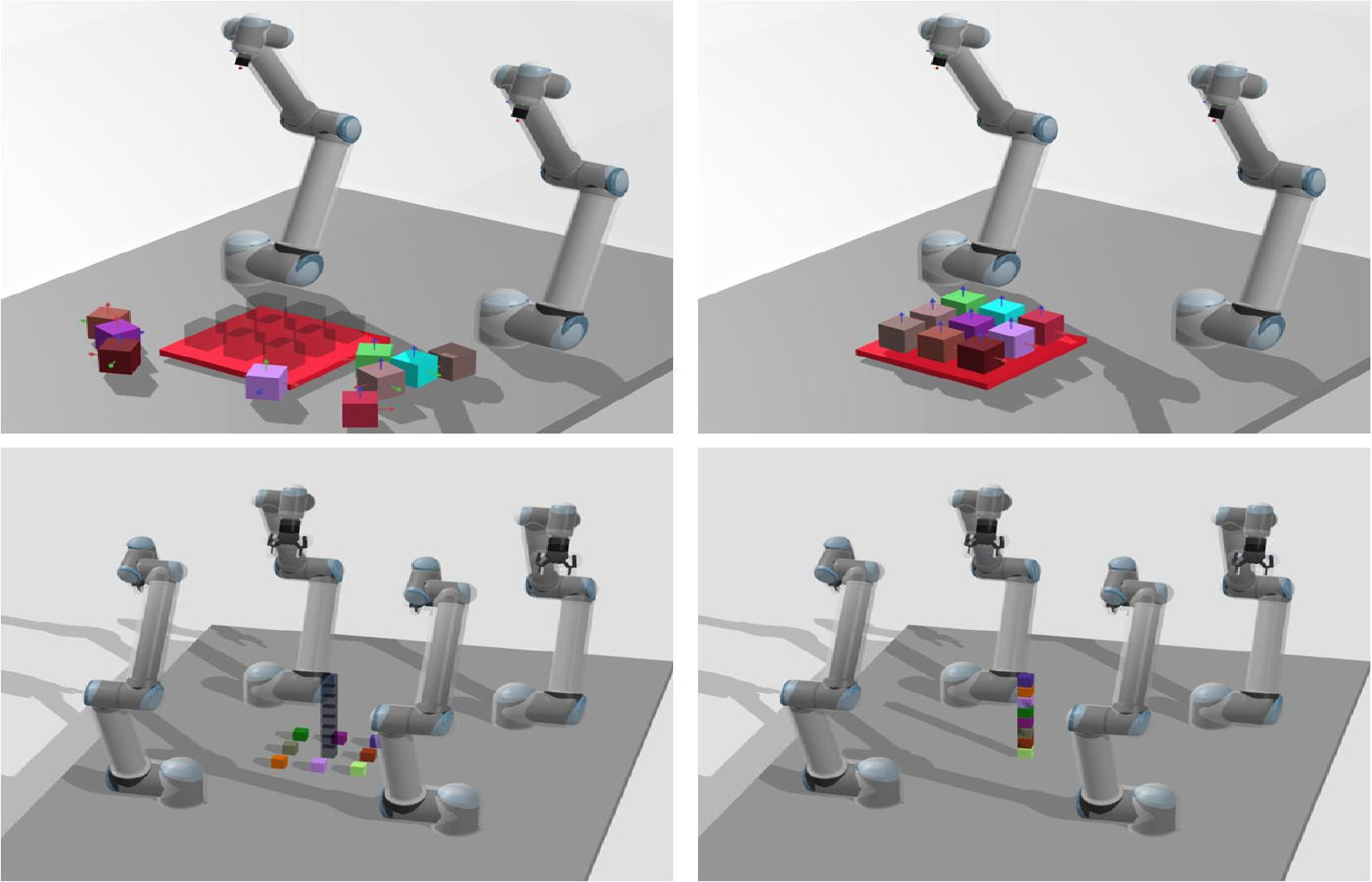}     
	}
	\subfloat[CARIC \cite{Cao2025MABench14}]{ 
		\label{fig:CARIC}     
		\includegraphics[width=0.24\columnwidth]{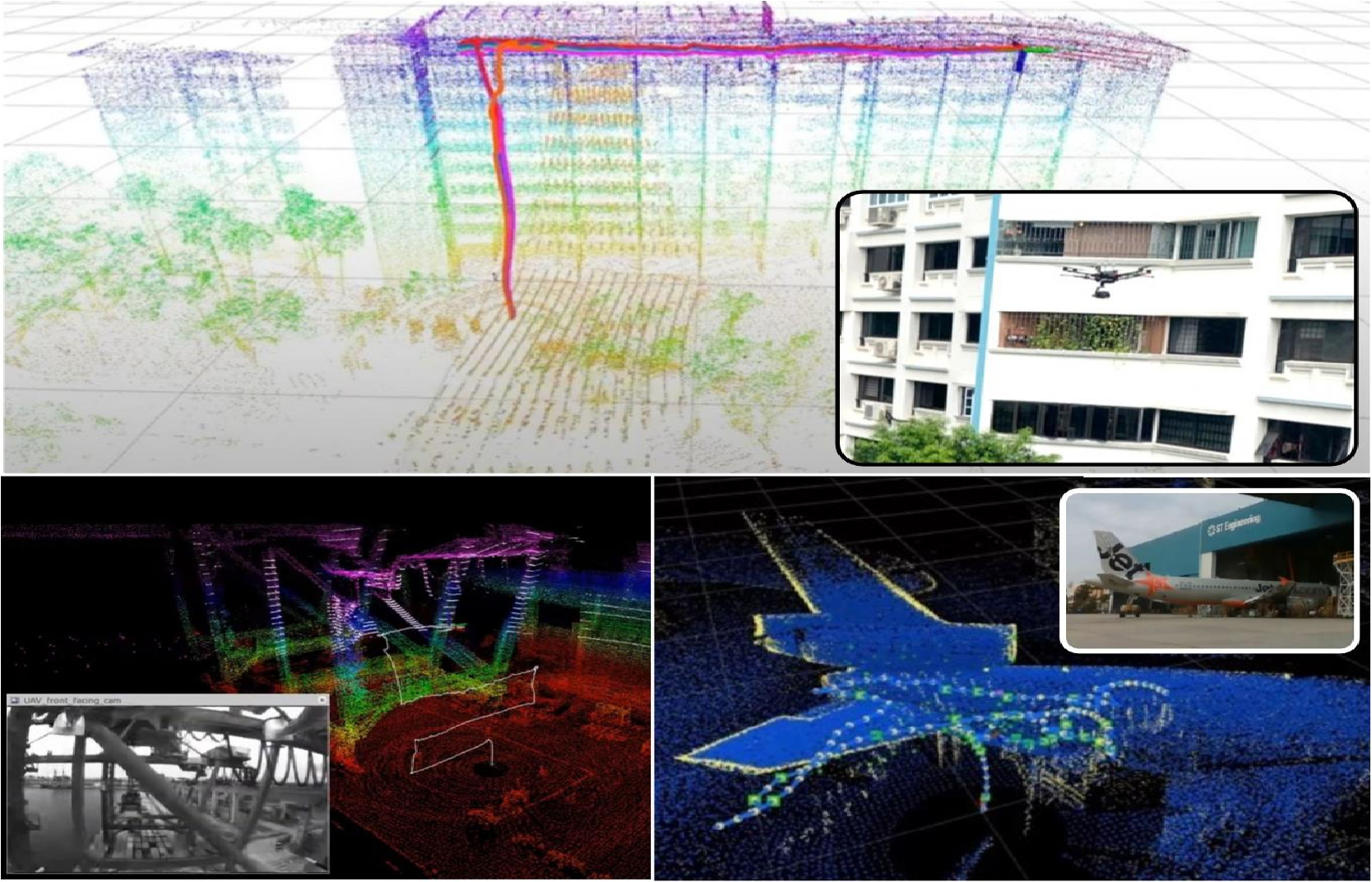}     
	}   
    \caption{An overview of multi-embodied benchmarks listed in Table \ref{tbl:MABenchmarks}.}
    \label{fig:MABenchmarks}
\end{figure*}

\begin{table*}[h]
\centering
\caption{Benchmarks for testing multi-agent embodied AI.}
\small
\label{tbl:MABenchmarks}
\resizebox{\textwidth}{!}{
\begin{tabular}{l|c|ccc|cccc|cc|ccc|ccccc}
\toprule
& \multicolumn{1}{c|}{\textbf{Year}} 
& \multicolumn{3}{c|}{\textbf{Category}} 
& \multicolumn{4}{c|}{\textbf{Input}} 
& \multicolumn{2}{c|}{\textbf{View}} 
& \multicolumn{3}{c|}{\textbf{Data}} 
& \multicolumn{5}{c}{\textbf{Agent Type}} 
\\
& \rotatebox{90}{\quad}
& \rotatebox{90}{Perception} & \rotatebox{90}{Planning} & \rotatebox{90}{Control}
& \rotatebox{90}{Image} & \rotatebox{90}{Pointcloud} & \rotatebox{90}{Language} & \rotatebox{90}{Proprioception}
& \rotatebox{90}{Global} & \rotatebox{90}{Local}
& \rotatebox{90}{Real} & \rotatebox{90}{Sim} & \rotatebox{90}{Hybrid}
& \rotatebox{90}{Humanoid/Human} & \rotatebox{90}{Quadruped} & \rotatebox{90}{UAV/UGV/USV} & \rotatebox{90}{Manipulator} & \rotatebox{90}{No certain type}
\\
\midrule
\href{https://movingai.com/benchmarks/mapf.html}{Grid-MAPF} \cite{Stern2019MABench17} & 2019 &  & \checkmark &  &  &  &  & \checkmark & \checkmark &  &  & \checkmark &  &  &  &  &  & \checkmark \\
\href{https://unnat.github.io/cordial-sync}{FurnMove} \cite{Jain2020MABench19} & 2020 & \checkmark & \checkmark &  & \checkmark &  & \checkmark &  &  & \checkmark &  & \checkmark &  &  &  & \checkmark & \checkmark &  \\
\href{https://github.com/huawei-noah/SMARTS}{SMARTS} \cite{Zhou2021MABench20} & 2021 & \checkmark & \checkmark & \checkmark & \checkmark &  &  & \checkmark &  & \checkmark &  & \checkmark &  &  &  & \checkmark &  &  \\
\href{www.megaverse.info}{Megaverse} \cite{Petrenko2021MABench1} & 2021 & \checkmark & \checkmark & \checkmark & \checkmark &  &  &  &  & \checkmark &  &  \checkmark &  &  &  &  &  & \checkmark \\
\href{https://ai4ce.github.io/V2X-Sim/}{V2X-Sim} \cite{Li2022MABench18} & 2022 & \checkmark &  &  & \checkmark & \checkmark &  & \checkmark &  & \checkmark &  & \checkmark &  &  &  & \checkmark &  &  \\
\href{https://github.com/PKU-MARL/DexterousHands}{Bi-DexHands} \cite{Chen2022MABench2} & 2022 & \checkmark & \checkmark & \checkmark & \checkmark & \checkmark &  & \checkmark & \checkmark & \checkmark &  & \checkmark &  &  &  &  & \checkmark &  \\
\href{https://github.com/boschresearch/mrp_bench}{MRP-Bench} \cite{Schaefer2023MABench3} & 2023 & \checkmark & \checkmark & \checkmark &  & \checkmark &  & \checkmark & \checkmark & \checkmark &  & \checkmark &  &  &  &  \checkmark &  &  \\
\href{https://lemma-benchmark.github.io}{LEMMA} \cite{Gong2023MABench4} & 2023 & \checkmark & \checkmark & \checkmark & \checkmark &  & \checkmark &  &  & \checkmark &  & \checkmark &  &  &  &  & \checkmark &  \\
\href{https://github.com/UMass-Foundation-Model/CHAIC}{CHAIC} \cite{Du2024MABench16} & 2024 & \checkmark & \checkmark & \checkmark & \checkmark &  &  & \checkmark &  & \checkmark &  & \checkmark &  &  &  &  &  & \checkmark \\
\href{https://project-roco.github.io}{RoCo} \cite{Mandi2024EnvReflection1} & 2024 & \checkmark & \checkmark & \checkmark & \checkmark &  & \checkmark & \checkmark & \checkmark & \checkmark &  &  & \checkmark & \checkmark &  & \checkmark & \checkmark &  \\
\href{https://github.com/nsidn98/LLaMAR}{MAP-THOR} \cite{Nayak2024MABench15} & 2024 & \checkmark & \checkmark & \checkmark & \checkmark &  & \checkmark & \checkmark &  & \checkmark &  & \checkmark &  &  &  &  &  & \checkmark \\
\href{https://github.
 com/boschresearch/remroc}{REMROC} \cite{Heuer2024MABench6} & 2024 & \checkmark & \checkmark & \checkmark &  & \checkmark &  & \checkmark &  & \checkmark &  & \checkmark &  &  &  & \checkmark &  &  \\
\href{https://github.com/Dr-Xiaogaren/T2E}{T2E} \cite{Zhang2024MABench7} & 2024 & \checkmark & \checkmark & \checkmark &  &  &  & \checkmark &  & \checkmark &  & \checkmark &  &  &  & \checkmark &  &  \\
\href{https://github.com/ut-amrl/SocialGym2}{SocialGym 2.0} \cite{Chandra2024MABench8} & 2024 &  & \checkmark & \checkmark &  &  &  & \checkmark &  & \checkmark &  & \checkmark &  &  &  & \checkmark &  &  \\
\href{https://proroklab.github.io/cambridge-robomaster}{Cambridge RoboMaster} \cite{Blumenkamp2024MABench9} & 2024 & \checkmark & \checkmark & \checkmark & \checkmark &  &  & \checkmark &  & \checkmark &  &  & \checkmark &  &  & \checkmark &  &  \\
\href{https://pengyu-team.github.io/S3E/}{S3E} \cite{Feng2024MABench10} & 2024 & \checkmark &  &  & \checkmark & \checkmark &  & \checkmark &  & \checkmark & \checkmark &  &  &  &  & \checkmark &  &  \\
\href{https://github.com/facebookresearch/partnr-planner}{PARTNR} \cite{Chang2025MABench11} & 2024 & \checkmark & \checkmark & \checkmark & \checkmark & \checkmark & \checkmark & \checkmark & \checkmark & \checkmark &  & \checkmark &  &  & \checkmark &  &  &  \\
\href{https://embodiedbench.github.io}{EmbodiedBench} \cite{Yang2025MABench12} & 2025 & \checkmark & \checkmark & \checkmark & \checkmark &  & \checkmark & \checkmark &  & \checkmark &  & \checkmark &  &  &  & \checkmark & \checkmark &  \\
\href{https://github.com/vhartman/multirobot-pathplanning-benchmark}{MRMG-Planning }\cite{Hartmann2025MABench13} & 2025 &  & \checkmark &  &  &  &  & \checkmark & \checkmark &  &  & \checkmark &  &  &  & \checkmark & \checkmark &  \\
\href{https://ntu-aris.github.io/caric/}{CARIC} \cite{Cao2025MABench14} & 2025 & \checkmark & \checkmark & \checkmark & \checkmark & \checkmark &  & \checkmark &  & \checkmark &  & \checkmark &  &  &  & \checkmark &  &  \\
\bottomrule
\end{tabular}
}
\end{table*}
In contrast to the rapid advances observed in single-agent embodied intelligence, progress in multi-agent embodied intelligence has been relatively limited, especially in the development of standardized benchmarks. Existing benchmarks in this area frequently target highly specialized tasks or narrowly-defined scenarios, restricting their broader applicability and generalizability. To address this issue, this paper provides a comprehensive and systematic review of current benchmarks for multi-agent embodied intelligence(see Table~\ref{tbl:MABenchmarks} and Figure~\ref{fig:MABenchmarks}). Through detailed analyses and structured comparisons, we aim to offer researchers a clear overview of the state-of-the-art, highlight critical gaps in existing benchmarks, and facilitate the development of more robust, versatile, and widely applicable evaluation frameworks to support future research in this domain.
\begin{itemize}
    \item Grid-MAPF \cite{Stern2019MABench17} is a comprehensive benchmark for MAPF featuring 24 diverse grid maps that cover city layouts, game environments and warehouse-style grids, with 25 scenarios provided for each map.
    It systematically generates source-target pairs using random, clustered or designated methods and supports progressive agent scaling to test solver performance under increasing difficulty. Grid-MAPF formalizes various conflict types (e.g., vertex, edge, swapping) and supports multiple objectives such as makespan and sum of costs, enabling standardized and reproducible comparisons across MAPF algorithms. 
    \item FurnMove \cite{Jain2020MABench19} is a simulated benchmark designed to evaluate interactive furniture rearrangement, requiring agents to perceive, navigate, and manipulate multiple objects to achieve a target room layout. Built upon Habitat 2.0, it offers diverse 3D environments with varying furniture types and spatial configurations. Tasks are framed as multi-step, object-centric transformations toward a predefined goal state. The benchmark defines structured scenarios with different levels of difficulty and introduces comprehensive evaluation metrics, including layout accuracy, motion efficiency, and collision count, making it a robust testbed for embodied planning and control in visually complex environments.
    \item SMARTS \cite{Zhou2021MABench20} is a scalable simulation platform and benchmark suite specifically designed for MARL in autonomous driving. It enables diverse and realistic interaction scenarios through a modular architecture, incorporating heterogeneous social agents from an extensible ``Social Agent Zoo.'' SMARTS provides fine-grained simulation control via ``bubbles'' for localized agent interactions, supports distributed multi-agent training, and maintains compatibility with mainstream MARL libraries. The benchmark includes challenging driving tasks such as double merges and unprotected intersections, and offers a comprehensive set of evaluation metrics covering safety, agility, cooperation, and game-theoretic behavior, establishing it as a powerful testbed for complex MARL-based driving research.
    \item Megaverse \cite{Petrenko2021MABench1} is an eight-task embodied AI benchmark. It features procedurally generated 3D environments designed to evaluate a wide range of agent capabilities, including maze navigation, obstacle traversal, exploration, puzzle solving, object manipulation and memory-based reasoning. The platform achieves high-throughput simulation with over 1M observations per second through GPU-accelerated batched execution. It also supports multi-agent teams and enables infinite random instantiations of tasks, promoting generalization and robustness. As an open-source benchmark, Megaverse facilitates reproducible evaluation of embodied AI methods in navigation, exploration, planning, and manipulation.
    \item V2X-Sim \cite{Li2022MABench18} is a comprehensive benchmark for multi-agent collaborative perception in autonomous driving, supporting realistic vehicle-to-vehicle and vehicle-to-infrastructure communication scenarios. Built upon CARLA-SUMO co-simulation, it provides multi-modal sensor data including RGB, LiDAR, GPS and semantic labels from both vehicles and roadside units. V2X-Sim evaluates collaborative 3D object detection, tracking and bird eye view semantic segmentation under various fusion strategies. With rich annotations, spatial diversity and support for intermediate feature sharing, V2X-Sim offers a rigorous testbed for assessing perception robustness under occlusion, sensor noise and communication limitations.
    \item Bi-Dexhands \cite{Chen2022MABench2} is a unified benchmark built on Isaac Gym \cite{Makoviychuk2021IssacGym} that achieves over 30,000 FPS on a single NVIDIA RTX 3090 by running thousands of environments in parallel, enabling highly efficient RL. It includes twenty bi-manual manipulation tasks inspired by the Fine Motor Sub-test, supporting single-agent, multi-agent, offline, multi-task and meta-RL settings within a coherent benchmark. Unlike prior benchmarks, Bi-DexHands treats each joint, finger and hand as a truly heterogeneous agent, and leverages a diverse set of everyday objects from the YCB \cite{Calli2015YCBdataset} and SAPIEN datasets to evaluate task generalization. The platform also provides rich multi-modal observations including contact forces, RGB images, RGB-D images and point clouds, and aligns task difficulty with human motor development milestones, enabling direct comparisons between robotic performance and infant dexterity.
    \item MRP-Bench \cite{Schaefer2023MABench3} is an open-source benchmark for evaluating full-stack multi-robot navigation systems, integrating both centralized and decentralized MAPF algorithms as well as task assignment methods within realistic 3D environments from intralogistics and household domains. Built upon ROS2 \cite{Steven2022ROS2}, Gazebo \cite{Koenig2004Gazebo}, Nav2 package and the robotics middle-ware framework, it supports end-to-end evaluation of planning and coordination methods. MRP-Bench offers modular interfaces, configurable scenarios and standardized evaluation metrics including success rate, planning time, makespan and execution quality, enabling direct comparisons of multi-robot solutions under cluttered, dynamic and complex conditions.
    \item LEMMA \cite{Gong2023MABench4} is a language-conditioned benchmark for multi-robot manipulation, consisting of eight procedurally generated, long-horizon tabletop tasks. Each involved task is designed to require true collaboration, as successful completion depends on two heterogeneous robot arms operating under minimal high-level or crowd-sourced natural language instructions. To support learning and evaluation, the benchmark provides 6,400 expert demonstrations paired with corresponding language commands, along with rich multi-modal observations such as RGB-D images and fused point clouds. The tasks are structured to enforce strong temporal dependencies and require exclusive multi-agent participation, reflecting the complexity of real-world collaborative manipulation. Evaluation is based on standardized success-rate metrics under a strict 100-second time constraint, enabling comprehensive assessment of agents’ capabilities in language-guided planning, task allocation, and fine-grained coordination.
    \item CHAIC \cite{Du2024MABench16} introduces the first large-scale benchmark for embodied social intelligence with a specific focus on accessibility. It simulates interactions between helper agents and four types of physically constrained human partners, including wheelchair users, children, frail adults, and bicyclists. These interactions are carried out across eight long-horizon tasks set in both indoor and outdoor environments, many of which involve emergency scenarios. To successfully assist their partners, agents must infer individual goals and physical limitations based on egocentric RGB-D observations, and generate user-adaptive cooperative plans grounded in realistic physics. The benchmark evaluates performance using metrics such as transport success rate, efficiency improvement, and goal inference accuracy, enabling rigorous assessment of socially aware embodied collaboration. 
    \item RoCo \cite{Mandi2024EnvReflection1} is a benchmark consisting of six tasks for evaluating multi-robot manipulation. It is specifically designed to probe key dimensions of collaboration, including task decomposition into parallel or sequential subtasks, symmetry of observations between agents, and varying degrees of workspace overlap. The environment is built on MuJoCo and is accompanied by a text-only dataset. Tasks in RoCo involve everyday household objects and require agents to demonstrate zero-shot adaptation, feedback-driven re-planning and language-grounded task reasoning. The evaluations in RoCo are based on metrics including success rate, environment step efficiency and the number of re-planning attempts, allowing rigorous assessment of both high-level coordination and low-level motion planning performance.
    \item MAP-THOR \cite{Nayak2024MABench15} is a benchmark consisting of 45 long-horizon, language-instructed multi-agent tasks situated in the AI2-THOR simulator under partial observability. To evaluate spatial generalization, each task is instantiated across five distinct floor plans. The benchmark also categorizes tasks into four levels according to the explicitness of item types, quantities, and target locations. These levels span a spectrum from fully specified instructions to entirely implicit goals, progressively increasing the demands on planner robustness in the face of linguistic ambiguity. To support consistent and comprehensive evaluation, the benchmark adopts standardized metrics such as success rate, transport rate, coverage, balance, and average decision steps. Together, these metrics enable rigorous end-to-end comparisons of multi-agent planning approaches without relying on privileged simulator information. 
    \item REMROC \cite{Heuer2024MABench6} is a modular, open-source benchmark designed to evaluate multi-robot coordination algorithms in realistic and unstructured environments that are shared with humans. It is built on ROS2, Gazebo Ignition and Navigation2, and supports heterogeneous robot platforms, configurable environments with varying human densities, and integrated human motion models. To enable systematic assessment, the benchmark records a variety of performance metrics, including time-to-goal and path efficiency, which allow researchers to quantify the impact of human presence on coordination effectiveness.
    \item T2E \cite{Zhang2024MABench7}  is a 2D simulation benchmark where heterogeneous captor and target robots must coordinate to trap opponents, rather than simply pursue them. This is achieved by combining inter-agent cooperation with environmental constraints. The trapping process is formalized using the absolutely safe zone (ASZ) metric, which measures the degree of spatial confinement. The benchmark includes open-source environments based on real-world maps and provides tailored MARL baselines through a unified TrapNet architecture. It supports both fully cooperative and competitive evaluation settings. Standardized metrics, including completion time, success rate, path length, and ASZ area change, are used to rigorously assess collaborative behaviors in complex spatial layouts. 
    \item SocialGym 2.0 \cite{Chandra2024MABench8} is a multi-agent social navigation benchmark for training and evaluating autonomous robots with individual kino-dynamic constraints. It supports a variety of environments including both open spaces and geometrically constrained areas such as doorways, hallways, intersections and roundabouts. The simulator integrates the PettingZoo \cite{Terry2021PettingZoo} with ROS messaging, allowing for fully customizable observation and reward functions. SocialGym2.0 includes a set of social mini-games along with realistic building layouts. It defines evaluation metrics such as success rate, collision rate and stopping time, which are used to rigorously assess policy performance in diverse social interaction scenarios.
    \item Cambridge RoboMaster \cite{Blumenkamp2024MABench9} offers an agile and low-cost research testbed for MAS by retrofitting DJI RoboMaster S1 platforms with advanced onboard compute such as NVIDIA Jetson Orin, custom ROS2 drivers, and dual-network communication capabilities using both CAN and WiFi. The platform supports seamless sim-to-real transfer through integration with the VMAS framework\cite{Bettini2024VMAS}, allowing zero-shot deployment of both traditional model-based controllers and decentralized MARL policies driven by GNNs. This hardware-software integration enables real-world experimentation under controlled yet realistic conditions. The testbed supports a variety of embodied multi-agent tasks, including trajectory tracking, visual SLAM, and cooperative navigation, thereby facilitating reproducible, scalable, and physically grounded evaluation of coordination strategies and policy generalization in embodied systems.
    \item S3E \cite{Feng2024MABench10} is a large-scale benchmark for collaborative SLAM (C-SLAM). It features synchronized 360$\circ$ LiDAR, high-resolution stereo imagery, high-frequency IMU and novel UWB ranging data collected from three mobile platforms. The dataset covers 18 indoor and outdoor sequences and totals over 263 GB. To support systematic evaluation, S3E defines four representative trajectory patterns: concentric circles, intersecting circles, intersection curve and rays. These patterns test the ability of C-SLAM methods to handle both intra- and inter-loop closures under different levels of spatial overlap.
    \item PARTNR \cite{Chang2025MABench11} ~\cite{Chang2025MABench11} is a large-scale benchmark for embodied human–AI collaboration driven by natural language. It includes 100,000 procedurally generated tasks distributed across 60 diverse multi-room houses containing 5,819 unique objects. All the included tasks are divided into four types: constraint-free, spatial, temporal and heterogeneous, which are designed to evaluate long-horizon planning and inter-agent coordination under realistic physical conditions. Each instruction is paired with a simulation-grounded evaluation function that is automatically generated, enabling rigorous assessment of task completion. The benchmark is built using a semi-automated pipeline that integrates LLM-based task generation, evaluation synthesis, in-the-loop simulation filtering, and human validation. This design enables PARTNR to reach an unprecedented scale and diversity. It also highlights the performance gap between current LLM-based planners and human collaborators in complex multi-agent embodied tasks.
    \item EmbodiedBench \cite{Yang2025MABench12} provides a comprehensive suite of 1,128 testing instances across four embodied environments. These include household scenarios (EB-ALFRED and EB-Habitat) and low-level control domains (EB-Navigation and EB-Manipulation). The tasks span both high- and low-level action hierarchies and are organized into six capability-oriented subsets: base, common sense, complex instruction, spatial awareness, visual perception and long-horizon reasoning. The benchmark supports fine-grained, multi-modal evaluation of 19 state-of-the-art proprietary and open-source multi-modal large language models (MLLMs), enabling systematic assessment across a wide range of embodied AI challenges.
    \item MRMG-Planning \cite{Hartmann2025MABench13} formalizes the problem of multi-modal, multi-robot, multi-goal path planning in continuous spaces. The benchmark includes 21 diverse scenarios, ranging from simple validation tasks to complex cases involving up to five heterogeneous robots and 74 sequential goals. MRMG-Planning supports dynamic environment changes by modeling mode transitions that modify both robot assignments and scene geometry, such as during object handovers. Each scenario is instantiated in a composite configuration space and paired with two baseline planners based on RRT* and PRM*, both of which are asymptotically optimal and probabilistically complete. This design enables rigorous evaluation of end-to-end planning performance under varying robot counts, goal sequences and cost formulations. 
    \item CARIC \cite{Cao2025MABench14} provides a benchmark for heterogeneous multi-UAV inspection, in which LiDAR-equipped explorers and camera-only photographers must collaboratively map unknown structures using only bounding box constraints while capturing high-quality structural imagery. CARIC incorporates realistic limitations on kinematics, communication and sensing capabilities. It includes a variety of inspection scenarios such as buildings, cranes and aircraft, and defines rich evaluation metrics that decompose the inspection score into components for visibility, motion blur and spatial resolution. These metrics enable systematic evaluation of task allocation and planning under tight constraints and support reproducible comparisons for real-world aerial inspection tasks demonstrated in CDC 2023 and IROS 2024. 
\end{itemize}

\section{The Challenges and Future Works} \label{sec:disscussions} %

Despite of the rapid progress of embodied AI and the primary development of multi-agent embodied AI, it faces several challenges and presents exciting future directions.
\paragraph{Theory for complex Embodied AI interaction.}
Building upon the theoretical foundations provided by Markov games, MARL has introduced various frameworks for effectively modeling cooperative relationships among agents in complex environments~\cite{zhang2021multi}. Methods such as value decomposition (e.g., VDN~\cite{Sunehag2017VDN}, QMIX~\cite{rashid2020monotonic}) have improved scalability by decomposing joint objectives into individual utilities, while counterfactual reasoning approaches like COMA~\cite{foerster2018counterfactual} have significantly enhanced attribution accuracy for individual agent contributions. Advances in transfer learning and networked MARL have mitigated challenges posed by agent heterogeneity and environmental non-stationarity. Additionally, the integration of control and game theory has fostered effective coordination strategies, exemplified by consensus protocols and distributed task allocation methods~\cite{Chu2020Multi-agent}. However, embodied multi-agent intelligence introduces unique theoretical challenges, including asynchronous sensing, delayed actions, limited observability, communication constraints, and substantial heterogeneity, complicating both theoretical formulation and practical implementation. Furthermore, while large generative models, particularly LLMs, offer significant potential in facilitating planning and inter-agent communication, their theoretical characteristics such as stability, generalization and interpretability still remain inadequately understood~\cite{strachan2024testing}. To address these limitations, future research should explore new theoretical paradigms specifically tailored to embodied MAS, leveraging approaches such as causal inference to uncover inter-agent dependencies, complex systems theory to understand emergent behaviors, and hierarchical, bio-inspired coordination frameworks adaptable to real-world complexity.
\paragraph{New algorithm design.}
Recent developments have solidified the role of MARL as a foundational framework for cooperative and competitive multi-agent scenarios. Notable algorithms, including QMIX~\cite{rashid2020monotonic} with its monotonic value decomposition and MAPPO~\cite{yu2022surprising}, an adaptation of proximal policy optimization, have demonstrated compelling results in benchmark environments such as SMAC and Google research football (GRF)\cite{Yuan2023OpenSurvey}. Nevertheless, these successes predominantly rely on the CTDE paradigm\cite{amato2024introduction}, assuming ideal conditions and unrestricted access to agent actions—conditions rarely achievable in embodied scenarios. Physical deployment introduces significant complexities such as sensor noise, limited actuation, delayed feedback, and partial observability. Moreover, existing multi-agent learning frameworks typically struggle with generalization beyond training distributions and have limited scalability when dealing with heterogeneous teams. Embodied multi-agent tasks involve multi-modal sensory data, diverse capabilities, and dynamic interactions, necessitating alternative algorithmic frameworks beyond traditional paradigms such as CTDE, DTDE (decentralized training and execution), or CTCE (centralized training and execution). Promising future directions include hierarchical coordination structures, agent-grouping mechanisms, and the integration of structured priors or classic control theories to achieve scalable and robust performance in embodied MAS.

\paragraph{Effective and efficient learning.}
Most multi-agent learning methodologies have been developed in simulation or gaming environments, benefiting from low-cost, highly repeatable scenarios that allow extensive policy optimization through frequent interaction sampling. However, embodied multi-agent tasks exponentially increase complexity, severely reducing sample efficiency and complicating joint exploration due to vast state and policy spaces~\cite{albrecht2024multi}. These challenges are amplified in real-world contexts, where each interaction incurs considerable time, financial cost, and hardware wear~\cite{yu2018towards}. Although recent advancements in single-agent embodied learning—such as extracting knowledge from large multi-modal models like LLMs~\cite{cao2024survey}, employing world-model-based simulations for policy rollouts~\cite{luo2024survey}, and leveraging offline data-driven policy calibration~\cite{jiang2024multi}—have improved sample efficiency, these strategies often struggle in multi-agent contexts owing to complex interactions and non-stationary dynamics. To overcome these barriers, specialized multi-agent world models accurately simulating interactive dynamics, exploration strategies informed by structured prior knowledge, rapid initialization through IL, and meta-learning for improved generalization are necessary. Additionally, developing robust methodologies supporting effective multi-task and sim-to-real transfer is vital for real-world applicability.

\paragraph{Large generative models assisted learning.}
Recent advances in pre-trained large models, such as GPT-4, PaLM, CLIP, SAM and Gemini, have significantly reshaped language, vision and multi-modal learning landscapes by offering powerful capabilities in representation, perception, and reasoning~\cite{huang2024large}. These foundation models, trained on vast and diverse datasets, provide rich prior knowledge and robust cross-modal alignment, presenting promising avenues for enhancing embodied AI. Leveraging these models allows agents to attain deeper semantic understanding, superior generalization, and more adaptive interactions. However, directly deploying these foundation models in embodied multi-agent scenarios remains challenging~\cite{Chen2024MALLMPlanning4}, as they typically derive from static, single-agent contexts and lack inductive biases essential for dynamic interactions involving asynchronous communication, partial observability, tightly coupled policies, and non-stationarity. Furthermore, embodied MAS usually involve high-dimensional multi-modal inputs, extensive action spaces and sparse feedback, limiting the adaptability and efficiency of the learning algorithms. Insights from recent single-agent embodied learning—such as few-shot adaptation, multi-modal joint pretraining and simulation-driven data augmentation provide valuable transferable knowledge~\cite{ma2024survey} for multi-agent research. Future research should focus on developing scalable multi-agent pretraining paradigms, integrating foundation models with reinforcement learning and graph-based coordination methods, and enhancing sim-to-real transfer, thereby forming robust and generalizable frameworks for multi-agent embodied AI in complex, open-ended environments.

\paragraph{General multi-agent Embodied AI framework.}
Recent advancements in multi-agent embodied AI have primarily emphasized solving specific tasks within constrained scenarios. While progress is notable, this focus significantly limits the generalizability and scalability of policies. Benchmarks such as SMAC and GRF have driven progress in MARL, yet existing methods continue to struggle with variable task objectives, environmental dynamics, and agent team compositions~\cite{Yuan2023OpenSurvey}. Single-agent generalist models like Gato~\cite{reed2022a} and RT-X~\cite{Neill2024SABenchmark} demonstrate unified architectures' potential in broadly generalizing across vision, language, and control domains. Extending generalist principles to multi-agent settings introduces additional complexity from inherent non-stationarity, intricate agent interactions, and multiple equilibria. Moreover, current architectures often assume fixed team sizes and agent homogeneity, restricting scalability and adaptability. Recent innovations such as the multi-agent Transformer (MAT)~\cite{wen2022multi} address scalability via permutation-invariant representations but substantial challenges remain, particularly in integrating multimodal knowledge (e.g., physical embodiment, social reasoning, linguistic interaction) into unified frameworks. Future advancements will require equilibrium-aware training methodologies, modular and scalable architectures, and robust multimodal representation learning techniques, enabling reliable coordination and generalization in dynamic multi-agent environments.

\paragraph{Adaptation to open environments.}
Unlike the stable and closed task settings that commonly assumed in embodied AI research, scenarios with open environments~\cite{zhou2022open} present significant challenges for embodied MAS due to their inherently dynamic, uncertain and non-stationary characteristics. Agents operating within such environments often encounter unpredictable sensory inputs, shifting reward structures, evolving tasks and frequently changing agent populations. For instance, the environments may include noisy observations, delayed or disrupted feedback and task transitions ranging from navigation and exploration to cooperative or competitive interactions. Additionally, the shifting team compositions with teammates and adversaries changing unpredictably further complicate coordinations. The human involvement also introduces extra layers of uncertainty due to diverse, less predictable behaviors. Such variability undermines the consistency between training and deployment environments, revealing the fragility of policies~\cite{Yuan2023OpenSurvey}. Thus, achieving effective performance in open environments requires agents with robust policies resilient to sensory and environmental disturbances, continual learning capabilities that integrate new experiences without compromising previously acquired skills and contextual reasoning abilities for dynamic interactions and coordinations with novel agents. Methods enabling rapid strategy adaptation such as policy reuse, latent variable modeling and meta-learning have become indispensable for generalizing quickly to unseen circumstances. Although recent advances in robust MARL, lifelong learning, and context-aware decision-making are promising, critical challenges remain unsolved. Future research should prioritize accurate modeling of open-environment dynamics, anticipating heterogeneous agent and human behaviors, and enabling real-time algorithmic adaptation.
\paragraph{Evaluation and verification.}
Establishing rigorous and comprehensive evaluation frameworks is crucial yet challenging for advancing multi-agent embodied AI. Recent platforms, including CHAIC~\cite{Du2024MABench16} (further benchmarks in Table~\ref{tbl:MABenchmarks}), have provided valuable resources but often suffer from limited representational complexity and insufficient ecological validity. Current benchmarks predominantly focus on single-modality tasks, neglecting essential multi-modal signals (vision, language, audio, video) necessary for realistic perception and decision-making. Furthermore, benchmarks commonly involve small-scale, homogeneous MAS, inadequately representing the complexity encountered in heterogeneous MAS comprising diverse embodiments such as drones, robotic arms, autonomous vehicles, and quadrupedal robots. Additionally, idealized assumptions such as ignoring partial observability, communication latency, adversarial behaviors and asynchronous execution widen the sim-to-real gap, undermining policy robustness and transferability. Subsequently, future research should aim to develop general-purpose, modular and scalable evaluation frameworks analogous to Gym, MuJoCo or PyMARL~\cite{samvelyan2019starcraft}, incorporating multi-modal interactions, heterogeneous coordination and reproducibility features. Physical testbeds, including robot soccer platforms (e.g., RoboCup~\cite{farias2025robocup}) and drone swarm arenas are also essential to validate real-world feasibility and bridge sim-to-real gaps. Moreover, the absence of standardized and interpretable evaluation metrics remains unresolved, as existing measures (e.g., coordination scores, network-based metrics) often lack consistency and generalizability. Therefore, developing unified benchmarks with clearly defined tasks, comprehensive evaluation criteria (robustness, scalability, energy efficiency, behavioral diversity), open-access leader-boards, and formal verification methods is imperative, especially for safety-critical applications.
\paragraph{Applications and implementations.}
Multi-agent embodied AI holds immense promise across diverse domains including robotics, education, healthcare, military operations, interactive simulations and smart-city infrastructures. In robotics, particularly within industrial manufacturing, warehouse logistics, and autonomous driving, successful deployment necessitates robust real-time coordination, reliable collision avoidance, and adaptable platforms, emphasizing algorithmic generalizability. Educational applications involving virtual agent teams demand controlled, socially appropriate behaviors, thus placing substantial requirements on advanced natural language processing and affective computing capabilities. In healthcare, integrating robotic agents within medical teams highlights the necessity for reliable decision-making, robust privacy safeguards, and strict adherence to ethical standards. Military applications, such as coordinated drone swarms, further underline these challenges, stressing policy robustness under adversarial conditions and the crucial role of human oversight to mitigate ethical and operational risks associated with autonomy. Interactive simulations and competitive gaming environments, exemplified by platforms like OpenAI Five~\cite{berner2019dota} and AlphaStar~\cite{vinyals2019grandmaster}, have provided valuable benchmarks but face limitations due to high computational requirements and difficulties in transferring learned strategies to real-world contexts. Smart-city scenarios, including traffic management and energy-grid optimization, involve extensive interactions among numerous autonomous agents, necessitating highly reliable, efficient, and safe solutions with direct societal impacts. Addressing these diverse yet interconnected challenges requires integrating expert domain knowledge, rigorous validation procedures, and comprehensive safety frameworks, enabling practical real-world deployments of embodied intelligent MAS.
\paragraph{Other future directions.}
While the previously sections highlight significant challenges within multi-agent embodied AI, several additional fundamental issues warrant deeper investigation. First, multi-modal perception and collaborative learning remain underexplored, with current research typically focusing on isolated sensory modalities and rarely effectively integrating vision, audio and language inputs. Future studies should thus prioritize developing robust cross-modal fusion methods and efficient communication protocols to handle modality disparities, information inconsistencies and latency issues~\cite{duan2024multimodal}. Second, social learning presents a promising pathway toward emergent collective AI. Although approaches such as IL and policy distillation enable agent populations to exhibit coordination surpassing individual capabilities, theoretical foundations and conditions necessary for stable emergent behaviors remain inadequately understood~\cite{ndousse2021emergent}. Lastly, as MAS are increasingly gain autonomy, addressing safety and ethical considerations becomes critically important, especially within sensitive or high-stakes environments~\cite{ruane2019conversational}. Explicit safety constraints, clear ethical guidelines and transparent explainability mechanisms must be developed to ensure fair, accountable and interpretable decision-making, thereby minimizing unintended risks arising from complex agent interactions and ensuring trustworthy AI deployments.
\section{Conclusion} \label{sec:conclusions} 
Embodied AI equips agents with the capability to sense, perceive and effectively interact with various entities in both cyberspace and the physical world, marking a significant step toward achieving artificial general intelligence. To provide a systematic understanding of this rapidly evolving field, this survey comprehensively reviews recent developments in embodied AI, bridging the gap from single-agent to multi-agent scenarios, and further discusses key challenges and promising future directions.. Specifically, we first introduce foundational concepts essential to embodied AI. We then systematically examine critical methodologies, including classic control and planning approaches, learning-based techniques, and generative-model-based frameworks. Additionally, widely-adopted benchmarks employed for evaluating embodied AI in both single-agent and multi-agent contexts are carefully discussed. Finally, we highlight significant challenges and explore promising future directions within this emerging research domain. Through a comparative analysis of recent advances, this survey aims to offer researchers a clear and integrated perspective, thereby facilitating continued innovation in multi-agent embodied AI.

\bibliography{fzh}
\bibliographystyle{ACM-Reference-Format}
\end{document}